\documentclass[10pt,journal,compsoc]{IEEEtran}

\usepackage{amsmath}
\usepackage{amssymb}
\usepackage{mathptmx}      
\usepackage{algorithm}
\usepackage{amsthm}
\usepackage{algorithmic}
\usepackage{subfigure}
\usepackage{graphicx}
\usepackage{tabularx}
\usepackage{color}
\usepackage{caption}
\usepackage{ragged2e}
\usepackage{epsfig}
\usepackage{algorithm}
\usepackage{multirow}
\usepackage{mathrsfs}
\usepackage{mathtools}
\usepackage{enumitem}
\usepackage{tensor}
\usepackage{xcolor}
\usepackage{booktabs}
\usepackage{balance}
\usepackage{latexsym}
\usepackage{multirow}
\usepackage{color, colortbl}
\usepackage[normalem]{ulem}
\algsetup{linenosize=\small}
\newcommand{\SKIP}[1]{} 

\newcommand{\mbegin} {\left [ \begin{array}}

\newcommand{\mend}   {\end{array} \right ]}

\newcommand{\detbegin} {\left | \begin{array}}
\newcommand{\detend}   {\end{array} \right |}

\newcommand{\vbegin} {\left ( \begin{array}{c}}

\newcommand{\vend} {\end{array}\right )}

\def\squareforqed{\hbox{\rlap{$\sqcap$}$\sqcup$}}

\def\qed{\ifmmode\squareforqed\else{\unskip\nobreak\hfil
	\penalty50\hskip1em\null\nobreak\hfil\squareforqed
	\parfillskip=0pt\finalhyphendemerits=0\endgraf}\fi}

\newcommand{\showeqnlabel}{
	\hbox to 0pt{\quad\quad\relax\fbox{\scriptsize\rm\eqnlblx}%
	\gdef\eqnlblx{xxxx}}} \newcommand{\eqnlblx}{}

\def\@eqnnum{\rm (\theequation)\showeqnlabel}

\newcommand{\nofig}[1]{\centerline{\bf Figure here}}

\def\mat#1{\mathchoice{\mbox{\bf$\displaystyle\tt#1$}}
	{\mbox{\bf$\textstyle\tt#1$}}
	{\mbox{\bf$\scriptstyle\tt#1$}}
	{\mbox{\bf$\scriptscriptstyle\tt#1$}}}
\def\m#1{\protect\mat #1}

\pdfoutput=1
\DeclareRobustCommand{\rchi}{{\mathpalette\irchi\relax}}
\newcommand{\irchi}[2]{\raisebox{\depth}{$#1\chi$}}

\theoremstyle{definition}
\newtheorem{definition}{Definition}[section]
\definecolor{mygreen}{rgb}{0.000, 0.392, 0.000}
\definecolor{Gray}{gray}{0.9}

\ifCLASSOPTIONcompsoc
  \usepackage[nocompress]{cite}
\else
  \usepackage{cite}
\fi
\ifCLASSINFOpdf
\else
\fi
\begin{document}

%
\title{Dense Non-Rigid Structure from Motion: A Manifold Viewpoint}
%
%
%
%

\author{Suryansh Kumar, Luc Van Gool, Carlos E. P. de Oliveira, Anoop Cherian, Yuchao Dai, Hongdong Li
\IEEEcompsocitemizethanks{\IEEEcompsocthanksitem Suryansh Kumar, Luc Van Gool, Carlos E.P Oliveira is with Computer Vision Lab at ETH Z\"urich, Switzerland. \protect\\
E-mail: \{sukumar, vangool, coliveira\}@vision.ee.ethz.ch

\IEEEcompsocthanksitem Anoop Cherian is with MERL Cambridge, MA, USA. \protect\\
E-mail: cherian@merl.com
\IEEEcompsocthanksitem Yuchao Dai is with Northwestern  Polytechnical  University, X'ian China \protect\\
E-mail: daiyuchao@gmail.com

\IEEEcompsocthanksitem Hongdong Li is with Australian National University, Canberra \protect\\
E-mail: hongdong.li@anu.edu.au
}
}

\IEEEtitleabstractindextext{%
\justify
\begin{abstract}
Non-Rigid Structure-from-Motion (NRSfM) problem aims to recover 3D geometry of a deforming object from its 2D feature correspondences across multiple frames. Classical approaches to this problem assume a small number of feature points and, ignore the local non-linearities of the shape deformation, and therefore, struggles to reliably model non-linear deformations. Furthermore, available dense NRSfM algorithms are often hurdled by scalability, computations, noisy measurements and, restricted to model just global deformation.  In this paper, we propose algorithms that can overcome these limitations with the previous methods and, at the same time, can recover a reliable dense 3D structure of a non-rigid object with higher accuracy. Assuming that a deforming shape is composed of a union of \emph{local} linear subspace and, span a \emph{global} low-rank space over multiple frames enables us to efficiently model complex non-rigid deformations. To that end, each local linear subspace is represented using Grassmannians and, the global 3D shape across multiple frames is represented using a low-rank representation. We show that our approach significantly improves accuracy, scalability, and robustness against noise. Also, our representation naturally allows for simultaneous reconstruction and clustering framework which in general is observed to be more suitable for NRSfM problems. Our method currently achieves leading performance on the standard benchmark datasets.

\end{abstract}

}

\maketitle

\IEEEdisplaynontitleabstractindextext

%
\IEEEpeerreviewmaketitle

\ifCLASSOPTIONcompsoc
\IEEEraisesectionheading{\section{Introduction}\label{sec:introduction}}
\else
\section{Introduction}\label{sec:introduction}
\fi
\IEEEPARstart{I}{n} this work, we will be focused on the problem of Dense Non-rigid Structure from Motion (NRSfM). Generally, the goal of this problem is to solve dense 3D shape of a non-rigidly deforming object in the scene from its per pixel image correspondences across multiple frames. Application that benefits from dense NRSfM includes animation \cite{jain2010augmenting}, motion capture \cite{akhter2009nonrigid}, 3D facial expression capture \cite{garg2013dense}, human heart 3D model for bypass surgery \cite{garg2013dense} and many more. These examples demonstrate that NRSfM is central to a wide range of important real-world applications and therefore, a reliable solution to NRSfM can benefit several areas in science and engineering.

There are different ways to solve non-rigid structure-from-motion problem, among them, matrix factorization is one of the most popular and a well-known approach to find a solution to this problem \cite{bregler2000recovering} \cite{akhter2009nonrigid} \cite{dai2014simple} \cite{gotardo2011non}. Under the matrix factorization approach, a measurement matrix (a matrix with 2D trajectories as its column vectors) is decomposed into a motion matrix and a shape matrix \S \ref{ss:struct_motion}. Consequently, any solution to this problem using this approach depends on the proper modeling of \emph{structure}, and an efficient approach to estimate \emph{motion}. Mathematically, one can assume that the 3D shape belongs to some \emph{shape} manifold\footnote{Here, we assume a smooth, continuous surface for dense NRSfM problem} and the motion lies on a differentiable manifold \cite{fecko2006differential}. Keeping this perspective to solve dense NRSfM, it's quite natural to think of this problem in terms of manifolds, and how to model this problem efficiently using manifold representation.

Our survey reveals that the advancements in the non-rigid structure-from-motion for \emph{sparse} set of points has been steady over the years \cite{akhter2009nonrigid} \cite{gotardo2011non} \cite{gotardo2011kernel} \cite{paladini2012optimal} \cite{akhter2011trajectory} \cite{dai2014simple} \cite{lee2013procrustean} \cite{lee2016consensus} \cite{kumar2016multi} \cite{larsson2017compact} \cite{kumar2019non}, yet, the developments in dense NRSfM algorithm has been limited \cite{garg2013dense} \cite{ansari2017scalable} \cite{dai2017dense}. The reason for such a limited development in dense NRSfM is perhaps due to its dependence on per pixel reliable 2D image correspondences, across multiple frames, or the absence of resilient mathematical representation to model dense surface deformation. One can argue on the efficient motion estimation, however, from image correspondences, we can only estimate relative motion, and reliable algorithms with convincing theory exists to perform this task well \cite{dai2014simple} \cite{lee2013procrustean}. Additionally, with the recent developments in learning based approaches, per pixel correspondences can be achieved with a remarkable accuracy \cite{sun2018pwc} \cite{hur2017mirrorflow}, which leaves dense non-rigid shape representation and its modeling as a potential gray area for research in \textbf{dense} NRSfM. 

A  natural way to deal with dense NRSfM is to try classical sparse NRSfM algorithms, which in fact, works quite well for a few sets of points. Our experiments show that the existing sparse NRSfM algorithms do not cascade well to dense NRSfM settings. This is because the assumption and the formulation developed for the sparse NRSfM does not hold entirely for dense deforming surfaces. For example: The assumption that a non-rigid shape spans a global low-rank space  \cite{dai2014simple} \cite{lee2013procrustean}. Now, such an assumption may hold for the global structure of the problem, however, it fails to cater the inherent local deformation of the shape over time and space. Therefore,  dense NRSfM solution using sparse NRSfM formulations provides implausible results. This drawback with \cite{dai2014simple} \cite{lee2013procrustean} led to the development of union of subspace based methods in NRSfM \cite{zhu2014complex} \cite{kumar2016multi} \cite{kumar2017spatio}. Among these methods, Kumar et.al. work on the union of subspace demonstrated state-of-the-art results \cite{kumar2016multi} in the NRSfM challenge at CVPR 2017 \cite{jensen2018benchmark}. Nevertheless, these algorithms do not scale to dense feature points and their resilience to noise and outliers remains unsatisfactory.

In the past, researchers have developed dense NRSfM algorithm as well, but similar to sparse NRSfM they are mostly restricted to global shape constraint \cite{garg2013variational} \cite{garg2013dense} \cite{dai2017dense}. As a result, it fails to exploit the local deformation of the surface. Moreover, the optimization framework proposed by these approaches is critically expensive to process \S \ref{sref:relatedwork}. These deficiencies with past methods made us realize that a dense NRSfM algorithm needs a framework that should be able to exploit both local and global non-linearities, and at the same time must be computationally fast to process. Keeping these standards intact, we developed a new representation and modeling for dense NRSfM problem. Using our new representation, we can apply both local and global shape deformation constraints to model a dense NRSfM problem. We adhere to the assumption that the low-dimensional linear subspace spanned by a deforming shape is valid \textbf{locally} ---in both space and time, along with \textbf{global} low-rank space. Such an assumption about the surfaces has been well studied in topological manifold theory \cite{absil2004riemannian} \cite{dollar2007non}. \\

\noindent
\emph{Global low rank representation}: Matrix factorization approaches to NRSfM assumes that the deforming 3D structure intrinsically spans a low-rank space globally \cite{dai2014simple} \cite{garg2013dense}. This low-rank representation faithfully captures the global behavior of a non-rigidly deforming object over multiple frames. This representation can be obtained using Singular Value Decomposition, however, in the presence of noisy measurements such a solution can provide unsettling results. Due to \cite{elhamifar2009sparse} \cite{candes2011robust} \cite{liu2013robust} its possible to recover a low-rank solution under such circumstances. Dai et.al. research \cite{dai2014simple} ---which is a classical work in NRSfM, draws its inspiration using the following optimization to solve for low-rank non-rigid 3D structure 
\begin{equation}
\begin{aligned}
    \underset{\m X, \m E}{\textrm{argmin}} ~\|\m X\|_* + \|\m E\|_{l} \\
    \textrm{subject to:} ~\m Y = \m X + \m E
\end{aligned}
\end{equation}
Here `$\m Y$', `$\m X$' are the data matrix and its clean low-rank matrix representation respectively, `$\m E$' is the error matrix and $l$ stands for matrix norm ($l$ = 1 or 2). Inspired by Dai et.al. \cite{dai2014simple} sparse NRSfM work, in this paper, we assume that a dense non-rigid structure is of intrinsically low rank globally. We use this assumption to capture the global deformation of the surface. \\

\begin{figure}
\centering
\includegraphics[width=1.0\linewidth] {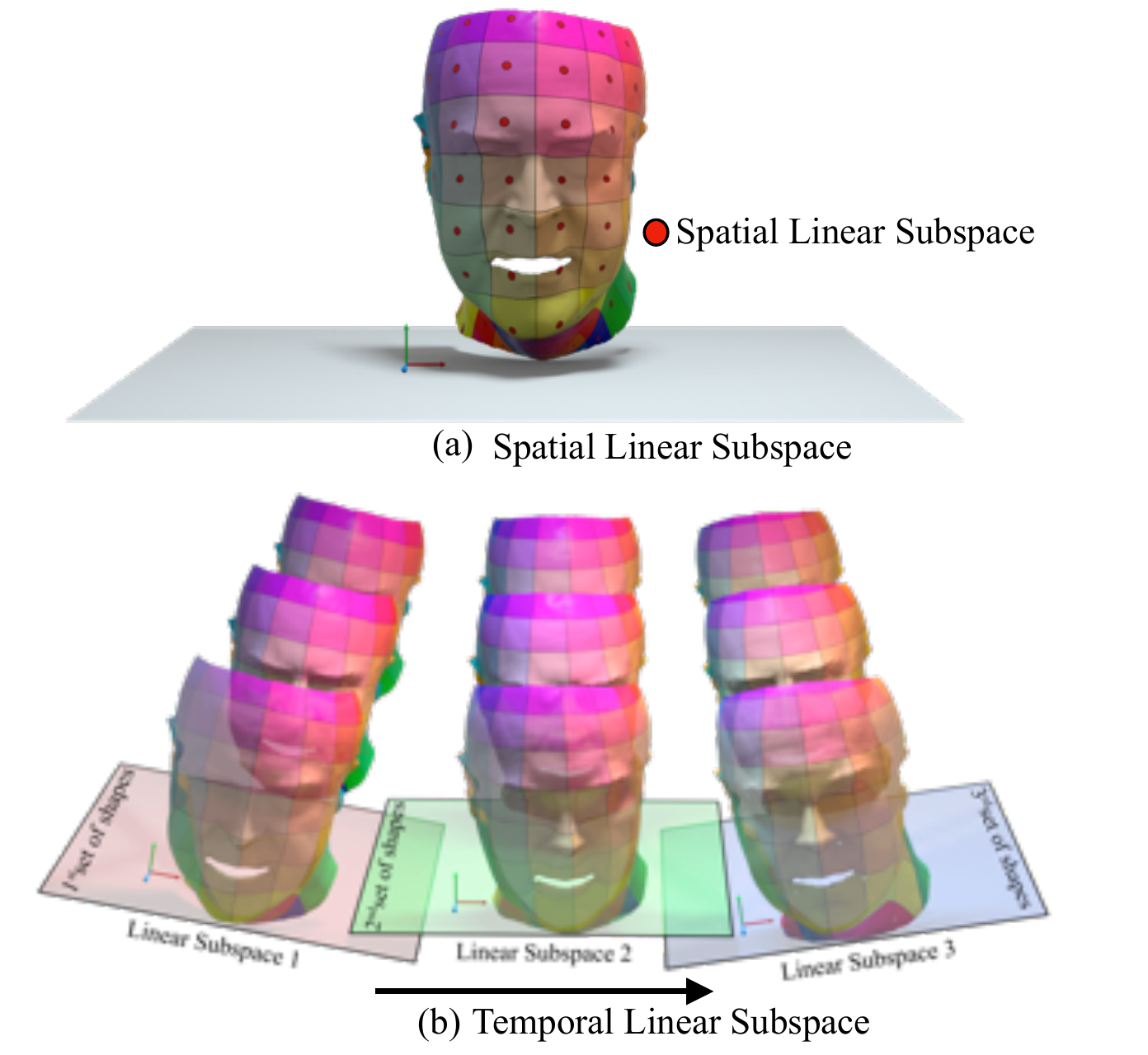}
\caption{\small Illustration show the decomposition of deforming surface into a set of local linear subspace. (a) Local Linear subspace representation of the trajectory space in the spatial domain (b) Local Linear Subspace representation of the shape space in the temporal domain. Each local plane shown in the temporal space can also be represented as a point on the Grassmann manifold. The above result show the reconstruction results using our method which is composed of 73,765 point samples from Actor dataset \cite{beeler2011high}.
}
\label{fig:spatial_temporal_rep}
\end{figure}

\noindent
\emph{Local linear subspace representation}: In contrast to the previous dense NRSfM methods \cite{garg2013dense} \cite{dai2017dense} \cite{ansari2017scalable}, we represent deforming surfaces as a union of locally linear subspace. We argue that most non-rigidly deforming object over time and space is composed of a union of linear subspaces. Therefore, the methods that uses only global nuclear norm to constrain the surface deformation gets a solution on the convex envelop over the underlying multiple subspaces \cite{garg2013dense} \cite{dai2017dense} \cite{ansari2017scalable}. More precisely, these methods look for a solution on the boundaries of a feasible region which is composed of summation of subspaces. As a result, their 3D reconstruction results are empirically inferior. In this work, we aim to jointly recover an implicit local subspace representation of the surface along with its 3D reconstruction for multiple frames. Consequently, we decompose the surface into a set of locally linear subspaces and represent each subspace as a point on a Grassmann Manifold \cite{absil2009optimization} (see Fig.\ref{fig:spatial_temporal_rep}). We will show, such a representation is well-suited for many dense deforming surfaces. Our approximation holds well to capture the local non-linearities in addition to the global deformation. 

Our representation to assign each local linear subspace as a point on a Grassmann manifold not only provides a richer representation to exploit each local linear subspace, but also helps to improve the scalability, robustness, and processing time of our dense NRSfM algorithm. To bootstrap the initial Grassmann points, we group the local trajectories and shapes via k-means++ \cite{arthur2007k}. We compute a large but finite set of linear subspaces and, cast dense NRSfM problem as a joint clustering of subspaces and 3D reconstruction tasks. Our representation easily blends into a joint clustering and reconstruction formulation which provides superior results than performing the two tasks separately \cite{zhu2014complex} \cite{kumar2016multi}. \\

\noindent
\textit{Contributions}: The main contributions of our work are as follows:
\begin{itemize}
    \item A new representation for dense NRSfM problem that utilizes both local and global structure of the deforming shape to solve the problem.
    \item An efficient framework for modeling non-rigidly deforming surface on Grassmann manifold, which jointly supply 3D reconstruction and compact subspace representation of the shape.
    \item A scalable, robust and fast algorithm which does not need any template prior to solve dense NRSfM \cite{yu2015direct} \cite{li2009robust}.
    \item A geometry aware extension that help exploit the Grassmannian representation of different dimensions which is extremely useful in handling noise and high-dimensional Grassmannians.
    
    \item Iterative solution to the proposed optimization based on ADMM \cite{boyd2011distributed} that achieves leading performance on the standard benchmark dataset \cite{garg2013dense} \cite{varol2012constrained}.
\end{itemize}

In addition to the 3D reconstruction accuracy analysis, we performed other relevant experiments and demonstrate the advantage of our formulation using a range of qualitative and quantitative analysis. The present journal paper is based on two CVPR conference papers \cite{kumar2018scalable} \cite{kumar2019jumping}. In this work, we described the approach in greater detail including the representation, modeling of the problem and the implementation of the algorithm. Additionally, how the first proposed algorithm \cite{kumar2018scalable} led the foundation for the development of the next algorithm \cite{kumar2019jumping}. We also present a more detailed derivation of the proposed optimization with deeper statistical analysis, minor corrections and extensive experimental results. Lastly, we provide a concise discussion on the potential limitations of the algorithm, and how it can be improved further for real-world applications. We believe our journal version is much more complete, and provide the readers comprehensive details on the advantages/limitations of Grassmannian representation to solve dense NRSfM under joint 3D reconstruction and subspace clustering framework.








\section{Related Work} \label{sref:relatedwork}
Non-Rigid Structure from Motion (NRSfM) is more than a two-decade-old problem and is still an active area of research in the geometric computer vision. NRSfM using matrix factorization introduced in the Bregler et.al. seminal work \cite{bregler2000recovering} was one the first working algorithm for NRSfM, which in fact, was an extension to the rigid factorization method \cite{tomasi1992shape}. This problem is challenging due to the inherent unconstrained nature of the problem, as many 3D varying configurations can have similar image projections and as a result, the problem remains unsolved for any arbitrary deformations. However, many profound algorithms under some or the other prior assumptions ---about the object deformation or the camera projection, have been proposed to achieve a reliable solution to this problem \cite{dai2014simple} \cite{akhter2011trajectory} \cite{park20103d} \cite{lee2013procrustean} \cite{zhu2014complex} \cite{kumar2017spatio} \cite{kumar2016multi} \cite{garg2013dense} \cite{kumar2020non}.  The literature on this topic is very extensive and therefore, for brevity, we review the works that are of close relevance to the \textbf{dense} NRSfM methods under classical NRSfM setting\footnote{By {classical NRSfM setting}, we mean the input to the algorithm is only image feature correspondences rather than depth or 3D template.}.

Earlier attempts to solve this problem used piecewise 3D reconstruction of the shape parts, which were further processed via a stitching step to get a global 3D shape \cite{collins2010locally} \cite{russell2012dense}.  To our knowledge, Garg et.al. \cite{garg2013variational} variational approach was one of the first to propose and demonstrate a practical dense NRSfM algorithm that do not rely on a 3D template prior. This method introduced a discrete total variational constraint with trace norm constraint on the global shape, which leads to a biconvex optimization problem. Despite the algorithm's outstanding performance, it's computationally expensive and needs a \textbf{GPU} to provide the solution.

Recently, Dai et.al. \cite{dai2017dense} extended his simple prior free approach \cite{dai2014simple} to solve dense NRSfM problem. They proposed a spatial-temporal formulation to tackle the problem. The author revisits the temporal smoothness term from \cite{dai2014simple} and integrate it with a spatial smoothness term using the Laplacian of the non-rigid shape. The resultant optimization leads to a series of least squares to be minimized, thus making it \textbf{extremely slow} to process, hence, not scalable for practical settings.

The consecutive frame-based formulation in recent years has shown some promising results to solve dense 3D reconstruction of a general dynamic scene, including non-rigid object \cite{kumar2017monocular,ranftl2016dense}. Nevertheless, motion segmentation, triangulation, as rigid as possible assumption, scale consistency and inter-frame consistency quite often breaks down for the deforming object. Therefore, these methods are still not mature to solve dense NRSfM for multiple frames. Not long ago, Gallardo et.al. \cite{gallardo2017dense} combined shading, motion and generic physical deformation to model dense NRSfM. Nevertheless, such information is generally not available on many real-world devices.

Other variants of dense non-rigid structure from motion algorithm involve solving the problem in a sequential manner \textit{i.e.},
rather than using an entire batch of frames, solve for dense 
3D reconstruction as the image arrives. However, the proposed method under such a setting uses an initial set of frames to initialize the algorithm using a rigid factorization algorithm \cite{tomasi1992shape, marquez2008optimal}. Such heuristics greatly limits the use of such a sequential method to real-world scenarios. Recently proposed method CMDR \cite{golyanik2019intrinsic} proposed a hybrid approach extracts prior shape knowledge from an input sequence and, uses it as a dynamic shape prior for sequential surface recovery.

\begin{figure}
    \subfigure[\label{fig:nmanifold1}]{\includegraphics[width=0.5\linewidth]{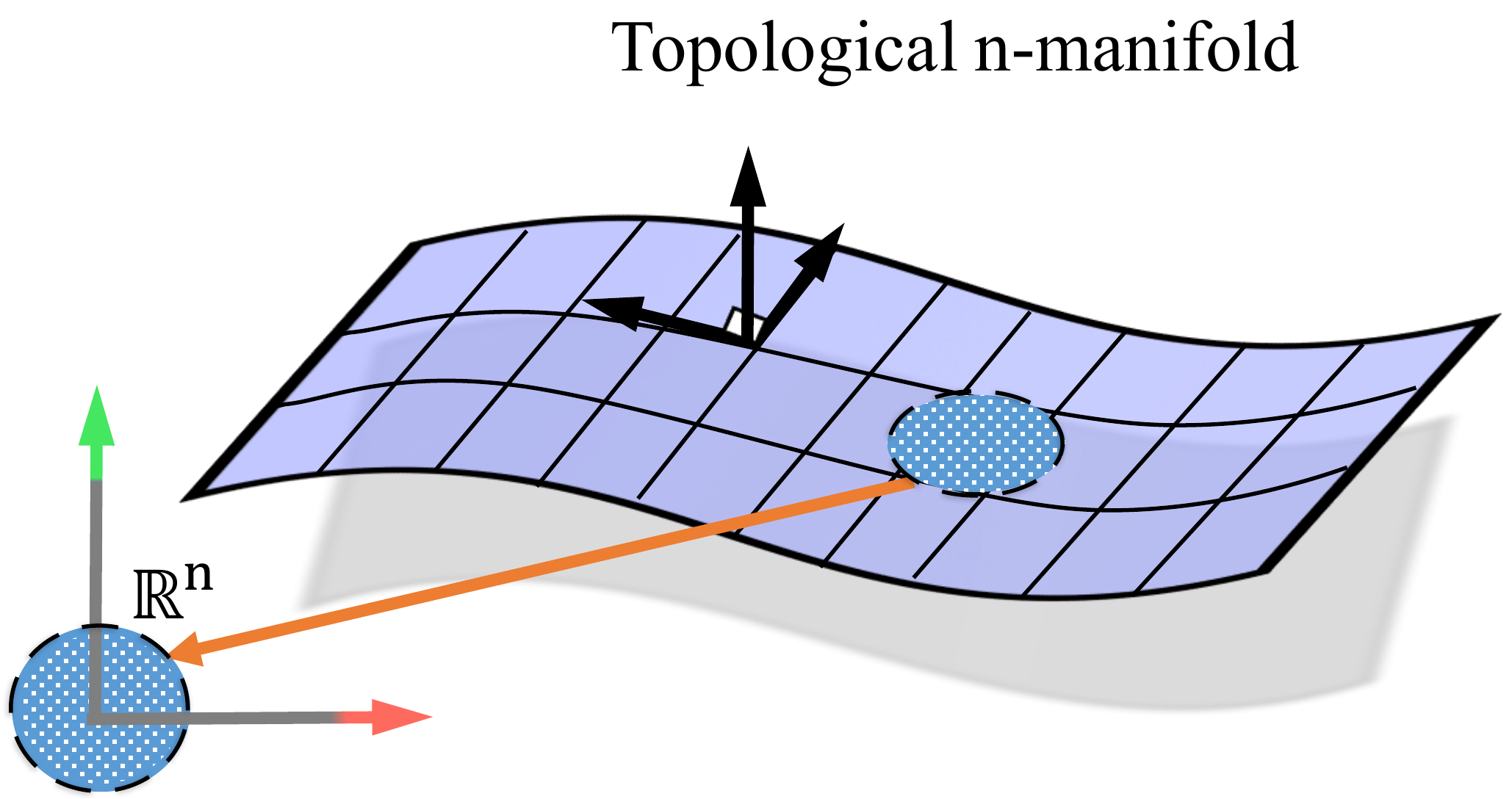}}
    \subfigure[\label{fig:nmanifold2}]{\includegraphics[width=0.5\linewidth]{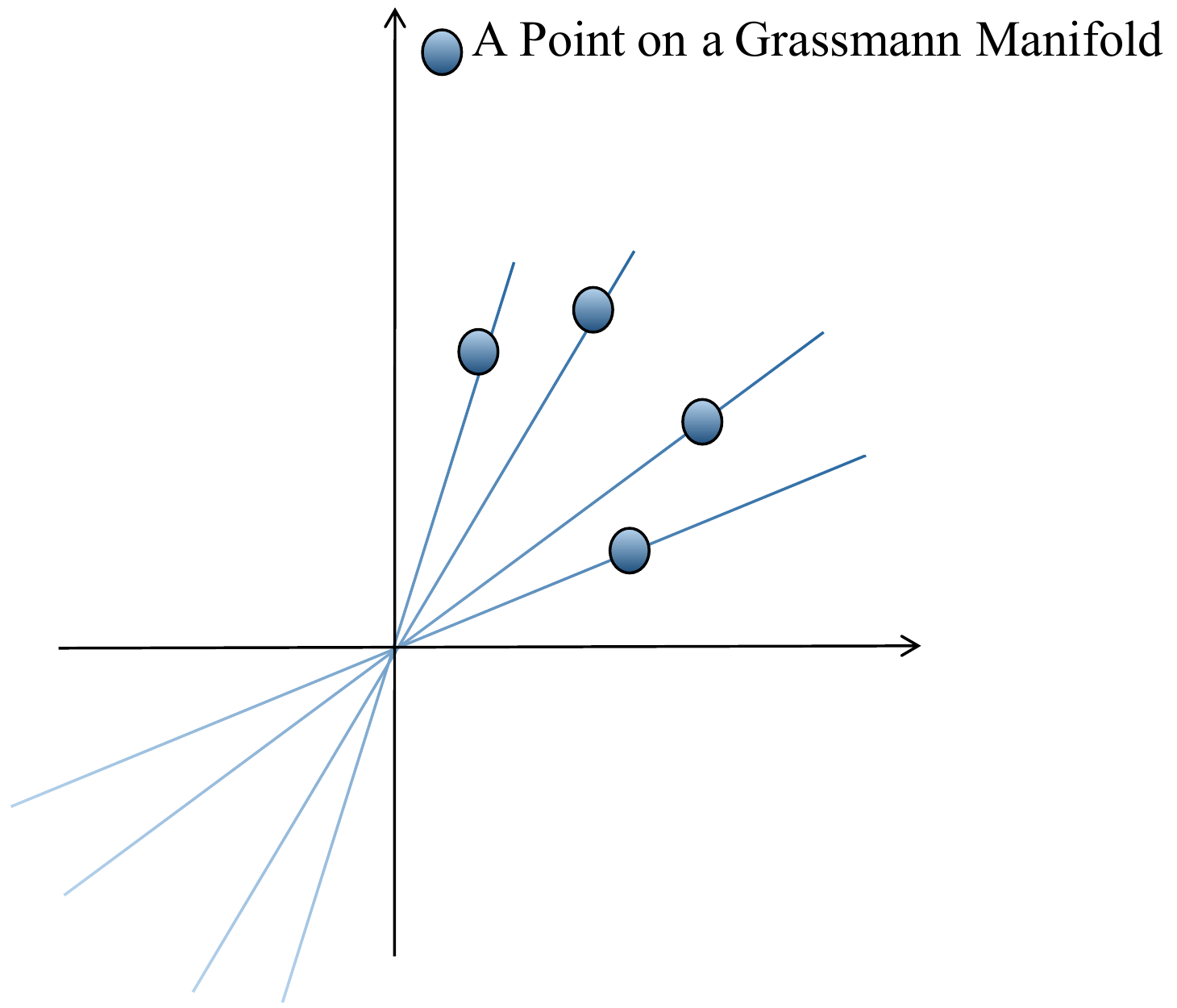}}
\caption{\small Grassmannian $\mathcal{G}$(1, 2). (a) Intuition of n-manifold. (b) 1-dimensional Euclidean subspace of $\mathbb{R}^2$ as a point on a Grassmann Manifold. Fig.(b) is inspired from \cite{hauberg2014grassmann} work.}
\label{fig:manifoldintuition}
\end{figure}

Recent state-of-the-art in sparse NRSfM  uses joint subspace clustering and reconstruction formulation \cite{kumar2016multi} \cite{kumar2017spatio}. Yet, the nature of the formulations fails to cope up with a large number of features points, and its inherent representation is unable to exploit the local surface deformation (spatially). However, the construction of simultaneous clustering and 3D reconstruction framework does provide an inspiration to extend such an idea to dense NRSfM. In this work, we want to take a step further and would like to show that it is, in fact, possible to develop the elementary idea proposed in \cite{kumar2017spatio} for dense NRSfM using a new representation and formulation.

\section{Preliminaries}
In this paper, $\| .\|_{\m F}$, $\| .\|_{*}$ denotes the Frobenius norm and nuclear norm respectively. $\| .\|_{\mathcal{G}}$ represents the notion of a norm on the Grassmann manifold. Single angle bracket $<.,.>$ denotes the Euclidean inner product. For ease of understanding and completeness, in this section, we briefly review a few important definitions related to the Grassmann manifold. Firstly, a topological n-manifold ($\mathcal{M}$) is a \emph{topological space} which is \emph{locally homeomorphic} to a \emph{n}-ball, where $n$ is a positive integer which is well-defined, which is the dimension of the manifold. Additionally, space ($\mathcal{M}$) is assumed to be Hausdorff and second countable. Avoiding the mathematical rigor, intuitively, one can think of a continuous surface to be locally similar to the Euclidean space (see Fig. \ref{fig:nmanifold1}). Out of several manifolds, the Grassmann manifold is a topologically rich non-linear manifold, each point of which represents the set of all right invariant subspace of the Euclidean space \cite{dollar2007non,absil2009optimization,kumar2018scalable} (see Fig. \ref{fig:nmanifold2}).
\begin{definition}\label{def:1}
   The Grassmann manifold, denoted by $\mathcal{G}(\m p, \m d)$, consists of all the linear $\m p$-dimensional subspace embedded in a `$\m d$' dimensional Euclidean space $\mathbb{R}^{\m d}$ such that $0\leq \m p \leq \m d$ \cite{absil2009optimization}.
\end{definition}
A point `$\m \Phi$' on the Grassmann manifold can be represented by $ \mathbb{R}^{\m d \times \m p}$ matrix whose columns are composed of orthonormal basis. The space of matrices with orthonormal columns is a Riemannian manifold such that $\m \Phi^{\m T} \m \Phi = \mathbf{I}_{\m p}$, where $\mathbf{I}_{\m p}$ is a $\m p \times \m p$ identity matrix. 

\begin{definition}\label{def:2}
	Grassmann manifold can be embedded into the space of symmetric matrices via mapping $\m \Pi$: $\mathcal{G}(\m p, \m d)$ $\mapsto$ $\textrm{Sym}$ $(\m d)$, $\m \Pi( \m \Phi) = \m \Phi \m \Phi^{\m T}$, where $\m \Phi$ is a Grassmann point \cite{hamm2008grassmann, harandi2013dictionary}.
   Given two Grassmann points $\m \Phi_1$ and $\m \Phi_2$, the distance between them can be measured using a projection metric:
\end{definition}
\begin{equation}
\begin{aligned}
& \displaystyle d_g^2(\m \Phi_1,  \m \Phi_2) = 0.5\|\m \Pi(\m \Phi_1) - \m \Pi(\m \Phi_2)\|_{\m F}^{2}.        
\end{aligned}
\end{equation}
These two properties of Grassmann manifold has been used in many computer vision applications \cite{hamm2008grassmann} \cite{cetingul2009intrinsic} \cite{kumar2018scalable} \cite{harandi2013dictionary}. Second definition is very important as it allows to measure the distance on the Grassmann manifold, hence, $(\mathcal{G}, d_g$) forms a metric space. We used these properties in the construction of our formulation. For comprehensive details on this topic readers may refer to \cite{hamm2008grassmann}.


\subsection{Why Grassmann Manifold?}
It is well-known that the complex non-rigid deformations are composed of multiple subspaces that quite often fit a higher-order parametric model \cite{pasko2002function} \cite{sheng2011facial} \cite{zhu2014complex}. To handle such complex models globally can be very challenging ---both numerically and computationally \cite{garg2013dense} \cite{garg2013variational}. Consequently, for an appropriate representation of such a model, we decompose the overall non-linearity of the shape by a set of locally linear models that span a low-rank subspace of a vector space. The space of all $\m d$-dimensional linear subspaces of $\mathbb{R}^{\m N}$ ($0 < \m d < \m N$) forms the Grassmann manifold \cite{absil2004riemannian} \cite{absil2009optimization}. Modeling the deformation on this manifold allows us to operate on the number of subspaces rather than on the number of vectorial data points (on the shape), which reduces the complexity of the problem significantly. Moreover, since each local surface is a low-rank subspace, it can be faithfully reconstructed using a few singular values and corresponding singular vectors, which makes such a representation scalable and robust to noise \cite{wang2015low}.



\section{Problem Formulation}

\subsection{Structure and Motion Representation} \label{ss:struct_motion}
Tomasi et.al. \cite{tomasi1992shape} matrix factorization method to represent the shape and motion under orthographic camera projection appropriately summarizes the behavior of the 3D points over frames. The relation between 3D shape, motion and its projection over frames was defined as
\begin{equation}\label{eq:classicalrep}
\small
\m W = \m R \m S
\end{equation}
where, $\m W \in \mathbb{R}^{2\m F \times \m P}$ is the measurement matrix with `$\m P$' as the total number of feature points tracked across `$\m F$' frames. $\m R = \textrm{{\fontfamily{cmtt}\selectfont blockdiagonal}}(\m R_1, \m R_2,..,\m R_{\m F}) \in \mathbb{R}^{2 \m F \times 3 \m F}$ denotes the orthographic camera rotation matrix with each $\m R_{\m i} \in \mathbb{R}^{2 \times 3}$ as per frame rotation. $\m S \in \mathbb{R}^{3 \m F \times \m P}$ represent the shape matrix with each row triplet as a 3D shape. This representation was originally formulated to solve \textbf{rigid} structure from motion under orthographic projection which was later extended by \cite{bregler2000recovering} to recover the 3D shape of a \textbf{non-rigidly} deforming object for multiple frames.

This classical representation entails that given the input measurement matrix, solve for rotation ($\m R$) and 3D shape ($\m S$). For our method, we solve for rotation matrix using the Intersection method \cite{dai2014simple} by assuming that per frame relative camera motion ($\m R$) can faithfully represent the global deformation of the subject in the scene. Accordingly, our goal reduces to develop a systematic approach that can reliably explain the non-rigid shape deformations and provides better 3D reconstruction. We used this relation to enforce our first constraint to solve for shape. This constraint is also known as a \textbf{re-projection error constraint} \textit{i.e,}
\begin{equation}
\small
\underset{\m S}  {\text{minimize}} ~\frac{1}{2}\|\m W - \m R \m S\|_{\m F}^2
\end{equation}

\subsection{Non-Rigid Object Representation}
As alluded to above, given the matrix $\m R$, our goal is to solve for the 3D structure $\m S \in \mathbb{R}^{3\m F \times \m P}$.  Eq:(\ref{eq:classicalrep}) show that we can get infinite family of solution to $\m S$ using such representation. Nonetheless, Bregler et.al. \cite{bregler2000recovering} $3\m K$ matrix factorization to obtain low-order linear model suggest that $\textrm{{\fontfamily{cmtt}\selectfont rank}}(\m S) \leq 3\m K$. Consequently, the non-rigid 3D shape must lie in a low-rank space. Later, Akther et.al. \cite{akhter2009nonrigid} and Dai et.al. \cite{dai2014simple} suggested the idea to provide stronger rank bound to the shape matrix by shuffling the arrangements of rows and columns of shape matrix \textit{i.e}, $\m S \in \mathbb{R}^{3\m F \times \m P} \mapsto \m S^\sharp \in \mathbb{R}^{3\m P \times \m F}$. Accordingly, we enforce the low-rank constraint on $\m S^\sharp$ which equivalently represent the low order constraint \cite{bregler2000recovering} \cite{akhter2009nonrigid} \cite{dai2014simple} with a tighter rank bound $\textrm{{\fontfamily{cmtt}\selectfont rank}}(\m S^{\sharp}) \leq \m K$. Combining re-projection error constraint with the low-rank constraint, we have 

\begin{equation}\label{eq:daiform}
\small
    \begin{aligned}
     & \displaystyle \underset{\m S, \m S^{\sharp}}  {\text{minimize}} ~\frac{1}{2}\|\m W - \m R \m S\|_{\m F}^2 + \gamma \|\m S^{\sharp}\|_*\\
     & \displaystyle \text{subject to:} ~\m S^{\sharp} = f(\m S)
    \end{aligned}
\end{equation}

\noindent
where $\|\m S^{\sharp}\|_*$ represent the nuclear norm of the shape matrix. Here, we define $f: \m S \in \mathbb{R}^{3 \m F \times \m P} \mapsto \m S^\sharp \in \mathbb{R}^{3\m P \times \m F}$. In general, the exact rank minimization problem is NP-hard, hence, we relax this with a nuclear norm minimization problem. Dai et.al. \cite{dai2014simple} proposed this formulation to solve non-rigid structure from motion problem. Although this formulation provides a decent result for sparse feature points, it fails to estimate dense non-rigid structure from motion with reasonable accuracy. One of the main reasons for its failure is that a  non-rigid deforming surface is mostly composed of a union of several local linear subspaces. Consequently, a global low-rank shape constraint fails to cater the local shape deformation, therefore, it provides questionable 3D reconstruction results for a dense deforming object.

To overcome this limitation with \cite{dai2014simple} formulation, joint  subspace clustering and reconstruction methods are proposed \cite{kumar2016multi} \cite{kumar2017spatio}. Although the method proposed by \cite{kumar2016multi} \cite{kumar2017spatio} provides state-of-the-art results for sparse features points \cite{jensen2018benchmark}, the algorithm cannot process a large set of feature points, hence, not scalable. To come up with an algorithm that is scalable and also utilize the idea of spatial-temporal clustering approach for dense non-rigid surfaces, we use grassmannian representation in our formulation \cite{absil2004riemannian} \cite{harandi2014manifold}.

\subsection{Grassmannian Representation in Trajectory Space}

Let `$\m \Phi_{\m s \m i}$' $\in \mathcal{G}(\m p, {\m d})$ be a Grassmann point representing the $\m i^{\textrm{th}}$ local linear subspace spanned by $\m i^{\textrm{th}}$ set of columns of `$\m S$'. Using this notion, we decompose the entire trajectories of the structure into a set of `$\m K_{\m s}$' Grassmannians $\xi_{\m s} = \{\m \Phi_{\m s \m 1}, \m \Phi_{\m s \m 2}, \m \Phi_{\m s \m 3}, ...., \m \Phi_{\m s \m K_{\m s} } \}$. To explain the complex non-rigid deformations, we reduce the overall non-linear space as a union of several local low-dimensional linear spaces which are sample points on the Grassmann manifold. But, the notion of self-expressiveness is valid only for Euclidean linear or affine subspace. To apply self-expressiveness on the Grassmann manifold one has to adopt linearity onto the manifold. Since, Grassmann manifold is isometrically equivalent to the symmetric idempotent matrices \cite{chikuse2012statistics}, we embed the Grassmann manifold into the symmetric matrix manifold, where the self-expressiveness can be defined in the embedding space. Let `$\rchi_{\m s}$' be a tensor which is constructed by mapping trajectory space Grassmann points. Concretely, $\rchi_{\m s} = \{ (\m \Phi_{\m s \m 1} \m \Phi_{\m s \m 1}^{\m T}), (\m \Phi_{\m s \m 2} \m \Phi_{\m s \m 2}^{\m T}), ..., (\m \Phi_{{\m K \m s}} \m \Phi_{{\m K \m s}}^{\m T}) \}$ is its embedding onto symmetric matrix manifold which is constructed by mapping trajectory space Grassmann points. Since the high-dimensional complex deformation is composed of several low-dimensional subspace, its low rank representation shall reveal the subspace information. This motivation leads to the following optimization in the trajectory space

\begin{equation}\label{eq:GRTS}
\begin{small}
\begin{aligned}
& \displaystyle \underset{\m E_{\m s}, \m C_{\m s}}  {\text{minimize}} ~\|\m E_{\m s}\|_{\m F}^2 + \lambda_1\|\m C_{\m s}\|_* \\
& \displaystyle \text{subject to:} ~\rchi_{\m s} = \rchi_{\m s}\m C_{\m s} + \m E_{\m s}
\end{aligned}
\end{small}
\end{equation}

We denote $\m C_{\m s} \in \mathbb{R}^{\m K_{\m s} \times \m K_{\m s}}$ as the coefficient matrix with `$\m K_{\m s}$' as the total number of spatial groups. Here, $\m E_{\m s}$ measures the trajectory subspace reconstruction error as per the manifold geometry. Also, we would like to emphasize that since the object undergoes deformations in the 3D space, we operate in 3D space rather than in the projected 2D space. $\|~\|_*$ is enforced on $\m C_{\m s}$  for a low-rank solution.

\subsection{Grassmannian Representation in Shape Space}
Similarly, let `$\m \Phi_{\m t \m i}$' $\in \mathcal{G}(\m p, {\m d})$ be a Grassmann point representing the $\m i^{\textrm{th}}$ local linear subspace spanned by $\m i^{\textrm{th}}$ set of columns of `$\m S^\sharp$'. We decompose the set of shapes into `$\m K_{\m t}$' Grassmannians $\xi_{\m t} = \{\m \Phi_{\m t \m 1}, \m \Phi_{\m t \m 2}, \m \Phi_{\m t \m 3}, ...., \m \Phi_{\m t \m K_{\m t} } \}$. To accomplish the notion of self-expressiveness in the temporal space as well, we define $\rchi_{\m t} = \{ (\m \Phi_{\m t \m 1} \m \Phi_{\m t \m 1}^{\m T}), (\m \Phi_{\m t \m 2} \m \Phi_{\m t \m 2}^{\m T}), ..., (\m \Phi_{{\m K \m t}} \m \Phi_{{\m K \m t}}^{\m T}) \}$. Previous literature and experiments revealed that deforming object attains different state over time which adheres to distinct temporal local linear subspaces \cite{kumar2017spatio}. Assuming that the temporal deformation is smooth over-time, we express deforming shapes in terms of local self-expressiveness on grassmann manifold across frames as:
\begin{equation}\label{eq:GRSS}
\begin{small}
\begin{aligned}
& \displaystyle \underset{\m E_{\m t}, \m C_{\m t}}  {\text{minimize}} ~\|\m E_{\m t}\|_{\m F}^2 + \lambda_2\|\m C_{\m t}\|_* \\
& \displaystyle \text{subject to:} ~\rchi_{\m t} = \rchi_{\m t} \m C_{\m t} + \m E_{\m t}
\end{aligned}
\end{small}
\end{equation}
where, $\rchi_{\m t}$ is the set of all symmetric matrices constructed using a set of Grassmannian samples $\xi_{\m t}$, where $\xi_{\m t}$ contains the samples which are obtained from $\m S_{\m t}^{\sharp} \in \mathbb{R}^{3 \m P \times \m F}$. Intuitively, $\m S_{\m t}^{\sharp}$ is a shape matrix with each column as a deforming shape. $\m E_{\m t}$, $\m C_{\m t} \in \mathbb{R}^{\m K_{\m t} \times \m K_{\m t}}$ represent the temporal group reconstruction error and coefficient matrix respectively, with $\m K_{\m t}$ as the number of temporal groups. $\|~\|_*$ is enforced on $\m C_{\m t}$  for a low-rank solution.

\subsection{Simplified Low Rank Subspace Representation on the Grassmann Manifold}
The grassmannian representation in Eq.(\ref{eq:GRTS}) and Eq.(\ref{eq:GRSS}) are not straight-forward to solve, we simplified it further to an equivalent optimization problem which is easy to optimize. Consider the following minimization problem

\begin{equation}\label{eq:srefeq}
\begin{small}
\begin{aligned}
& \displaystyle \underset{\m C}  {\text{minimize}} ~\|\rchi - \rchi \m C\|_{\m F}^2 + \lambda\|\m C\|_*
\end{aligned}
\end{small}
\end{equation}

\noindent
Here, we choose the notations that stands for both Eq.(\ref{eq:GRTS}) and Eq.(\ref{eq:GRSS}). To simplify the form of previous optimization problem, let's consider the error term that involves the tensor structure.
\begin{equation}
\small
\begin{aligned}
& \displaystyle \|\m E\|_{\m F}^2 = \| \rchi - \rchi \m C \|_{\m F}^2
\end{aligned}
\end{equation}

\noindent
Using our notation $\rchi = \{ (\m \Phi_{\m 1} \m \Phi_{\m 1}^{\m T}), (\m \Phi_{ \m 2} \m \Phi_{\m 2}^{\m T}), ..., (\m \Phi_{{\m N}} \m \Phi_{{\m N}}^{\m T}) \}$ and $\m C \in \mathbb{R}^{\m N \times \m N}$ are the set of grassmann samples in the embedding space and its coefficient matrix respectively. Let's re-write the previous equation
\begin{equation}\label{eq:errori}
\small
    \begin{aligned}
     & \displaystyle \|\m E\|_{\m F}^2 = \sum_{\m i = 1}^{\m N} \|\m E_{\m i}\|_{\m F}^{2} = \sum_{\m i = 1}^{\m N} \textbf{Tr}(\m E_{\m i}^{\m T}\m E_{\m i})
    \end{aligned}
\end{equation}

\noindent
Using the per sample notion \textit{i.e,} any sample can be represented as a combination of other samples in the same space.
\begin{equation}
\small
\begin{aligned}
& \displaystyle \m E_{\m i} =  \m \Phi_{\m i} \m \Phi_{\m i}^{\m T} - \sum_{\m j = 1}^{\m N} \m c_{\m i \m j} (\m \Phi_{\m j} \m \Phi_{\m j}^{\m T})
\end{aligned}
\end{equation}

\noindent
Substituting the above expression in Eq.(\ref{eq:errori}) for the $\m i^\text{th}$ sample, we write
\begin{equation}
\small
    \begin{aligned}
    & \displaystyle  \| \m E_{\m i} \|_{\m F}^2 = \textbf{Tr}(\m E_{\m i}^{\m T}\m E_{\m i}) \\ 
    & \displaystyle = \textbf{Tr}\Big[ \Big(\m \Phi_{\m i} \m \Phi_{\m i}^{\m T} - \sum_{\m j = 1}^{\m N} \m c_{\m i \m j} (\m \Phi_{\m j} \m \Phi_{\m j}^{\m T})  \Big)^{\m T} \Big(\m \Phi_{\m i} \m \Phi_{\m i}^{\m T} - \sum_{\m j = 1}^{\m N} \m c_{\m i \m j} (\m \Phi_{\m j} \m \Phi_{\m j}^{\m T}) \Big) \Big]
    \end{aligned}
\end{equation}
\noindent
Expanding the above form
\begin{equation}
\small
\begin{aligned}
& \displaystyle \|\m E_{\m i}\|_{\m F}^2 =  
\textbf{Tr}\big( (\m \Phi_{\m i}\m \Phi_{\m i}^{\m T})^{\m T}  (\m \Phi_{\m i}\m \Phi_{\m i}^{\m T})\big) -2 \sum_{\m j=1}^{\m N} \m c_{\m i\m j} \textbf{Tr} \big( (\m \Phi_{\m i}\m \Phi_{\m i}^{\m T})^{\m T}  (\m \Phi_{\m j}\m \Phi_{\m j}^{\m T})\big) \\
& \displaystyle + \sum_{\m l=1}^{\m N} \sum_{\m m=1}^{\m N} \m c_{\m i \m l} \m c_{\m i \m m} \textbf{Tr}  \big( (\m \Phi_{\m l}\m \Phi_{\m l}^{\m T})^{\m T}  (\m \Phi_{\m m}\m \Phi_{\m m}^{\m T})\big)
\end{aligned}
\end{equation}
\noindent
Using the cyclic trace property and the orthonormality property of matrices (\textbf{Definition} \ref{def:1}).

\begin{equation}
\small
    \begin{aligned}
    & \displaystyle \|\m E_{\m i}\|_{\m F}^2 = \textbf{Tr}(\mathbf{I}_{\m p}) - 2 \sum_{\m j=1}^{\m N} \m c_{\m i\m j} \textbf{Tr} \big( (\m \Phi_{\m j}^{\m T} \m \Phi_{\m i})  (\m \Phi_{\m i}^{\m T} \m \Phi_{\m j})\big) \\
    & \displaystyle + \sum_{\m l=1}^{\m N} \sum_{\m m=1}^{\m N} \m c_{\m i \m l} \m c_{\m i \m m} \textbf{Tr}  \big( (\m \Phi_{\m l} ^{\m T} \m \Phi_{\m m})  (\m \Phi_{\m m}^{\m T} \m \Phi_{\m m})\big)
    \end{aligned}
\end{equation}
\noindent
Here $\m p$ is the magnitude of the lower dimensional space representation (\textbf{Definition} \ref{def:1}). By letting  $\m \Gamma_\textrm{ij} = \textbf{Tr} \big( (\m \Phi_{\m j}^{\m T} \m \Phi_{\m i})  (\m \Phi_{\m i}^{\m T} \m \Phi_{\m j})\big)$, we can rewrite the above form as

\begin{equation}\label{eq:trij}
\small
    \begin{aligned}
    & \displaystyle \|\m E_{\m i}\|_{\m F}^2 = \textbf{Tr}(\mathbf{I}_{\m p}) - 2 \sum_{\m j=1}^{\m N} \m c_{\m i\m j} \m \Gamma_\textrm{ij} + \sum_{\m l=1}^{\m N} \sum_{\m m=1}^{\m N} \m c_{\m i \m l} \m c_{\m i \m m} \m \Gamma_\textrm{lm}
    \end{aligned}
\end{equation}

\noindent Notice that $\m \Gamma_\textrm{ij}$ is $\mathbb{R}^{\m p \times \m p}$ matrix which is much easier to compute and process. Also, it's simple to verify that $\m \Gamma_\textrm{ij}$ is symmetric. Let $\m \Gamma = (\m \Gamma_\textrm{ij})_{\textrm{ij}=1}^{\m N} \in \mathbb{R}^{\m N \times \m N}$. By summing over all the samples , we can rewrite Eq:(\ref{eq:errori}) as:

\begin{equation}\label{eq:finallform}
\small
    \begin{aligned}
        & \displaystyle \|\m E\|_{\m F}^2 = \m N \m p - 2 \textbf{Tr}(\m C \m \Gamma) + \textbf{Tr}(\m C \m \Gamma \m C^{\m T}) \\
        & \displaystyle ~~~~~~~~\equiv \m N \m p - 2 \textbf{Tr}(\m C \m L \m L^{\m T}) + \textbf{Tr}(\m C \m L \m L^{\m T} \m C^{\m T}) \\
        & \displaystyle ~~~~~~~~\equiv \textrm{const} + \|\m L -  \m C \m L\|_{\m F}^2
    \end{aligned}
\end{equation}
\noindent
$\textrm{where}, \m L \m L^{\m T} = \textbf{Chol}(\m \Gamma)$, the Cholesky decomposition of the matrix. Note that adding and subtracting constant symbol w.r.t variable $\m C$ will not affect the solution to the targeted optimization problem. Using this form, we simplify the Eq:(\ref{eq:srefeq}) optimization problems as
\begin{equation}\label{eq:LRRGrass}
    \small
    \begin{aligned}
    & \displaystyle \underset{\m C}  {\text{minimize}} ~\|\m L - \m C \m L \|_{\m F}^2 + \lambda\|\m C\|_*
    \end{aligned}
\end{equation}

\noindent
This simplified equivalent problem is much easier to solve and process. We will use this form in our overall cost function to solve non-rigid 3D shape reconstruction.

\section{Spatial Temporal Formulation}
Combining the above developed objectives and their constraints give us our spatial temporal formulation for dense NRSfM. Our representation blends the local subspaces structure along with the global composition of a non-rigid shape. Thus, the overall objective is:

\begin{equation} \label{eq:spatialtempoverallOptimization} 
\begin{small}
\begin{aligned}
& \displaystyle \underset{\m S, \m S^{\sharp}, \m E_{\m s}, \m E_{\m t}, \m C_{\m s}, \m C_{\m t}} {\text{minimize}} ~\mathbf{E} = \frac{1}{2} ~\|\m W- \m R \m S\|_{\m F}^2 + \mu \| \m S^{\sharp} \|_* +  \lambda_1\|\m E_{\m s} \|_{\m F}^2 + \lambda_2\|\m E_{\m t}\|_{\m F}^{2} \\
& \displaystyle  + \lambda_3 \|\m C_{\m s}\|_* +  \lambda_4 \|\m C_{\m t}\|_*\\
& \displaystyle \text{subject to :} \\
& \displaystyle ~\rchi_{\m s} = \rchi_{\m s} \m C_{\m s} + \m E_{\m s}; ~\rchi_{\m t} = \rchi_{\m t} \m C_{\m t} + \m E_{\m t};\\
& \displaystyle \xi_{\m s}  = {f_g( \mathbf{P}_{\m s}, \m S, \m K_{\m s}, \m p_{\m s})}; ~\xi_{\m t}  = {f_g(\mathbf{P}_{\m t}, \m S^{\sharp}, \m K_{\m t}, \m p_{\m t})}; \\
& \displaystyle  \m S = f_{s}(\xi_{\m s}, \m \Sigma_{\m s}, \xi_{\m v \m s}, \m K_{\m s}, \m p_{\m s}); ~ \m S^{\sharp} = f_{s}(\xi_{\m t}, \m \Sigma_{\m t}, \xi_{\m v \m t}, \m K_{\m t}, \m p_{\m t}); \\
& \displaystyle  \mathbf{P}_{\m s} = f_p(\xi_{\m s}, \m C_{\m s}, \mathbf{P}_{\m s\m o}); ~\mathbf{P}_{\m t} = f_p(\xi_{\m t}, \m C_{\m t}, \mathbf{P}_{\m t\m o}); \\
& \displaystyle \m S^{\sharp} = f(\m S);  ~\m W = {f}_o(\m W, \mathbf{P}_{\m s}); 
\end{aligned}
\end{small}
\end{equation}

\noindent
We introduce few  constraint functions that provides a way to group Grassmannians and recover 3D shape simultaneously. Let $\mathbf{P}_{\m s}$ $\in \mathbb{R}^{1 \times \m P}$, $\mathbf{P}_{\m t}$ $\in \mathbb{R}^{1 \times \m F}$ be an ordering vector that contains the index of columns of $\m S$ and $\m S^\sharp$ respectively. Our function definition is of the form $\{(\textrm{output}, \textrm{function(.)}): \textrm{definition} \}$. Using it, we define the function $f_g$, $f_{s}$, $f_p$, $f$, $f_o$ as follows:
\begin{equation}\label{eq:fg}
\small
\begin{aligned}
& \displaystyle \Big\{ \big({\xi}, f_g(\mathbf{P}, \m X, \m K, \m p)\big): \textrm{order} ~\{\m X_{\m i}\}_{\m i=1}^{\m K} ~\text{columns of} ~\m X ~\text{using} ~{\mathbf{P} }, \\
& \displaystyle ~\xi := \big(\m \Phi_{\m i}\big|_{\m i=1}^{\m K} \big), ~\text{where}, [\m \Phi_{\m i}, \m \Sigma_{\m i}, \xi_\textrm{vi}^{\m T} ] = \textrm{svds}(\m X_{\m i}, \m p)\Big\}
\end{aligned}
\end{equation}

\begin{equation}\label{eq:fs}
\small
\begin{aligned}
& \displaystyle \Big\{\big(\m X, f_s({\xi}, \m \Sigma, {\xi}_{\m v}, \m K, \m p)\big): \m X_{\m i} = [{\xi}_{\m i}~{\m \Sigma_{\m i}}~{\xi}_{\m v\m i}^{\m T}], \text{where} ~{\m \Sigma_{\m i}} \in \mathbb{R}^{\m p \times \m p}, \\
& \displaystyle \m X =  \big( \m X_{\m i}\big|_{\m i = 1}^{\m K} \big) \Big\}
\end{aligned}
\end{equation}
\begin{equation}\label{eq:fp}
\small
\begin{aligned}
& \displaystyle \Big\{(\mathbf{P}, f_{p}({\xi}, {\m C}, \mathbf{P}_{\m o}): \mathbf{P} = \text{spectral\_clustering}({\xi}, {\m C}, \mathbf{P}_{\m o})\Big\} 
\end{aligned}
\end{equation}
\begin{equation}\label{eq:f}
    \small
    \begin{aligned}
    & \displaystyle \Big\{(\m X^{\sharp}, f(\m X) : \m X \in \mathbb{R}^{3 \m F \times \m P} \mapsto \m X^{\sharp} \in \mathbb{R}^{3 \m P \times \m F}\Big\}
    \end{aligned}
\end{equation}
\begin{equation}
    \small
    \begin{aligned}\label{eq:fo}
    & \displaystyle \Big\{(\m X, f_o(\m X, \mathbf{P})) : \m X = \text{arrange columns of $\m X$ using ordering vector $\mathbf{P}$}  \Big\}
    \end{aligned}
\end{equation}

\noindent
Note: $\big(\m X_{\m i} \big|_{\m i}^{\m K} \big)$ denoted the horizontal concatenation of the matrices. Intuitively, $f_g(.)$ provides the Grassmannian representation and  $f_s(.)$ reconstructs back each local low-rank subspace. $f_p(.)$ provides the ordering vector based on the inference drawn from coefficient matrix and $f_o$ rearranges the columns of $\m W$ in accordance with the columns of the shape matrix. The proposed cost function is minimized by solving for one variable at a time while treating others as constant, keeping the constraints intact over iteration. Next, we provide a detailed derivation to each sub-problem. 

\subsection{Solution}

The formulation in Eq.(\ref{eq:spatialtempoverallOptimization}) is a non-convex problem due to the bilinear optimization variables ($\rchi_{\m s} \m C_{\m s}$, $\rchi_{\m t} \m C_{\m t}$), hence a global optimal solution is hard to achieve. However, it can be efficiently solved using Augmented Lagrangian Methods (ALMs) \cite{boyd2011distributed}, which has proven its effectiveness for many non-convex problems. Using the result of Eq.(\ref{eq:LRRGrass}) with introduction of Lagrange multipliers $(\{\mathbf{L}_{\m i}\}_{\m i=1}^3)$ and auxiliary variables $(\m J_{\m s}, \m J_{\m t})$ to Eq. (\ref{eq:spatialtempoverallOptimization}) gives us the overall cost function as follows:

\begin{equation} \label{eq:spatialtempOptimizationLM} 
\begin{small}
\begin{aligned}
& \displaystyle \underset{\m S, \m S^{\sharp}, \m J_{\m s}, \m J_{\m t}, \m C_{\m s}, \m C_{\m t}} {\text{minimize}} ~\mathbf{E} = \frac{1}{2} ~\|\m W- \m R \m S\|_{\m F}^2 + \frac{\beta}{2}\| \m S^{\sharp} - f(\m S)\|_{F}^2 + <\mathbf{L}_1, \m S^{\sharp} - f(\m S)> \\
& \displaystyle \gamma \| \m S^{\sharp} \|_* +  \lambda_1\|\m L_{\m s} - \m C_{\m s} \m L_{\m s}  \|_{\m F}^2 +  \lambda_3\|\m J_{\m s}\|_* + \lambda_2\|\m L_{\m t} - \m C_{\m t} \m L_{\m t} \|_{\m F}^{2} + \lambda_4\|\m J_{\m t}\|_* + \\
& \displaystyle  \frac{\beta}{2} \|\m C_{\m s}-\m J_{\m s}\|_{\m F}^2 + <\mathbf{L}_2, \m C_{\m s}-\m J_{\m s}>  + \frac{\beta}{2} \|\m C_{\m t}-\m J_{\m t}\|_{\m F}^2 + <\mathbf{L}_3, \m C_{\m t}-\m J_{\m t}>\\
& \displaystyle \text{subject to :} \\
& \displaystyle \xi_{\m s}  = {f_g( \mathbf{P}_{\m s}, \m S, \m K_{\m s}, \m p_{\m s})}; ~\xi_{\m t}  = {f_g(\mathbf{P}_{\m t}, \m S^{\sharp}, \m K_{\m t}, \m p_{\m t})}; \\
& \displaystyle  \m S = f_{s}(\xi_{\m s}, \m \Sigma_{\m s}, \xi_{\m v \m s}, \m p_{\m s}, \m K_{\m s}); ~ \m S^{\sharp} = f_{s}(\xi_{\m t}, \m \Sigma_{\m t}, \xi_{\m v \m t}, \m p_{\m t}, \m K_{\m t}); \\
& \displaystyle  \mathbf{P}_{\m s} = f_p(\xi_{\m s}, \m C_{\m s}, \mathbf{P}_{\m s\m o}); ~\mathbf{P}_{\m t} = f_p(\xi_{\m t}, \m C_{\m t}, \mathbf{P}_{\m t\m o}); \\
& \displaystyle \m S^{\sharp} = f(\m S);  ~\m W = {f}_o(\m W, \mathbf{P}_{\m s}); 
\end{aligned}
\end{small}
\end{equation}

\noindent
\textbf{Solution to $\m S$}
\begin{equation}\label{eq:ss}
\begin{small}
\begin{aligned}
& \displaystyle \underset{\m S} {\text{argmin}} \frac{1}{2} ~\|\m W-\m R\m S\|_{\m F}^2 + \frac{\beta}{2}\| \m S^{\sharp} - f(\m S)\|_{\m F}^2 + <\mathbf{L}_1, \m S^{\sharp} - f(\m S)> \\
& \displaystyle \underset{\m S} {\text{argmin}} \frac{1}{2} ~\|\m W-\m R \m S\|_{\m F}^2 + \frac{\beta}{2}\| f^{-1}(\m S^{\sharp}) - \m S\|_{\m F}^2 + 
<f^{-1}(\mathbf{L}_1), f^{-1}(\m S^{\sharp})-\m S> \\
& \displaystyle \equiv \underset{\m S} {\text{argmin}}  \frac{1}{2} ~\|\m W-\m R\m S\|_{\m F}^2 + \frac{\beta}{2}\|\m S - \big(f^{-1}(\m S^{\sharp}) + \frac{f^{-1}(\mathbf{L}_1)}{\beta}\big)\|_{\m F}^2.
\end{aligned}
\end{small}
\end{equation}

\noindent
The solution to the variable `$\m S$' can be derived by differentiating the above term w.r.t $\m S$ and equating it to zero.
\begin{equation}
\begin{small}
\begin{aligned}
 \m S \equiv \Big(\m R^{\m T}\m R + \beta \m I\Big)^{-1} \Big(\beta\big(f^{-1}(\m S^{\sharp}) + \frac{f^{-1}(\mathbf{L}_1)}{\beta}\big) + \m R^{\m T} \m W \Big)
\end{aligned}
\end{small}
\end{equation}

\noindent
\textbf{Solution to $\m S^\sharp$}
\begin{equation}
\begin{small}
\begin{aligned}
& \displaystyle \equiv \underset{\m S^{\sharp}} {\text{argmin}} ~\gamma \| \m S^{\sharp} \|_*  + \frac{\beta}{2}\| S^{\sharp} - f(\m S)\|_{\m F}^2 + 
<\mathbf{L}_1, \m S^{\sharp} - f(\m S)> \\
& \displaystyle \equiv  \underset{\m S^{\sharp}} {\text{argmin}} ~\gamma\| \m S^{\sharp} \|_*  + \frac{\beta}{2}\|\m S^{\sharp} - \big( f(\m S) - \frac{\mathbf{L}_1}{\beta}\big)\|_F^2
\end{aligned}
\end{small}
\end{equation}

\noindent
Let's define the soft-thresholding operation as $\mathcal{S}_{\tau}[x] = \mathrm{sign}(x)$ $\max(|x|-\tau, 0)$. The optimal solution to $\m S^{\sharp}$ is given by

\begin{equation} \label{eq:St}
\begin{small}
\begin{aligned}
& \displaystyle \m S^{\sharp} \equiv \m U \mathcal{S}_{\frac{\gamma}{\beta}}(\m \Sigma) \m V^{\m T} 
~\text{where, } [\m U, \m \Sigma, \m V^{\m T}] = \text{svd}(f(\m S) - \frac{\mathbf{L}_1}{\beta})
\end{aligned}
\end{small}
\end{equation}

\noindent
\textbf{Solution to $\m C_{\m s}$}
\begin{equation}
\begin{small}
\begin{aligned}
& \displaystyle \equiv \underset{\m C_{\m s}} {\text{argmin}} ~\lambda_1\|\m L_{\m s}- \m C_{\m s}\m L_{\m s}\|_{\m F}^2 + \frac{\beta}{2}\|\m C_{\m s} - \m J_{\m s}\|_{\m F}^2 + <\mathbf{L}_2, \m C_{\m s} - \m J_{\m s}> \\
& \displaystyle \equiv \underset{\m C_{\m s}} {\text{argmin}} ~\lambda_1\|\m L_{\m s}-\m C_{\m s}\m L_{\m s}\|_{\m F}^2 + \frac{\beta}{2} \|\m C_{\m s} - \big(\m J_{\m s}-\frac{\mathbf{L}_2}{\beta}\big)\|_{\m F}^2
\end{aligned}
\end{small}
\end{equation}

\noindent
The solution to $\m C_{\m s}$ can be derived by differentiating the above term w.r.t $\m C_{\m s}$ and equating it to zero.

\begin{equation}
\begin{small}
\begin{aligned}
\m C_{\m s} \equiv  \Big(2\lambda_1 \m L_{\m s} \m L_{\m s}^{\m T} + \beta(\m J_{\m s} -\frac{\mathbf{L}_2}{\beta}) \Big)  \Big(2 \lambda_1 \m L_{\m s} \m L_{\m s}^{\m T} + \beta \m I_{\m s}  \Big)^{-1}
\end{aligned}
\end{small}
\end{equation}

\noindent
\textbf{Solution to $\m C_{\m t}$} 

Similar to the $\m C_{\m s}$ derivation, it's solution can be derived as follows:
\begin{equation}
\begin{small}
\begin{aligned}
& \displaystyle \equiv \underset{\m C_{\m t}} {\text{argmin}} ~\lambda_2\|\m L_{\m t}-\m C_{\m t} \m L_{\m t}\|_{\m F}^2 + \frac{\beta}{2}\|\m C_{\m t} - \m J_{\m t}\|_{\m F}^2 + <\mathbf{L}_3, \m C_{\m t} - \m J_{\m t}> \\
& \displaystyle \equiv \underset{\m C_{\m t}} {\text{argmin}} ~\lambda_2\|\m L_{\m t}-\m C_{\m t} \m L_{\m t}\|_{\m F}^2 + \frac{\beta}{2} \|\m C_{\m t} - \big(\m J_{\m t}-\frac{\mathbf{L}_3}{\beta}\big)\|_F^2
\end{aligned}
\end{small}
\end{equation}

\begin{equation}
\begin{small}
\begin{aligned}
\m C_{\m t} \equiv   \Big(2\lambda_2 \m L_{\m t} \m L_{\m t}^{\m T} + \beta(\m J_{\m t} -\frac{\mathbf{L}_{3}}{\beta}) \Big)  \Big(2 \lambda_2 \m L_{\m t} \m L_{\m t}^{\m T} + \beta \m I_{\m t}  \Big)^{-1}
\end{aligned}
\end{small}
\end{equation}

\begin{algorithm*}[t!]
\caption{Dense Non-Rigid Structure from Motion using Grassmannians}\label{alg:Algorithm1}
\begin{algorithmic}[1]
\REQUIRE
$\m W$, $\m R$ using \cite{dai2014simple}, tuning parameters: $\lambda_1$, $\lambda_2$, $\lambda_3$, $\lambda_4$, $\gamma$, $\rho=1.1$, $\beta=1e^{-3}$, $\beta_m=1e^{6}$, $\epsilon = 1e^{-12}$, $\m K_{\m s}$, $\m K_{\m t}$.\\
\hspace{-0.3cm}{\bf Initialize:} $\m S$ = \textbf{pinv}($\m R$)$\m W$ and ${ \m S^{\sharp}} = f(\m S)$.\\

\hspace{-0.3cm}{\bf Initialize:} `$\m K_{\m t}$' temporal data points on the Grassmann manifold using  $\mathbf{P}_{\m t \m o}$ = $\textrm{kmeans++}(\m S^\sharp, \m K_{\m t})$ index to `${\m S^{\sharp}}$' matrix, 
$\xi_{\m t}$ = $\{\m \Phi_{\m t \m i}\}_{\m i=1}^{\m K_{\m t}}$\\
\hspace{-0.3cm}{\bf Initialize:} `$\m K_{\m s}$' spatial data points on the Grassmann manifold using $\mathbf{P}_{\m s \m o}$ = $\textrm{kmeans++}(\m S, \m K_{\m s})$ index to `${\m S}$' matrix, $\xi_{\m s}$ = $\{\m \Phi_{\m s \m i}\}_{\m i=1}^{\m K_{\m s}}$\\
\hspace{-0.3cm}{\bf Initialize:} The auxiliary variables $\m J_{\m s}$, $\m J_{\m t}$ and Lagrange multiplier $\{\mathbf{L}_{\m i}\}_{\m i=1}^{3}$ as zero matrices.\\
\hspace{-0.3cm}{\bf Initialize:} $\m \Gamma_\textrm{ij}^{\m s} = \textbf{Tr}[({\m \Phi_{\m s\m j}^{\m T}} {\m \Phi_{\m s\m i}} ) ({\m \Phi_{\m s\m i}^{\m T}} {\m \Phi_{\m s\m j}})]$,
$\m \Gamma_\textrm{ij}^{\m t} = \textbf{Tr}[({\m \Phi_{\m t\m j}^{\m T}} {\m \Phi_{\m t\m i}} ) ({\m \Phi_{\m t\m i}^{\m T}} {\m \Phi_{\m t\m j}})]$, ${\m \Gamma}_{\m s} = (\m \Gamma_\textrm{ij}^{\m s})_{\m i, \m j = 1}^{\m K_{\m s}}, {\m \Gamma}_{\m t} = (\m \Gamma_\textrm{ij}^{\m t})_{\m i, \m j = 1}^{\m K_{\m t}}$\\

\hspace{0.95cm} $\m L_{\m s} \m L_{\m s}^{\m T}$ = \textbf{Chol}(${\m \Gamma}_{\m s}$), $\m L_{\m t}\m L_{\m t}^{\m T}$ = \textbf{Chol}(${\m \Gamma}_{\m t}$)\\
\hspace{-0.3cm}{\bf Initialize:} $\mathbf{P}_{\m s} := \mathbf{P}_{\m s \m o}$, $\mathbf{P}_{\m t} := \mathbf{P}_{\m t \m o}$, $\textrm{iter} = 1$ \\
\hspace{-0.3cm}{\bf Define:} ~~~~$\mathcal{S}_{\tau}(\m x) := \textcolor{mygreen}{@}(\m x, \tau) \textrm{sign}(\m x).*\textrm{max}(\textrm{abs}(\m x)-\tau, 0);$ \hspace{1.5cm} \COMMENT {MATLAB function script}\\
\vspace{0.2cm}
\WHILE {not converged}
\STATE $\m S$ $\gets$ $\Big(\m R^{\m T}\m R + \beta \m I\Big)^{-1} \Big(\beta\big(f^{-1}(\m S^{\sharp}) + \frac{f^{-1}(\mathbf{L}_1)}{\beta}\big) + \m R^{\m T} \m W \Big)$  \\

\STATE $\m C_{\m s}$ $\gets$ $ \Big(2\lambda_1 \m L_{\m s} \m L_{\m s}^{\m T} + \beta(\m J_{\m s} -\frac{\mathbf{L}_2}{\beta}) \Big) \Big(2 \lambda_1 \m L_{\m s} \m L_{\m s}^{\m T} + \beta \m I_{\m s}  \Big)^{-1}$\\

\STATE $\xi_{\m s}$ $\gets$ ${f_g( \mathbf{P}_{\m s}, \m S, \m K_{\m s}, \m p_{\m s})};$ \hspace{5.0cm} \COMMENT {Update spatial Grassmann points}

\STATE $\m S$ ~$\gets$ $f_{s}(\xi_{\m s}, \m \Sigma_{\m s}, \xi_{\m v \m s}, \m K_{\m s}, \m p_{\m s})$; \hspace{4.45cm} \COMMENT {Refine based on top $\m p_{\m s}$ singular values}
\STATE $\m J_{\m s} \gets  \m U_{\m J_{\m s}} \mathcal{S}_{\frac{\lambda_3}{\beta}}(\m \Sigma_{\m J_{\m s}}) \m V_{\m J_{\m s}}^{\m T},\text{ where } [\m U_{\m J_{\m s}}, \m \Sigma_{\m J_{\m s}}, \m V_{\m J_{\m s}}^{\m T}] = \text{svd}(\m C_{\m s} + \frac{\mathbf{L}_2}{\beta})$
\STATE $\m S^{\sharp} \gets \m U \mathcal{S}_{\frac{\gamma}{\beta}}(\m \Sigma) \m V^{\m T}
\text{where,} ~[\m U, \m \Sigma, \m V^{\m T}] = \text{svd}(f(\m S) - \frac{\mathbf{L}_1}{\beta})$\\

\STATE $\m C_{\m t} \gets  \Big(2\lambda_2 \m L_{\m t} \m L_{\m t}^{\m T} + \beta(\m J_{\m t} -\frac{\mathbf{L}_{3}}{\beta}) \Big) \Big(2 \lambda_2 \m L_{\m t} \m L_{\m t}^{\m T} + \beta \m I_{\m t}  \Big)^{-1}$

\STATE $\xi_{\m t}  \gets {f_g(\mathbf{P}_{\m t}, \m S^{\sharp}, \m K_{\m t}, \m p_{\m t})}$ \hspace{5.0cm} \COMMENT {Update temporal Grassmann points}

\STATE $\m S^{\sharp} \gets f_{s}(\xi_{\m t}, \m \Sigma_{\m t}, \xi_{\m v \m t},  \m K_{\m t}, \m p_{\m t})$;\hspace{4.45cm} \COMMENT {Refine based on top $\m p_{\m t}$ singular value} \\

\STATE $\m J_{\m t} \gets  \m U_{\m J_{\m t}} \mathcal{S}_{\frac{\lambda_4}{\beta}}(\m \Sigma_{\m J_{\m t}}) \m V_{\m J_{\m t}}^{\m T}, \text{where} ~[\m U_{\m J_{\m t}}, \m \Sigma_{\m J_{\m t}}, \m V_{\m J_{\m t}}^{\m T}] = \text{svd}(\m C_{\m t} + \frac{\mathbf{L}_3}{\beta})$

\STATE $\m \Gamma_\textrm{ij}^{\m s}$ $\gets$ $\textbf{Tr}[ ({\m \Phi_{\m s\m j}^{\m T}} {\m \Phi_{\m s\m i}} ) ({\m \Phi_{\m s\m i}^{\m T}} {\m \Phi_{\m s\m j}})]$,
$\m \Gamma_\textrm{ij}^{\m t} \gets \textbf{Tr}[({\m \Phi_{\m t\m j}^{\m T}} {\m \Phi_{\m t\m i}} ) ({\m \Phi_{\m t\m i}^{\m T}} {\m \Phi_{\m t\m j}})]$;\\

\STATE ${\m \Gamma}_{\m s} \gets (\m \Gamma_\textrm{ij}^{\m s})_{\m i, \m j = 1}^{\m K_{\m s}}, {\m \Gamma}_{\m t} \gets (\m \Gamma_\textrm{ij}^{\m t})_{\m i, \m j = 1}^{\m K_{\m t}}$;\hspace{4.0cm} \COMMENT{${\m \Gamma}_{\m s} \succeq 0,{\m \Gamma}_{\m t} \succeq 0$, if ${\m \Gamma}_{\m s} || {\m \Gamma}_{\m t} = 0$ add $\delta \mathbf{I}$ to make it  $\succ 0$ }\\

\STATE $\m L_{\m s}\m L_{\m s}^{\m T}$ = \textbf{Chol}(${\m \Gamma}_{\m s}$), $\m L_{\m t}\m L_{\m t}^{\m T}$ = \textbf{Chol}(${\m \Gamma}_{\m t}$); \\

\STATE $\mathbf{P}_{\m s} = f_p(\xi_{\m s}, \m C_{\m s}, \mathbf{P}_{\m s}); ~\mathbf{P}_{\m t} = f_p(\xi_{\m t}, \m C_{\m t}, \mathbf{P}_{\m t})$; \\

\STATE $\m W$ $\gets$ $f_o(\m W, \mathbf{P}_{\m s})$ \hspace{6.2cm}\COMMENT {Note: Column ordering of $\m W$ and $\m S$ must be same.}

\STATE $\mathbf{L}_1$ := $\mathbf{L}_1 + \beta(\m S^{\sharp} - f(\m S))$, $\mathbf{L}_2$ := $\mathbf{L}_2 + \beta(\m C_{\m s} - \m J_{\m s})$, $\mathbf{L}_3$ := $\mathbf{L}_3 + \beta(\m C_{\m t} - \m J_{\m t})$; \COMMENT {Update Lagrange multipliers}\\
\STATE $\beta\gets \textrm{min}(\rho\beta, \beta_m)$ \\
\STATE \textrm{maxgap} := \textrm{max}($[\| \m S^\sharp - f(\m S)\|_{\infty}, \|\m C_{\m s} - \m J_{\m s} \|_{\infty}, \|\m C_{\m t} - \m J_{\m t}\|_{\infty} ]$)
\IF{(maxgap $<$ $\epsilon$ $ \lor ~\beta > \beta_m$)}
	\STATE break;
\ENDIF \hspace{7.2cm} \COMMENT  {Check for the convergence}
\STATE $\textrm{iter} := \textrm{iter} + 1$\\
\ENDWHILE  \hspace{3.0cm} \COMMENT {Note: $\delta$ is a very small positive number and $\mathbf{I}$ symbolizes identity matrix.}
\ENSURE $\m S$, $\m S^\sharp$, $\m C_{\m s}$, $\m C_{\m t}$.
\hspace{2.0cm} \COMMENT {Note: Kindly use economical version of `{\textrm{svd()}}' on a regular desktop.}
\end{algorithmic}
\end{algorithm*}

\noindent
\textbf{Solution to $\m J_{\m s}$ } 
\begin{equation}
\begin{small}
\begin{aligned}
& \displaystyle \equiv \underset{\m J_{\m s}} {\text{argmin}} ~\lambda_3 \|\m J_{\m s}\|_* + \frac{\beta}{2}\|\m C_{\m s} - \m J_{\m s}\|_{\m F}^2 + <\mathbf{L}_2, \m C_{\m s} - \m J_{\m s}> \\
& \displaystyle \equiv \underset{\m J_{\m s}} {\text{argmin}} ~\lambda_3 \|\m J_{\m s}\|_* + \frac{\beta}{2} \|\m J_{\m s} - \big(\m C_{\m s} + \frac{\mathbf{L}_2}{\beta}\big)\|_{\m F}^2
\end{aligned}
\end{small}
\end{equation}

\noindent
Similar to Eq.(\ref{eq:St}) derivation, we use the soft-thresholding operation. It's optimal solution can be obtained as 

\begin{equation}
\begin{small}
\begin{aligned}
\m J_{\m s} \equiv  \m U_{\m J_{\m s}} \mathcal{S}_{\frac{\lambda_3}{\beta}}(\m \Sigma_{\m J_{\m s}}) \m V_{\m J_{\m s}}^{\m T},\text{ where } [\m U_{\m J_{\m s}}, \m \Sigma_{\m J_{\m s}}, \m V_{\m J_{\m s}}^{\m T}] = \text{svd}(\m C_{\m s} + \frac{\mathbf{L}_2}{\beta})
\end{aligned}
\end{small}
\end{equation}

\noindent
\textbf{Solution to $\m J_{\m t}$} 
\begin{equation}
\begin{small}
\begin{aligned}
& \displaystyle \equiv \underset{\m J_{\m t}} {\text{argmin}} ~\lambda_4 \|\m J_{\m t}\|_* + \frac{\beta}{2}\|\m C_{\m t} - \m J_{\m t}\|_F^2 + <\mathbf{L}_3, \m C_{\m t} -\m J_{\m t}> \\
& \displaystyle \equiv \underset{\m J_{\m t}} {\text{argmin}} ~\lambda_4 \|\m J_{\m t}\|_* + \frac{\beta}{2} \|\m J_{\m t} - \big(\m C_{\m t} + \frac{\mathbf{L}_3}{\beta}\big)\|_{\m F}^2
\end{aligned}
\end{small}
\end{equation}

\begin{equation}
\begin{small}
\begin{aligned}
\m J_{\m t} \equiv  \m U_{\m J_{\m t}} \mathcal{S}_{\frac{\lambda_4}{\beta}}(\m \Sigma_{\m J_{\m t}}) \m V_{\m J_{\m t}}^{\m T}, \text{ where} [\m U_{\m J_{\m t}}, \m \Sigma_{\m J_{\m t}}, \m V_{\m J_{\m t}}^{\m T}] = \text{svd}(\m C_{\m t} + \frac{\mathbf{L}_3}{\beta})
\end{aligned}
\end{small}
\end{equation}

\noindent
The pseudo-code with few MATLAB script of our implementation is provided in {\bf{Algorithm (\ref{alg:Algorithm1}}}). This divide and conquer approach works well for most of the available benchmark dataset, however, to make our method more robust to a real-world setting, we took our idea a step further. We know that high-dimensional data representation can be inferior in the presence of noise and outliers unless some filtering techniques are employed. Therefore, we must introduce the notion of low-dimensional representation on Grassmann manifold which preserves the important geometrical information and lets us get rid of noisy information in the data. Inquest of implementing this idea, we proposed a geometry aware extension to our previously proposed dense NRSfM algorithm.



\section{Geometry Aware Idea}

\subsection{Motivation}

\begin{figure}[t]
\begin{center}
\includegraphics[width=0.9\linewidth]{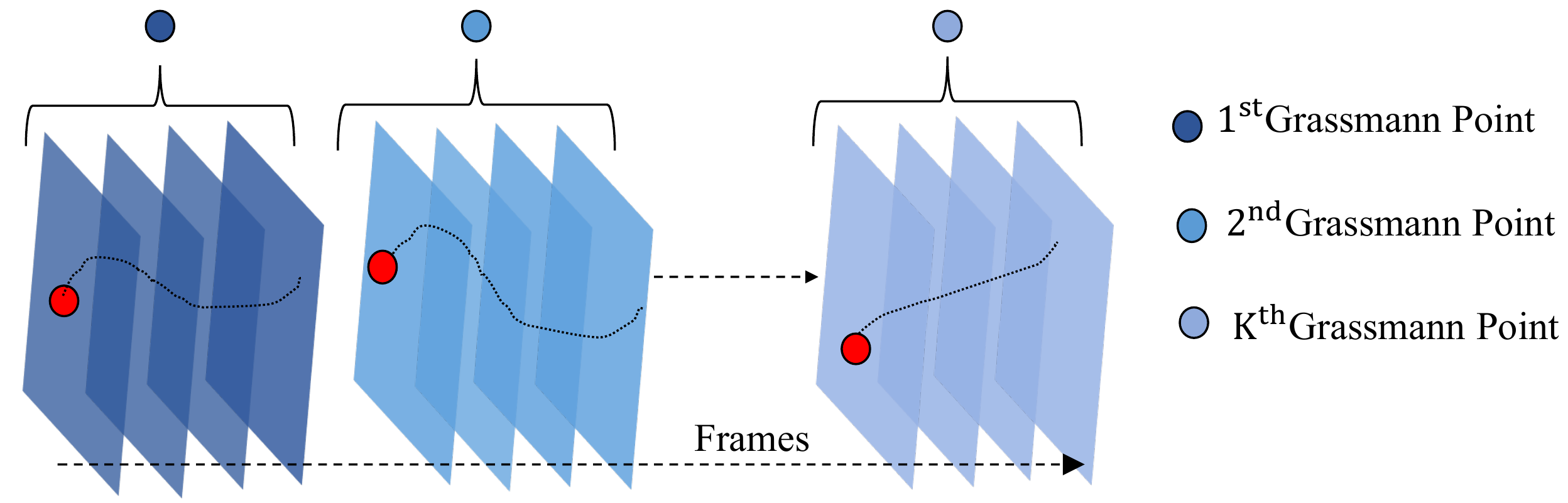}
\caption{\small{Temporal representation using Grassmannians in the shape space introduces discontinuity in the overall trajectory of the feature point. Also, to define neighboring subspace dependency graph in the time domain seems very challenging keeping in mind that the activity/expression may repeat.  Red circle shows the feature point with its trajectory over frames (Black). } \label{fig:temporaldis}}
\end{center}
\end{figure}

The key insight in the last algorithm is; even though the overall complexity of the deforming shape is high, each local deformation may be less complex. Using this idea, we developed a union of local linear subspace approach to solve dense NRSfM problem. Despite its excellent performance, it has some practical concerns.
\emph{Firstly}, the intrinsic issues associated with the modeling of a non-rigidly deforming surface via a \textbf{high-dimensional} Grassmannian representation. Now, such a representation may help reconstruct complex 3D deformation but can lead to wrong clustering ---curse of dimensionality \cite{duda2012pattern}, and it's very important in a joint reconstruction and clustering framework to have suitable clusters of subspaces, else reconstruction may suffer. \emph{Secondly}, the approach to represent local non-linear deformation completely ignores the neighboring surfaces, which may result in an inefficient representation of the Grassmannians in the trajectory space. \emph{Thirdly}, the representation of Grassmannians in the shape space can result in \emph{irredeemable} discontinuity of the trajectories (see Fig.(\ref{fig:temporaldis})). Hence, temporal representation of the set of shapes using Grassmannians seems not an extremely beneficial choice for modeling dense NRSfM on Grassmannian manifold, \textbf{unless prior information is available which in general is not known}. \emph{Lastly}, although the dense NRSfM algorithm proposed in Algorithm (\ref{alg:Algorithm1}) works better and faster than the previous methods, it depends on several manual parameters which are inadmissible for practical applications.

Hence, we extend the idea of our previous approach that can overcome the aforementioned limitations with \textbf{Algorithm (\ref{alg:Algorithm1})}. The main point we are trying to make is that; reconstruction and grouping of subspace on the same high dimensional Grassmann manifold seem like an unreasonable choice. Even recent research in the Riemannian geometry has shown that the low dimensional representation of the corresponding high dimensional Grassmann manifold is more favorable for grouping Grassmannians \cite{huang2015projection} \cite{harandi2014manifold}. So, inspired from these past work, we formulate dense NRSfM in a way that it takes advantage of both high and low dimensional representation of Grassmannians \textit{i.e.}, perform reconstruction in the original high-dimension manifold and cluster subspace on its corresponding low-dimension manifold representation.

We devise an unsupervised approach to efficiently represent the high-dimensional Grassmannians to a lower-dimensional Grassmann manifold via a projection operation. These low-dimensional Grassmannians are represented in such a way, it preserves the local structure of the surface deformation in accordance with its neighboring surfaces. These low-dimensional Grassmannians serves as a potential representative for its high-dimensional Grassmannians for suitable grouping, which subsequently helps improve the reconstruction and representation of the Grassmannians on the high-dimensional Grassmann manifold. Further, we drop the temporal grouping of shapes using Grassmannians to discourage the discontinuity of trajectories (see Fig.(\ref{fig:temporaldis})).

In essence, our modification is inspired by the previously developed idea and is oriented towards settling its important limitations. Moreover, in contrast to \textbf{Algorithm (\ref{alg:Algorithm1})}, we capture the notion of dependent local subspace in a union of subspace algorithm via Grassmannian modeling  \cite{larsson2017compact}. The algorithm we proposed is an attempt to supply a more efficient, reliable and practical solution to this problem. Our new formulation gives an efficient framework for modeling dense NRSfM on the Grassmann manifold. We observed empirically that this method is more useful and is as accurate and efficient than \textbf{Algorithm (\ref{alg:Algorithm1})}. The performance of this algorithm stands superior in handling noise. The main highlights of this algorithm are as follows:

\begin{enumerate}
\item An efficient framework for modeling non-rigidly deforming surface that exploits the advantage of Grassmann manifold representation of different dimensions based on its geometry.
\item A formulation that encapsulates the local non-linearity of the deforming surface w.r.t its neighbors to enable the proper inference and representation of local linear subspaces. 
\item An iterative solution to the proposed cost function based on ADMM \cite{boyd2011distributed}, which is simple to implement and provide results as good as \textbf{Algorithm(\ref{alg:Algorithm1})} and in addition to that, it helps improve the 3D reconstruction substantially, in the case of noisy trajectories.
\end{enumerate}

\subsection{New Grassmannian Representation}\label{ss:gr_rep}

To properly represent Grassmannian which respects the neighboring non-linearity in low-dimension, we introduce a different strategy to model non-rigid surface in low-dimension. For now, let $\Delta \in \mathbb{R}^{{\m d} \times \tilde{\m d} }$ be a matrix that maps `$\m \Phi_{\m i}$' $\in \mathcal{G}(\m p, {\m d})$ to  `$\m \phi_{\m i}$' $\in \mathcal{G}(\m p, \tilde{\m d})$ such that $ \tilde{\m d} < {\m d}$. Mathematically, 
\begin{equation}
\m \phi_{\m i} = \m \Delta^{\m T} \m \Phi_{\m i}
\end{equation}
Its quite easy to examine that $\m \phi_i$ is not a orthogonal matrix and, therefore, may not qualify as a potential point on a Grassmann manifold. However, by performing a orthogonal-triangular (QR) decomposition of $\m \phi_i$, we estimate the new representative of $\m \phi_i$ on the Grassmann manifold of `$\tilde{\m d}$' dimension.
\begin{equation}\label{eq:CVPR19_2}
\m \Theta_{\m i} \m U_{\m i} = \mathbf{qr}(\phi_{\m i}) = \m \Delta^{\m T} \m \Phi_{\m i} 
\end{equation}

\begin{figure}[t]
\begin{center}
\includegraphics[width=0.8\linewidth]{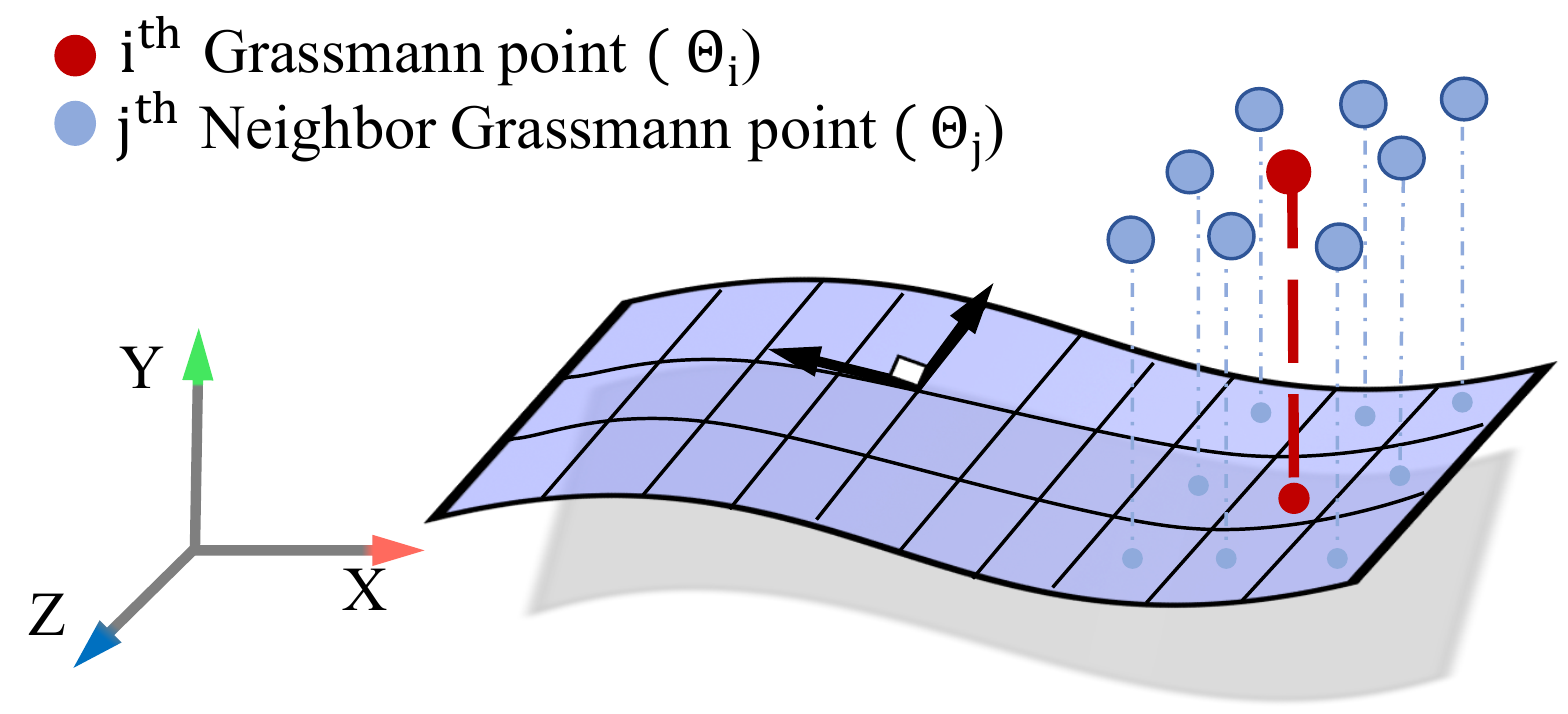}
\caption{\small{In contrast to Algorithm (1) representation, the modeling of surface using Grassmannians considers the similarity between the neighboring Grassmannians while representing it in the lower dimension. Based on the assumption that spatially neighboring surface tend to span similar subspace, defining neighboring subspace dependency graph is easy and, most of the real-world examples follows such an assumption. However, building such graph in shape space can be tricky.} \label{fig:grassmannneigh}}
\end{center}
\end{figure}

\begin{figure}[t]
\begin{center}
\includegraphics[width=0.8\linewidth]{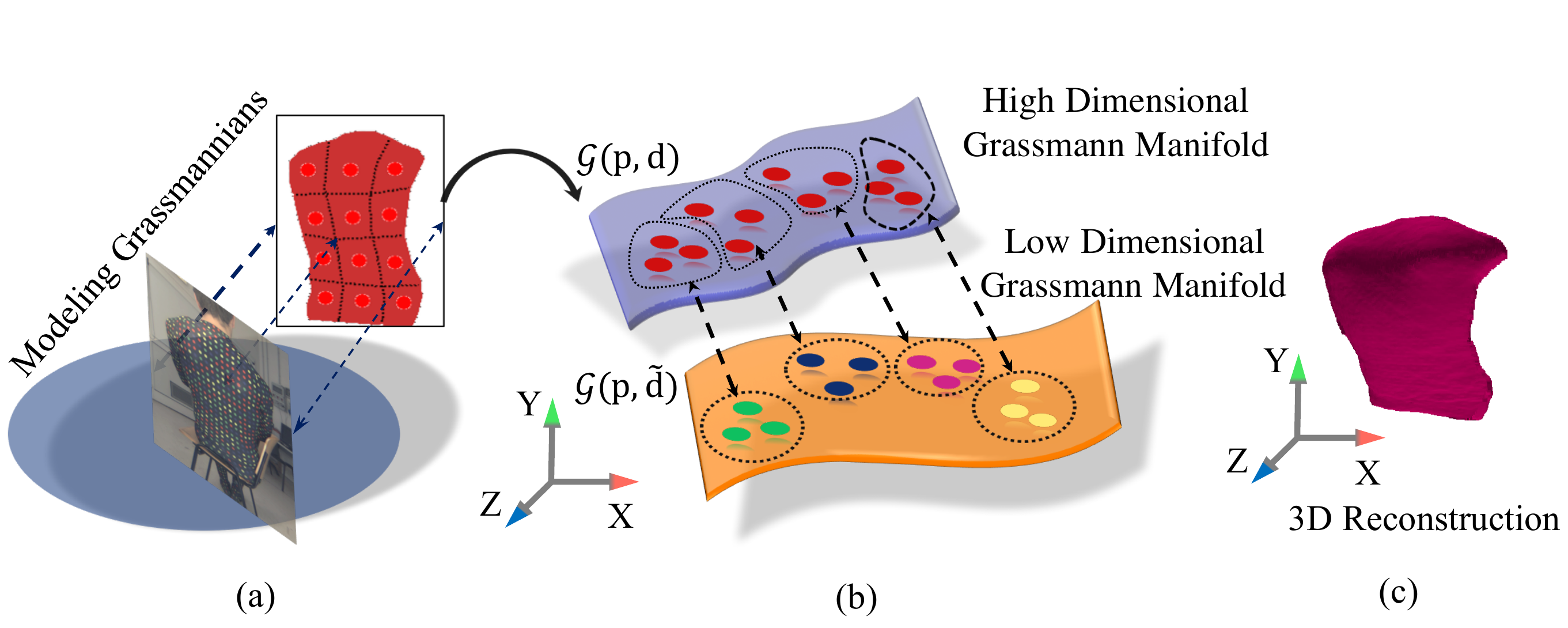}
\caption{\small{Conceptual illustration of our modeling (a) Modeling of 3D trajectories to Grassmann points (b) The two grassmann manifold and mapping of the points between them to infer better cluster index that leads to better reconstruction (c) The 3D reconstruction of the non-rigid deforming object.} \label{fig:grassmannneighconcept}}
\end{center}
\end{figure}

Here, $\mathbf{qr(.)}$ is a function that returns the QR decomposition of the matrix. The $\m \Theta_{\m i} \in \mathbb{R}^{\tilde{\m d} \times \m p}$ is an orthogonal matrix and $\m U_{\m i} \in \mathbb{R}^{\m p \times \m p} $ is the upper triangular matrix\footnote{Note: The value of $\tilde{\m d} \geq \m p$, Use $[\m \Theta_{\m i}, \m U_{\m i}]=\textbf{qr}(\m \phi_{\m i}, 0)$ in MATLAB to get a square $\m U_{\m i}$ matrix ($\m U_{\m i} \in \mathbb{R}^{\m p \times \m p}$)}. Using Eq.(\ref{eq:CVPR19_2}), we represent the equivalence of $\m \Phi_{\m i}$ in low dimension as

\begin{equation}\label{eq:CVPR19_3}
\begin{aligned}
& \displaystyle \m \Theta_{\m i} = \m \Delta^{\m T}(\m \Phi_{\m i} \m U_{\m i}^{-1})\\
& \displaystyle \m \Theta_{\m i} = \m \Delta^{\m T} \m \Omega_{\m i}
\end{aligned}
\end{equation}
where, $\m \Omega_{\m i} = \m \Phi_{\m i} \m U_{\m i}^{-1} \in \mathbb{R}^{{\m d} \times \m p}$. The key-point to note is that both $\m \Theta_{\m i}$ and $\phi_{\m i}$ has the same column space. In principle such a representation is useful however, it does not serve the purpose of preserving the non-linearity w.r.t its neighbors. In order to encapsulate the local dependencies (see Fig.(\ref{fig:grassmannneigh}), Fig.(\ref{fig:grassmannneighconcept})), we further constrain our representation as:
\begin{equation}\label{eq:CVPR19_4}
\m E (\m \Delta) = \underset{\m \Delta}  {\text{minimize}} \sum_{(\m i, \m j)}^{\m K} \m w_{\m i\m j} \frac{1}{2}\|\m \Pi(\m \Theta_{\m i})-\m \Pi(\m \Theta_{\m j})\|_2^{\m F}
\end{equation}
The parameter `$\m w_{\m i \m j}$' accommodate the similarity knowledge between the two Grassmannians. Using the {\bf{Definition}}(\ref{def:2}) and Eq.(\ref{eq:CVPR19_3}), we further simplify Eq.(\ref{eq:CVPR19_4}) as   
\begin{equation}\label{eq:CVPR19_5}
\begin{aligned}
& \displaystyle \m E (\m \Delta) \equiv \underset{\m \Delta}  {\text{minimize}} \sum_{(\m i, \m j)}^{\m K} \m w_{\m i\m j}\frac{1}{2}\|\m \Delta^{\m T} \m \Omega_{\m i} \m \Omega_{\m i}^{\m T} \m \Delta - \m \Delta^{\m T} \m \Omega_{\m j} \m \Omega_{\m j}^{\m T} \m \Delta \|_{\m F}^{2} \\
& \displaystyle \m E (\m \Delta) \equiv \underset{\m \Delta}  {\text{minimize}} \sum_{(\m i, \m j)}^{\m K} \m w_{\m i\m j}\frac{1}{2} \|\m \Delta^{\m T}(\m \Omega_{\m i}\m \Omega_{\m i}^{\m T} -\m \Omega_{\m j}\m \Omega_{\m j}^{\m T}) \m \Delta  \|_{\m F}^{2} \\
& \displaystyle \m E (\m \Delta) \equiv \underset{\m \Delta}  {\text{minimize}} \sum_{(\m i, \m j)}^{\m K} \m w_{\m i\m j}\frac{1}{2} \|\m \Delta^{\m T}(\m \Lambda_{\m i\m j})\m \Delta  \|_{\m F}^{2} \\
\end{aligned}
\end{equation}

where, $\m \Lambda_{\m i \m j} \in \textrm{Sym}({\m d})$. The parameter `$\m w_{\m i \m j}$' (\textbf{similarity graph}) is set as $\textrm{exp}(-d_{g}^{2}(\m \Phi_{\m i}, \m \Phi_{\m j}))$ with $d_{g}$ as the projection metric (see {\bf{Definition}} (\ref{def:2})). Eq.(\ref{eq:CVPR19_5}) is an unconstrained optimization problem and its solution may provide a trivial solution. To estimate the useful solution, we further constrain the problem. Using $\m i^{\m t \m h}$ Grassmann point `$\m \Omega_{\m i}$' and its neighbors, expand Eq.(\ref{eq:CVPR19_5}). By performing some simple algebraic manipulation, Eq.(\ref{eq:CVPR19_5}) reduces to 

\begin{equation}\label{eq:CVPR19_6}
\mathbf{Tr}\Big(\m \Delta^{\m T}\big(\sum_{\m i = 1}^{\m K}\m \lambda_{\m i \m i}\m \Omega_{\m i}\m \Omega_{\m i}^{\m T}\big)\m \Delta\Big)
\end{equation}
where, $\m \lambda_{\m i\m i} = \sum_{\m j = 1}^{\m K} \m w_{\m i \m j}$. Constraining the value of Eq.($\ref{eq:CVPR19_6}$) to 1 provides the overall optimization for an efficient representation of the local non-rigid surface on the Grassmann manifold.
\begin{equation}\label{eq:CVPR19_7}
\begin{aligned}
& \displaystyle \m E(\m \Delta) \equiv \underset{\m \Delta}  {\text{minimize}} \sum_{(\m i, \m j)}^{\m K} \m w_{\m i\m j}\frac{1}{2} \|\m \Delta^{\m T}(\m \Lambda_{\m i\m j})\m \Delta  \|_{\m F}^{2}\\
& \displaystyle \textrm{subject to:}\\
& \displaystyle \mathbf{Tr}\Big(\m \Delta^{\m T}\big(\sum_{\m i = 1}^{\m K}\m \lambda_{\m i \m i}\m \Omega_{\m i}\m \Omega_{\m i}^{\m T}\big)\m \Delta\Big) = 1
\end{aligned}
\end{equation}

Its easy to verify that the matrix `$\m \Lambda$' and `$\big(\sum_{\m i = 1}^{\m K}\m \lambda_{\m i \m i}\m \Omega_{\m i}\m \Omega_{\m i}^{\m T}\big)$' are symmetric and positive semi-definite, and therefore, the above optimization can be solved as a generalized eigen value problem.

\subsection{Solution to $\m E (\Delta)$}
\begin{equation}\label{eq:CVPR19_S15}
\begin{aligned}
& \displaystyle \m E(\m \Delta) \equiv \underset{\m \Delta}  {\text{minimize}} \sum_{(\m i, \m j)}^{\m K} \m w_{\m i\m j}\frac{1}{2} \|\m \Delta^{\m T}(\m \Lambda_{\m i\m j})\m \Delta  \|_{\m F}^{2}\\
& \displaystyle \textrm{subject to:}\\
& \displaystyle \textbf{Tr}\Big(\m \Delta^{\m T}\big(\sum_{\m i = 1}^{\m K}\lambda_{\m i \m i}\m \Omega_{\m i}\m \Omega_{\m i}^{\m T}\big)\m \Delta\Big) = 1
\end{aligned}
\end{equation}
The optimization equation proposed for $\m E (\m \Delta)$ is a well-studied optimization form and Riemann Conjugate gradient toolbox can be employed to achieve the solution. Nevertheless, we can also derive augmented lagrangian form to solve the same problem. By letting $\m X$ = $\big(\sum_{\m i = 1}^{\m K}\lambda_{\m i \m i}\m \Omega_{\m i}\m \Omega_{\m i}^{\m T}\big)$ and expanding the Frobenius norm term, we can re-write the equation as:
\begin{equation}\label{eq:CVPR19_S16}
\begin{aligned}
& \displaystyle \m E(\m \Delta) \equiv \underset{\m \Delta}  {\text{minimize}} \sum_{(\m i, \m j)}^{\m K} \frac{\m w_{\m i\m j}}{2} \textbf{Tr}\big(\m \Delta^{\m T}\m \Lambda_{\m i\m j}\m \Delta \m \Delta^{\m T} \m \Lambda_{\m i\m j} \m \Delta\big) \\
& \displaystyle \m E(\m \Delta) \equiv \underset{\m \Delta}  {\text{minimize}} ~\textbf{Tr}\Big({\m \Delta}^{\m T} \sum_{(\m i, \m j)}^{\m K} \frac{\m w_{\m i\m j}}{2} \m \Lambda_{\m i\m j}\m \Delta^{t-1} \m \Delta^{\m (t-1)\m T} \m \Lambda_{\m i\m j} \m \Delta\Big)\\
& \displaystyle \textrm{subject to:}\\
& \displaystyle \textbf{Tr}\Big(\m \Delta^{\m T}\m X \m \Delta\Big) = 1
\end{aligned}
\end{equation}
Here, $t-1$ refers to its known value before the current iteration. Now, by assuming $\m Y = \frac{\m w_{\m i\m j}}{2} \m \Lambda_{\m i\m j} \m \Delta^{t-1} \m \Delta^{\m (t-1)\m T} \m \Lambda_{\m i\m j} $, the above equation simplifies to standard eigen value decomposition problem \textit{i.e,}

\begin{equation}
\begin{aligned}
& \displaystyle \m E(\Delta) \equiv \underset{\m \Delta}  {\text{minimize}} ~\textbf{Tr}(\m \Delta^{\m T} \m Y \m \Delta) \\
& \displaystyle \textrm{subject to:}\\
& \displaystyle \textbf{Tr}\Big(\m \Delta^{\m T}\m X \m \Delta\Big) = 1
\end{aligned}
\end{equation}

\noindent
The equivalent Lagrangian function form is given by 
\begin{equation}\label{eq:CVPR19_S18}
\textrm{Tr}(\m \Delta^{\m T} \m Y \m \Delta) + \lambda\Big(1- \textbf{Tr}\big(\m \Delta^{\m T}\m X \m \Delta\big)\Big)
\end{equation}
Eq.(\ref{eq:CVPR19_S18}) is in the generalized eigen value problem form. Any standard linear algebra package can be used to solve it.

\subsection{Geometry Aware Extension}
Similar to the previous algorithm, we introduce the local subspace constraint on the shape, we use the notion of self-expressiveness on the non-linear Grassmann manifold space.
\begin{equation}\label{eq:CVPR19_10}
\small
\begin{aligned}
& \displaystyle \underset{\m E, {\m C}, \m S^{\sharp}}  {\text{minimize}} ~\| \m E \|_{\mathcal{G}}^2 + \beta_2\| \m S^{\sharp}\|_* + \beta_3 \| \m C\|_*\\
& \displaystyle \textrm{subject to:} ~\m S^{\sharp} = f(\m S), ~\m S = \m S \m C + \m E
\end{aligned}
\end{equation}

As defined before $f: \m S \in \mathbb{R}^{3 \m F \times \m P} \mapsto \m S^{\sharp} \in \mathbb{R}^{3 \m P \times \m F}$ and $\m C \in \mathbb{R}^{\m P \times \m P}$ as the coefficient matrix. We know from our previous discussion that the Grassmann manifold is isometrically equivalent to the symmetric idempotent matrix \cite{chikuse2012statistics}. So, we embed the Grassmann manifold into symmetric matrix manifold to define the self-expressiveness. Let $\tilde{\xi_{\m s}} = \{ \m \Theta_1, \m \Theta_2, ..., \m \Theta_{\m K} \}$ be the set of Grassmannians on a low dimensional Grassmann manifold. The elements of $\tilde{\xi_{\m s}}$ are the projection of high dimensional Grassmannian representation of the columns of `$\m S$' matrix. Let $\rchi = \{ (\m \Theta_1\m \Theta_1^{\m T}), (\m \Theta_2\m \Theta_2^{\m T}), ..., (\m \Theta_{\m K}\m \Theta_{\m K}^{\m T}) \}$ be its embedding onto symmetric matrix manifold. Using such embedding techniques we re-write Eq.(\ref{eq:CVPR19_10}) as

\begin{equation}\label{eq:CVPR19_11}
\small
\begin{aligned}
& \displaystyle \underset{\m E, \tilde{\m C}, \m S^{\sharp}}  {\text{minimize}} ~ \| \m E \|_{\m F}^2 + \beta_2\| \m S^{\sharp}\|_* + \beta_3 \| \tilde{\m C} \|_* \\
& \displaystyle \textrm{subject to:} \m S^{\sharp} = f(\m S), \rchi = \rchi \tilde{\m C} + \m E
\end{aligned}
\end{equation}
where, $\tilde{\m C} \in \mathbb{R}^{\m K \times \m K}$  and $\rchi \in \mathbb{R}^{\tilde{\m d} \times \tilde{\m d} \times \m K}$ denotes the coefficient matrix of Grassmannians and structure tensor respectively, with $\m K$ as the total number of Grassmannians. Generally, $\m K<<\m P$, which makes such representation scalable.

Similar to previous notations, let $\mathbf{P}$ $\in \mathbb{R}^{1 \times \m P}$ be an ordering vector that contains the index of columns of $\m S$. Also, using the the function definition form $\{(\textrm{output}, \textrm{function(.)}): \textrm{definition} \}$, we define $f_h$ as

\begin{equation}\label{eq:CVPR19_13}
\small
\begin{aligned}
& \displaystyle \big\{ \big(\tilde{\xi_{\m s}}, f_h(\m \Delta, \xi_{\m s})\big): \tilde{\xi_{\m s}} = \{\m \Theta_{\m i}\}_{\m i=1}^{\m K}, \m \Theta_{\m i} = \m \Delta^{\m T}(\m \Phi_{\m i} \m U_{\m i}^{-1}), \\ 
& \displaystyle \textrm{where,} ~\m \Delta  = \text{solution to the minimization of Eq.}(\ref{eq:CVPR19_7})\big\}
\end{aligned}
\end{equation}
Intuitively, The function ($f_h$) projects the Grassmannians to a lower dimension in accordance with the neighbors using Eq.(\ref{eq:CVPR19_7})

\noindent
{\bf{Objective Function:}}
Combining all the above terms and constraints provides our overall cost function. 
\begin{equation}\label{eq:CVPR19_15}
\small
\begin{aligned}
& \displaystyle \underset{\m E, \tilde{\m C}, \m S, \m S^{\sharp}}  {\text{minimize}} \frac{1}{2}\|\m W - \m R \m S\|_{\m F}^2 +  \beta_1\| \m E \|_{\m F}^2 + \beta_2\| \m S^{\sharp}\|_* + \beta_3 \| \tilde{\m C} \|_* \\
& \displaystyle \textrm{subject to:}\\
& \displaystyle \m S^{\sharp} = f(\m S), \rchi = \rchi \tilde{\m C} + \m E, \\
& \displaystyle {\xi_{\m s}} =  f_g(\mathbf{P}, \m S, \m K, \m p), \tilde{\xi_{\m s}} =  f_h(\m \Delta, \xi_{\m s}),  \\
& \displaystyle \m S =  f_s({\xi_{\m s}}, \m \Sigma, {\xi}_{\m v}, \m K, \m p), \mathbf{P} = f_p(\tilde{\xi_{\m s}}, \tilde{\m C}, \mathbf{P}_{\m s \m o})
\end{aligned}
\end{equation}

\noindent
where $\mathbf{P}_{\m o}$ vector contains the initial ordering of the columns of `$\m W$' and `$\m S$'. The function $(f_p)$ provides the ordering index to rearrange the columns of `$\m S$' matrix to be consistent with `$\m W$' matrix. This is important because, grouping the set of columns of `$\m S$' over iteration, disturbs its initial arrangements. The definition of $f_g$, $f_s$ and $f_p$ is same as outlined in Eq(\ref{eq:fg}), Eq(\ref{eq:fs}) and Eq(\ref{eq:fp}) respectively.

\subsection{Solution}
The optimization proposed in Eq.(\ref{eq:CVPR19_15}) is a coupled optimization problem. Several methods of Bi-level optimization can be used to solve such minimization problem \cite{bard2013practical, gould2016differentiating}. Nevertheless, we propose ADMM \cite{boyd2011distributed} based solution due to its application in many non-convex optimization problems. The key point to note is that one of our constraint is composed of separate optimization problem $(f_h)$ \textit{i.e.}, the solution to Eq.(\ref{eq:CVPR19_7}), and therefore, we cannot directly embed the constraint to the main objective function. Instead, we only introduce two Lagrange multiplier $\mathbf{L_1, L_2}$ to concatenate a couple of constraints back to the original objective function. The remaining constraints are enforced over iteration.  To decouple the variable $\tilde{\m C}$ from $\rchi$, we introduce auxiliary variable $\tilde{\m C} = \m Z$. We apply these operations to our optimization problem to get the following Augmented Lagrangian form:

\begin{equation}\label{eq:CVPR19_16}
\small
\begin{aligned}
& \displaystyle \underset{\m Z, \tilde{\m C}, \m S, \m S^{\sharp}}  {\text{minimize}} \frac{1}{2}\|\m W - \m R \m S\|_{\m F}^2 +  \beta_1\| \rchi - \rchi \tilde{\m C} \|_{\m F}^2 + \beta_2\| \m S^{\sharp}\|_* + \frac{\beta}{2}\| \m S^{\sharp} - f(\m S)\|_{\m F}^2 +\\
&\displaystyle ~~~~~~~~~~~~~ <\mathbf{L_1}, \m S^{\sharp} - f(\m S)> + \beta_3 \| \m Z \|_* + \frac{\beta}{2}\| \tilde{\m C} - \m Z\|_{\m F}^2 + <\mathbf{L_2}, \tilde{\m C} - \m Z>\\
& \displaystyle \textrm{subject to:} \\
& \displaystyle {\xi_{\m s}} =  f_g(\mathbf{P}, \m S, \m K, \m p), ~\tilde{\xi_{\m s}} =  f_h(\m \Delta, \xi_{\m s})  \\
& \displaystyle \m S =  f_s({\xi_{\m s}}, \m \Sigma, {\xi}_{\m v}, \m K, \m p), ~\mathbf{P} = f_p(\tilde{\xi_{\m s}}, \tilde{\m C}, \mathbf{P}_{\m o})
\end{aligned}
\end{equation}

\noindent
Note that $\tilde{\m C}$ provides the information about the subspace, not the vectorial points. However, we have the chart of the trajectories and its corresponding subspace. Once, we group the trajectories based on $\tilde{\m C}$, $f_g(.)$ provides new Grassmann sample corresponding to each group. The definition of $f_h(.)$ and $f_s(.)$ is provided in Eq.(\ref{eq:CVPR19_7}) and Eq.(\ref{eq:fs}) respectively. More generally, the solution to the optimization in Eq.(\ref{eq:CVPR19_7}) is obtained by solving it as a generalized eigenvalue problem. To keep the order of columns of `$\m S$' matrix consistent with `$\m W$' matrix $f_p(.)$ provides the ordering index. We provide the implementation details of our method with suitable MATLAB commands in the \textbf{Algorithm (\ref{algo:Algorithm2})}.

\begin{algorithm}[t!]
\caption{\small {Geometry Aware Dense NRSfM}}
\label{algo:Algorithm2}
\begin{algorithmic}
\REQUIRE
$\m W$, $\m R$, $\{\beta_{\m i}\}_{\m i=1}^3$, $\beta$=$e^{-2}$, $\beta_m$=$e^{8}$, $\epsilon$=$e^{-10}$, $c$ =$1.1$, $\m K$; \\ 
\hspace{-0.3cm}{\bf Initialize:} ${\m S}$=$\mathbf{pinv}(\m R) \m W$, ${\m S^{\sharp}}$=$f(\m S)$, $\m Z$=$\mathbf{0}$, $\{{\bf L}_{\m i}\}_{\m i=1}^2$=$\mathbf{0}$, $\tilde{\m d}$;\\ 
\hspace{0.5cm}$\Delta$ = $[\mathbf{I}_{\tilde{\m d} \times \tilde{\m d}};$ $\textrm{random values}],$ $\m p $ $\%$top singular values \\
\hspace{0.5cm}$\mathbf{P}_{\m s \m o}$ = $\textrm{kmeans++}(\m S, \m K)$, $\textrm{iter}$ = 1, $\mathbf{P}_\textrm{store}(\textrm{iter}, :)$ = $\mathbf{P}_{\m s \m o}$,\\
\hspace{0.5cm}$\mathbf{P}$ = $\mathbf{P}_{\m s \m o}$\\
{\bf Define:} ~$\mathcal{S}_{\tau}(\m x) := \textcolor{mygreen}{@}(\m x, \tau) \textrm{sign}(\m x).*\textrm{max}(\textrm{abs}(\m x)-\tau, 0);$

\WHILE {not converged}
{\small
\STATE 1. $\m S$ := mldivide\big($\m R^{\m T}\m R + \beta \m I$, $ \beta(f^{-1}(\m S^{\sharp}) + \frac{f^{-1}(\mathbf{L_1})}{\beta}) + \m R^{\m T}\m W$\big);
\STATE 2. $\xi_{\m s} := f_g(\mathbf{P}, \m S, \m K, \m p)$; see Eq.(\ref{eq:fg})
\STATE 3. $\m W := \text{arrange\_column}(\mathbf{P}, \m W)$
\STATE 4. Update the similarity matrix `$\m w_\textrm{ij}$' using $\xi_{\m s}$. \S \ref{ss:gr_rep}
\STATE 5. $\tilde{\xi_{\m s}} := f_{h}(\xi_{\m s}, \m \Delta); \text{s.t}, \Delta \equiv \underset{\m \Delta}{\textrm{minimize}}~~\m E(\m \Delta);$  see Eq.(\ref{eq:CVPR19_13})
\STATE 6. $\m \Gamma_\textrm{ij}=\textbf{Tr}[(\m \Theta_{\m j}^{\m T}\m \Theta_{\m i})(\m \Theta_{\m i}^{\m T}\m \Theta_{\m j})]; \m \Gamma=(\m \Gamma_\textrm{ij})_{\m i\m j=1}^{\m K};\m L = \textbf{Chol}(\m \Gamma)$ 
\STATE 7. $ \tilde{\m C}$ := $\big(2\beta_1 \m L \m L^{\m T} + \beta(\m Z - \frac{\mathbf{L_2}}{\beta})\big)$ $\big(2\beta_1 \m L \m L^{\m T} + \beta \m I\big)^{-1}$;
\STATE 8. $\mathbf{P} := f_{p}(\tilde{\xi_{\m s}}, \tilde{\m C}, \mathbf{P})$;

\STATE 9. $\m S := f_s({\xi_{\m s}}, {\m \Sigma}, {\xi}_v, \m K, \m p);$  see Eq.(\ref{eq:fg}), Eq.(\ref{eq:fs})

\STATE 10. $\m S^{\sharp}:= \m U_{\m s}\mathcal{S}_{\frac{\beta_2}{\beta}}(\m \Sigma_{\m s})\m V_{\m s}; \text{s.t},[\m U_{\m s},\m \Sigma_{\m s},\m V_{\m s}] := \textrm{svd}(f(\m S) - \frac{\mathbf{L_1}}{\beta})$

\STATE 11. $\m Z := \m U_{\m z}\mathcal{S}_{\frac{\beta_3}{\beta}}(\m \Sigma_{\m z})\m V_{\m z}; \text{s.t}, [\m U_{\m z}, \m \Sigma_{\m z},\m V_{\m z}] := \textrm{svd}(\tilde{\m C} + \frac{\mathbf{L_2}}{\beta});$
\STATE 12. $\mathbf{L_1} := \mathbf{L_1} + \beta(\m S^{\sharp} - f(\m S)); \mathbf{L_2} := \mathbf{L_2} + \beta(\tilde{\m C} - \m Z)$
\STATE 13. $\textrm{iter} := \textrm{iter} + 1;$ $\mathbf{P}_\textrm{store}(\textrm{iter}, :) := \mathbf{P};$ 
\STATE 14. $\beta := \textrm{min}(\beta_m, c\beta);$ \\
\STATE 15. $\textrm{gap} := ~~\textrm{max}\{\|\m S^{\sharp} - f(\m S)\|_\infty, \| \tilde{\m C} - \m Z\|_\infty\};$\\
$(\textrm{gap}<\epsilon) \lor (\beta > \beta_m)\rightarrow \textbf{break};\textrm{\%convergence check}$
} \\
\ENDWHILE
\RETURN $\m S$;\\
\hspace{-0.3cm}$e_{3 \m D}$ = \textbf{Estimate\_error} $(\m S_\textrm{est} = \m S, \m S_{\m G \m T}, \mathbf{P}_\textrm{store});$  \%use Eq.(\ref{eq:CVPR19_17})
\end{algorithmic}
\end{algorithm}

\begin{figure*}
\centering
\includegraphics[width=1.0\textwidth] {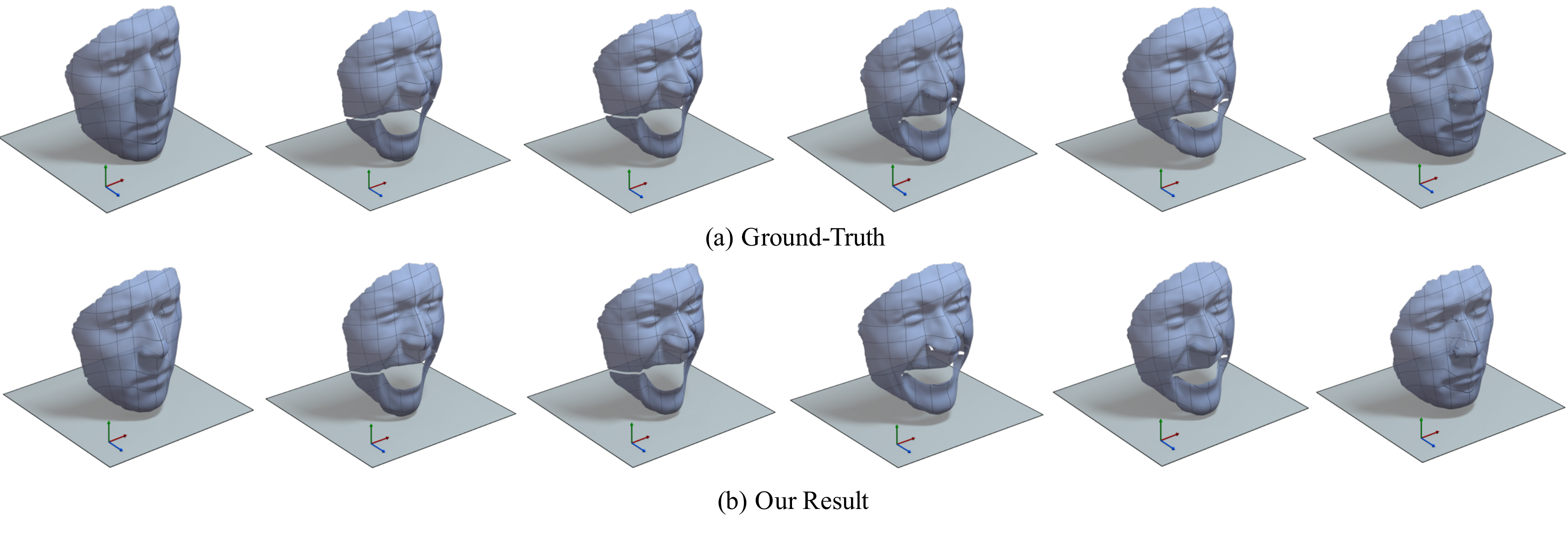}
\caption{\small{Reconstruction results obtained on synthetic dense face dataset (face sequence 4). {\bf{Top row}} : Ground-truth 3D points, {\bf{Bottom row}} : Recovered 3D shape using Algorithm 1. Visually, the 3D reconstruction results recovered using both the algorithms looks very similar. }}
\label{fig:syntheticSeq4res}
\end{figure*}

\section{Experimental Evaluation}
We performed extensive qualitative and quantitative evaluations on the available standard benchmark datasets \cite{garg2013dense} \cite{varol2012constrained} \cite{beeler2011high}. To keep our evaluations consistent with the previous methods, we compute the average 3D reconstruction quality of the estimated shape `$\m S_\textrm{est}$' using the following equation
\begin{equation}\label{eq:CVPR19_17}
\m e_{3 \m D} = \frac{1}{\m F} \sum_{\m i = 1}^{\m F}\frac{\|\m S_\textrm{ est}^{\m i} - \m S_{\m G\m T}^{\m i}\|_{\m F}}{\|\m S_{\m G\m T}^{\m i}\|_{\m F}}
\end{equation}
here, `$\m S_{\m G\m T}$' denotes the ground-truth 3D shape matrix. The qualitative results for both the algorithm looks very similar, yet, they are statistically different (see Table \ref{tab:statisticalresults}).\\
\noindent
\textbf{Initialization:} The initialization for both the algorithms are straight-forward and outlined in the respective algorithm table. In brief, we used Intersection method \cite{dai2014simple} to estimate the rotation matrix and initialize $\m S = \textbf{pinv}(\m R)\m W$. The initial grouping of the trajectories or columns of $\m S$ is done using k-means++ algorithm \cite{arthur2007k}. These initial grouping is used to initialize the ordering vector $\mathbf{P}_{\m o}$, $\mathbf{P}$ and, the Grassmann points $\{\Phi_{\m i} \}_{\m i=1}^{\m K} \in \xi$ via subset of singular vectors. To represent the Grassmannians in the lower-dimension, we solve Eq.(\ref{eq:CVPR19_S15}) to initialize $\tilde{\xi}$ and store corresponding singular values. The similarity matrix or graph in Eq.(\ref{eq:CVPR19_S15}) is constructed using the distance measure between the Grassmannians in the embedding space \S \ref{ss:gr_rep}.\\
\noindent
{\bf{1. Results on synthetic Face dataset:}}
The synthetic face dataset is composed of four distinct sequence \cite{garg2013dense} with 28,880 feature points tracked over multiple frames. Each sequence captures the human facial expression with a different range of deformations and camera motion. Sequence 1 and Sequence 2 are 10 frame long video with rotation in the range $\pm 30^{\circ}$ and $\pm 90^{\circ}$ respectively. Sequence 3 and Sequence 4 are 99 frame long video that contains high frequencies and low frequencies rotation respectively. It's a challenging dataset mainly due to different rotation frequencies and deformations in each of the facial expression sequence. Table (\ref{tab:statisticalresults}) shows the statistical results obtained on these four sequences using both of our algorithms. Fig.(\ref{fig:syntheticSeq4res}) show the qualitative results on face sequence 4 of the dataset.

\begin{figure*}
\centering
\includegraphics[width=1.0\textwidth] {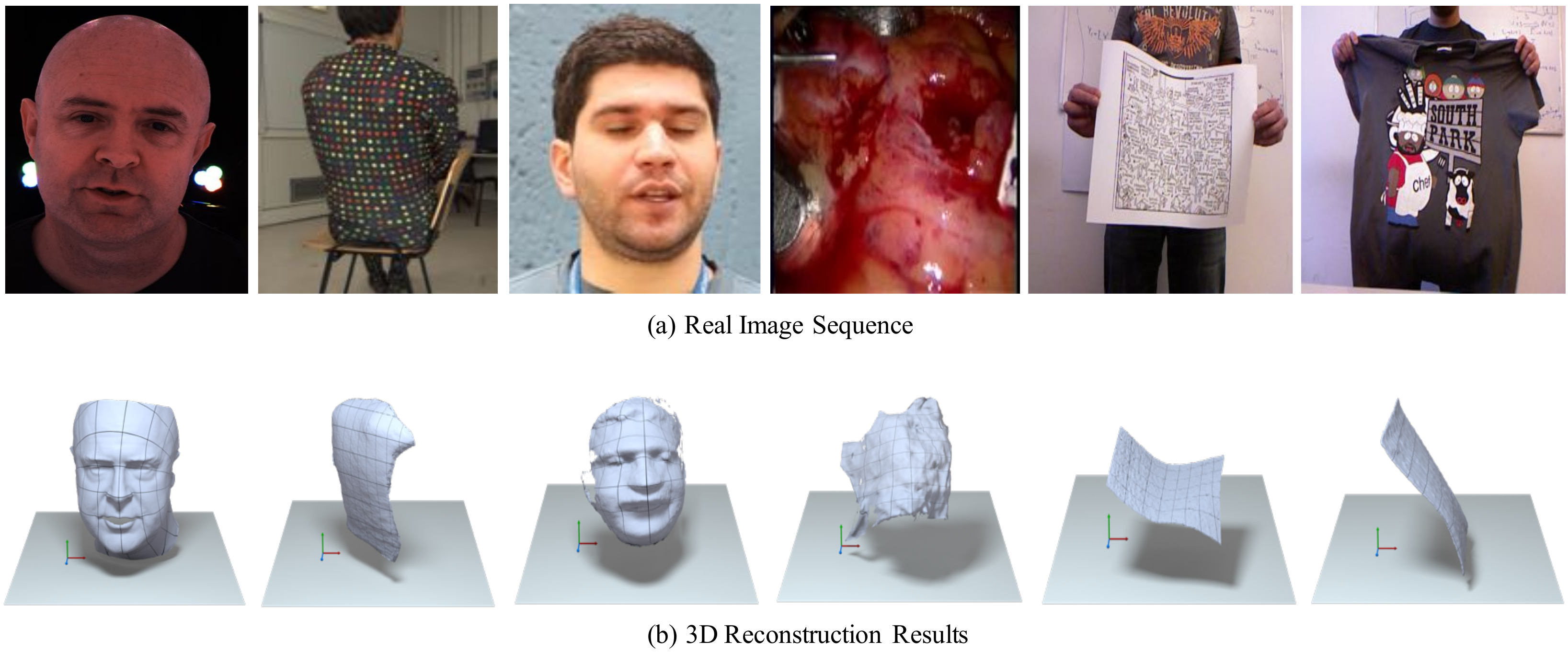}
\caption{\small{Reconstruction results obtained on real dataset sequence. {\bf{Top row}}: Real image sequence, {\bf{Bottom row}}: Recovered 3D shape using our approach. \textbf{Left to Right}: Actor \cite{beeler2011high}, Back \cite{garg2013dense}, Face \cite{garg2013dense}, Heart \cite{garg2013dense}, kinect\_paper \cite{varol2012constrained}, kinect\_tshirt \cite{varol2012constrained} dataset.}}
\label{fig:real6res}
\end{figure*}

\begin{table*}[h]
\centering
\small
\begin{tabular}{|>{\columncolor[gray]{0.85}}c|c|c|c|c|c|c|c|c|c|}
\hline
\rowcolor[gray]{0.65}
Dataset $\downarrow \slash$ Method $\rightarrow$& MP  & PTA  & CSF1 & CSF2 & DV & DS  & SMSR   & Algorithm 1 & Algorithm 2 \\ \hline
Face Sequence 1 &   0.0926  &   0.1559  & 0.5325    & 0.4677 & 0.0531 & 0.0636 & 0.1893 & 0.0443 & {\bf{0.0404}}\\ \hline
Face Sequence 2 &   0.0819  &   0.1503  & 0.9266    & 0.7909 & 0.0457 & 0.0569  & 0.2133  & {\bf{0.0381}} & 0.0392 \\ \hline
Face Sequence 3 &   0.1057 &   0.1252  & 0.5274    & 0.5474 & 0.0346 & 0.0374 & 0.1345  &  0.0294 & {\bf{0.0280}}\\ \hline
Face Sequence 4 &   0.0717 &   0.1348  & 0.5392    & 0.5292 & 0.0379 &  0.0428 & 0.0984 &  {\bf{0.0309}}  & 0.0327\\ \hline
Actor Sequence 1 &  0.5226  &   0.0418  & 0.3711    & 0.3708 & - & 0.0891 & 0.0352 & 0.0340 & {\bf{0.0274}} \\ \hline
Actor Sequence 2 &  0.2737  &   0.0532  & 0.2275    & 0.2279 & - & 0.0822 & 0.0334 & 0.0342 & {\bf{0.0289}} \\  \hline
Paper Sequence   &   0.0827 &   0.0918  &  0.0842   & 0.0801 &  - & 0.0612 & - & 0.0394 & {\bf{0.0338}} \\ \hline
T-shirt Sequence  &  0.0741 &   0.0712  &  0.0644   & 0.0628 &  - & 0.0636 & - & {\bf{0.0362}} & 0.0386 \\ \hline
\end{tabular}
\caption{ \small{Statistical comparison of our method with other competing approaches. Quantitative evaluations for SMSR \cite{ansari2017scalable} and DV \cite{garg2013dense} are not performed by us due to the unavailability of their code, and therefore, we tabulated their reconstruction error from their published work.}} \label{tab:statisticalresults}
\end{table*}

\noindent
{\bf{2. Results on Paper and T-shirt dataset:}}
To evaluate our performance on smooth deforming surfaces, we used Varol et.al. \cite{varol2012constrained} `kinect\_paper' and `kinect\_tshirt' datasets. This dataset provides real condition to test the performance of NRSfM algorithm. It provides sparse SIFT \cite{lowe1999object} feature tracks and noisy depth information captured from Microsoft Kinect for all the frames. As a result, to get dense 2D feature correspondences of the non-rigid object for all the frames becomes difficult. To circumvent this issue, we used Garg et.al.  algorithm \cite{garg2011robust} to estimate the measurement matrix. Numerically, we compute the correspondence of the deforming subject within $\m x_{\m w}$ = $(253, 253, 508, 508)$, $\m y_{\m w}$ = $(132, 363, 363, 132)$ rectangular window across 193 frames for kinect\_paper sequence. For kinect\_tshirt sequence, we considered rectangular window of $\m x_{\m w}$ = $(203, 203, 468, 468)$, $\m y_{\m w}$ = $ (112, 403, 403, 112)$ across 313 frames. Fig.(\ref{fig:real6res}) shows couple of 3D reconstruction results on these sequence with comparative results specified in Table (\ref{tab:statisticalresults}).

\noindent
{\bf{3. Results on Actor dataset:}}
Beeler et.al. \cite{beeler2011high} introduced Actor dataset for high-quality facial performance capture. This dataset is composed of 346 frames captured from seven cameras with 1,180,232 vertices. The dataset captures the fine details of facial expressions which is extremely useful in the testing of NRSfM algorithms. Nevertheless, for our experiment, we require dense 2D image feature correspondences across all images as input, which we synthesized using ground-truth 3D points and synthetically generated orthographic camera rotations. To maintain the consistency with the previous works in dense NRSfM for performance evaluations, we synthesized two different datasets namely Actor Sequence1 and Actor Sequence2 based on the head movement as described in Ansari et.al. work \cite{ansari2017scalable}. Fig.(\ref{fig:real6res}) show the dense detailed reconstruction that is achieved using our algorithms. Table (\ref{tab:statisticalresults}) clearly indicates the superior  performance of our approach on this high-quality dense dataset.

\noindent
{\bf{4. Results on Face, Heart, Back dataset:}}
To evaluate the variational approach to dense NRSfM  Garg et.al.\cite{garg2013dense} introduced this dataset. This dataset is composed of monocular video's captured in a natural environment with varying lighting condition and large displacements. It consists of three different videos with 120, 150 and 80 frames for face sequence, back sequence and heart sequence respectively. Furthermore, this dataset provides dense 2D feature tracks for these 3 categories. Specifically, face, back and heart sequence is composed of 28332, 20561, and 68295 features tracks respectively. Ground-truth 3D is not available with this dataset for evaluation. Fig.(\ref{fig:real6res}) show some qualitative 3D reconstruction results on face, back and heart sequence. The qualitative results ---shown in Fig.(\ref{fig:real6res}), demonstrate that our approach is able to estimate the 3D reconstruction of a deforming subject reliably and accurately.

\begin{figure}
\centering
\subfigure [\label{fig:timing_per}] 
{\includegraphics[width=0.24\textwidth, height=0.133\textheight]{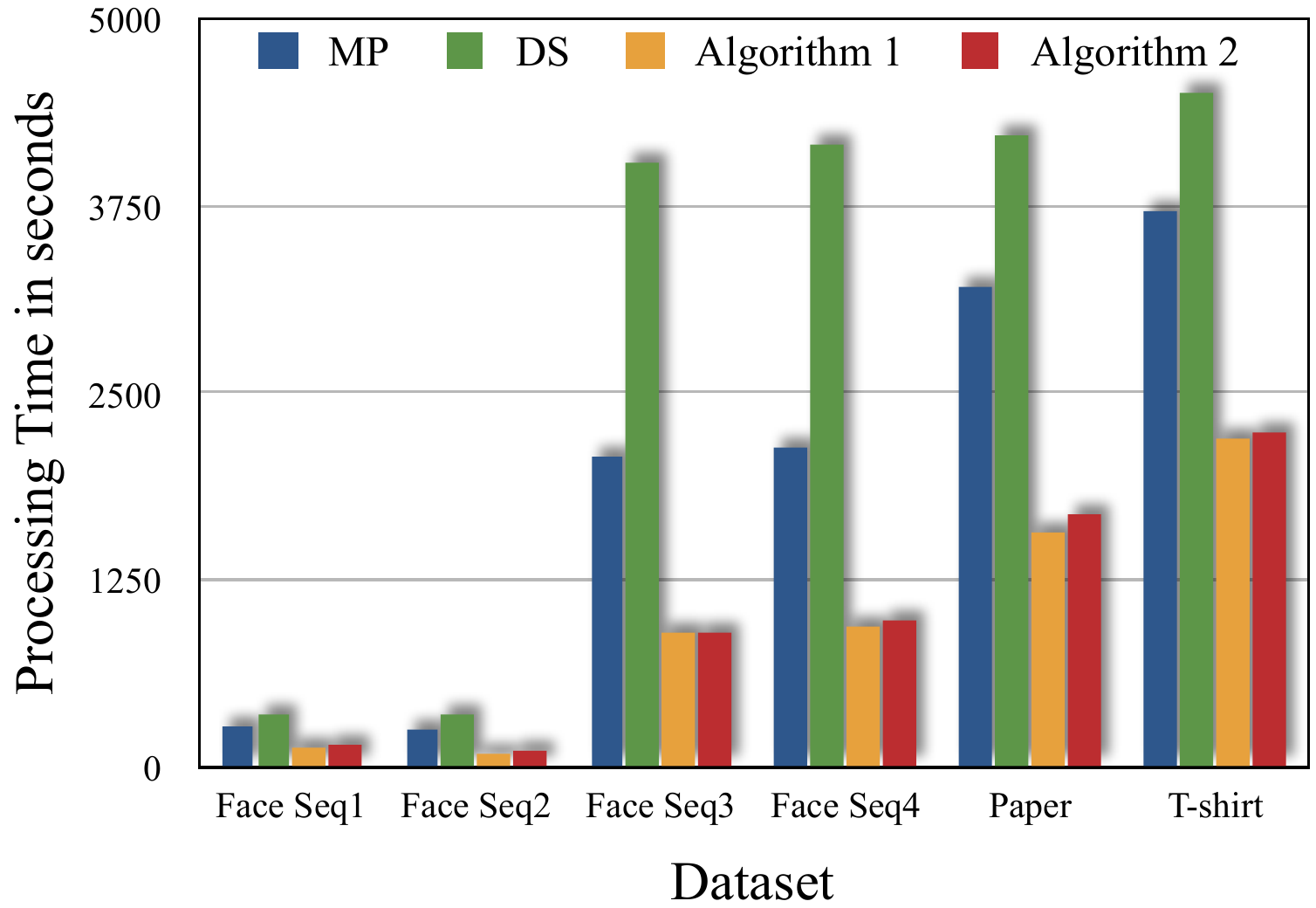}}
\subfigure [\label{fig:noise_per}] {\includegraphics[width=0.24\textwidth, height=0.14\textheight]{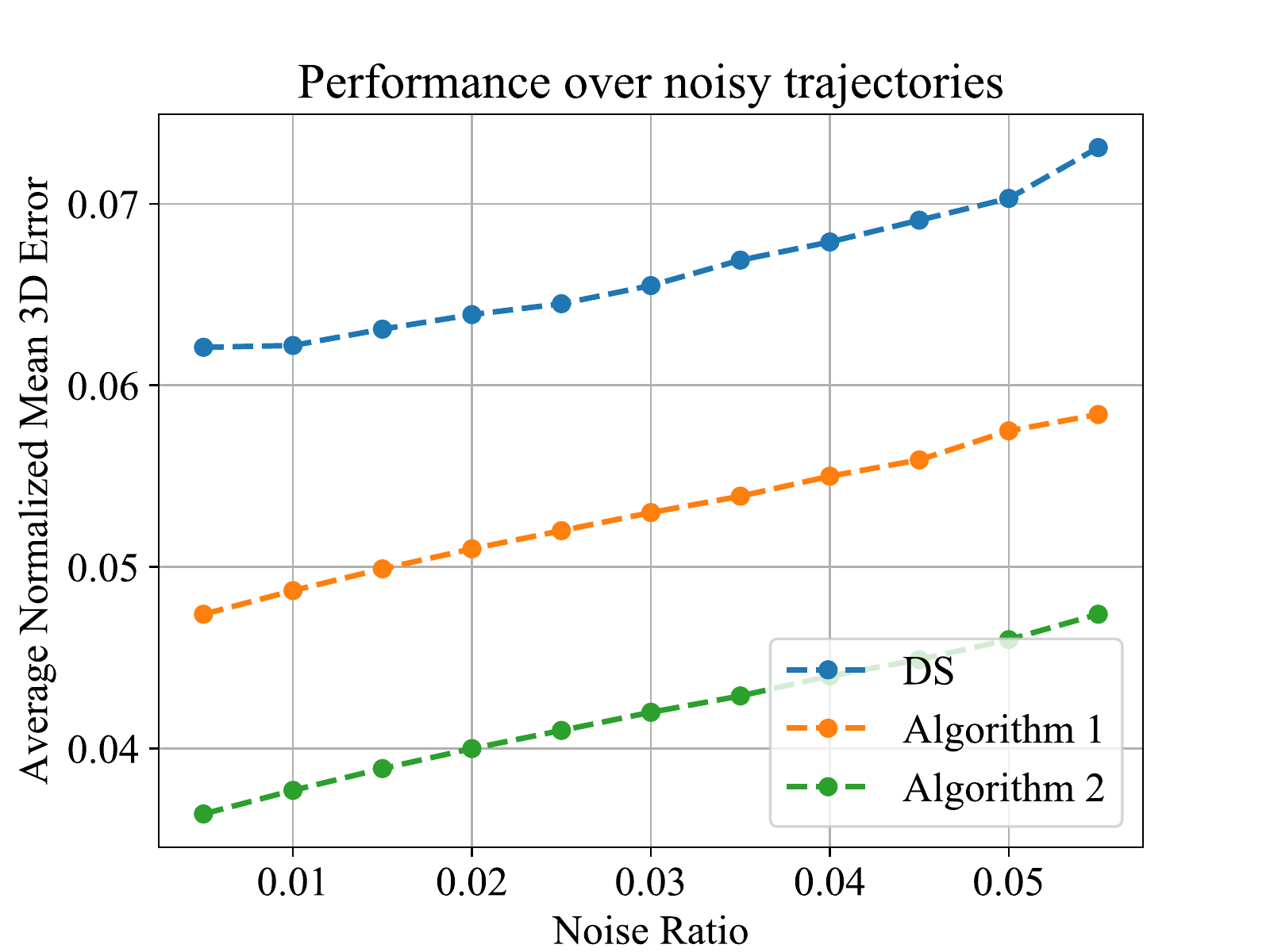}}
\caption{ \small{ (a) Processing Time Comparison (b) Performance Comparison on Noisy Trajectories} }
\label{fig:evaluationNoiseandMD}
\end{figure}

\subsection{Algorithmic Analysis}
\noindent
We performed several other experiments to understand the behavior of our algorithm under different input parameters and evaluation setup. In practice these experiments help analyze the practical applicability of our algorithm.\\

\noindent
{\bf{1. Processing Time and Convergence:}} The execution time for both the algorithm (Algorithm 1 and Algorithm 2) is more or less same. Nevertheless, in comparison to other previous methods our processing time is far better. We computed the processing time on commodity desktop machine with 16GB RAM suing MATLAB R2019b software. Fig.(\ref{fig:timing_per}) show the processing time of our method in comparison to other methods on different datasets. Ideally, our algorithm takes 120-150 iteration to provide an optimal solution to the problem.

\noindent
{\bf{2. Performance over noisy trajectories:}}
We utilized the standard experimental procedure to analyze the behavior of our algorithm under different noise levels. Similar to the work of Lee et.al.\cite{lee2013procrustean}, we added the Gaussian noise to the input trajectories. The standard deviation of the noise are adjusted as $\sigma_{g} = \lambda_g\textrm{max}\{|\m W|\}$ with $\lambda_g$ varying from 0.01 to 0.055. Fig.(\ref{fig:noise_per}) show the quantitative comparison of our approach with recent algorithm DS \cite{dai2017dense}. The graph show the average 3D reconstruction error of all the four synthetic face dataset \cite{garg2013dense}. The statistics indicate that our algorithms are more resilient to noise than other competing methods. Specifically, Algorithm 2 performs better with noisy data due to the low-dimensional projection of the Grassmannians to perform grouping, which inherently provides robust representation of subspace in presence of noisy trajectories.

\begin{figure}
\centering
\subfigure [\label{fig:algo1_sing_val}] 
{\includegraphics[width=0.24\textwidth, height=0.14\textheight]{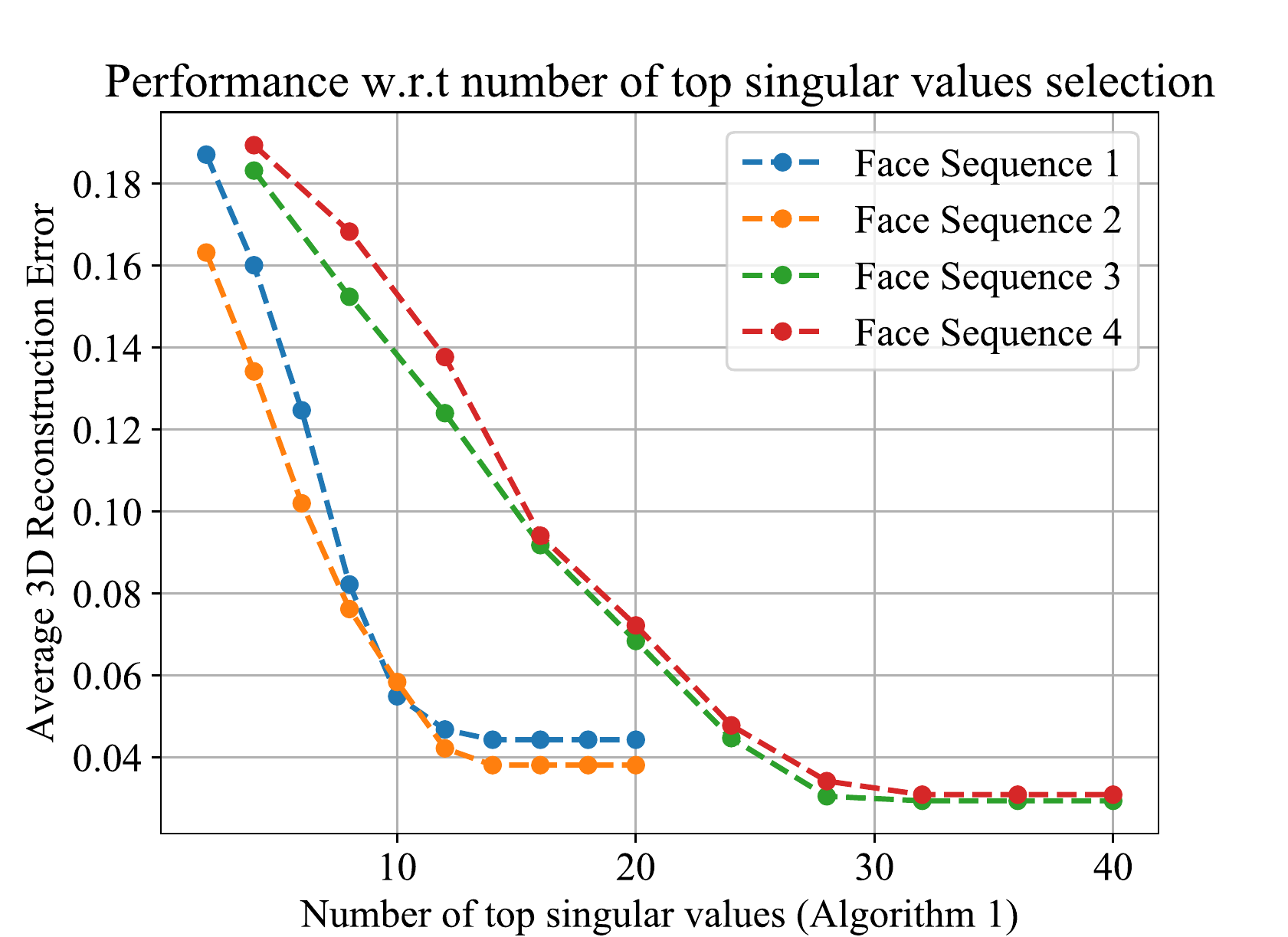}}
\subfigure [\label{fig:algo2_sing_val}] {\includegraphics[width=0.24\textwidth, height=0.14\textheight]{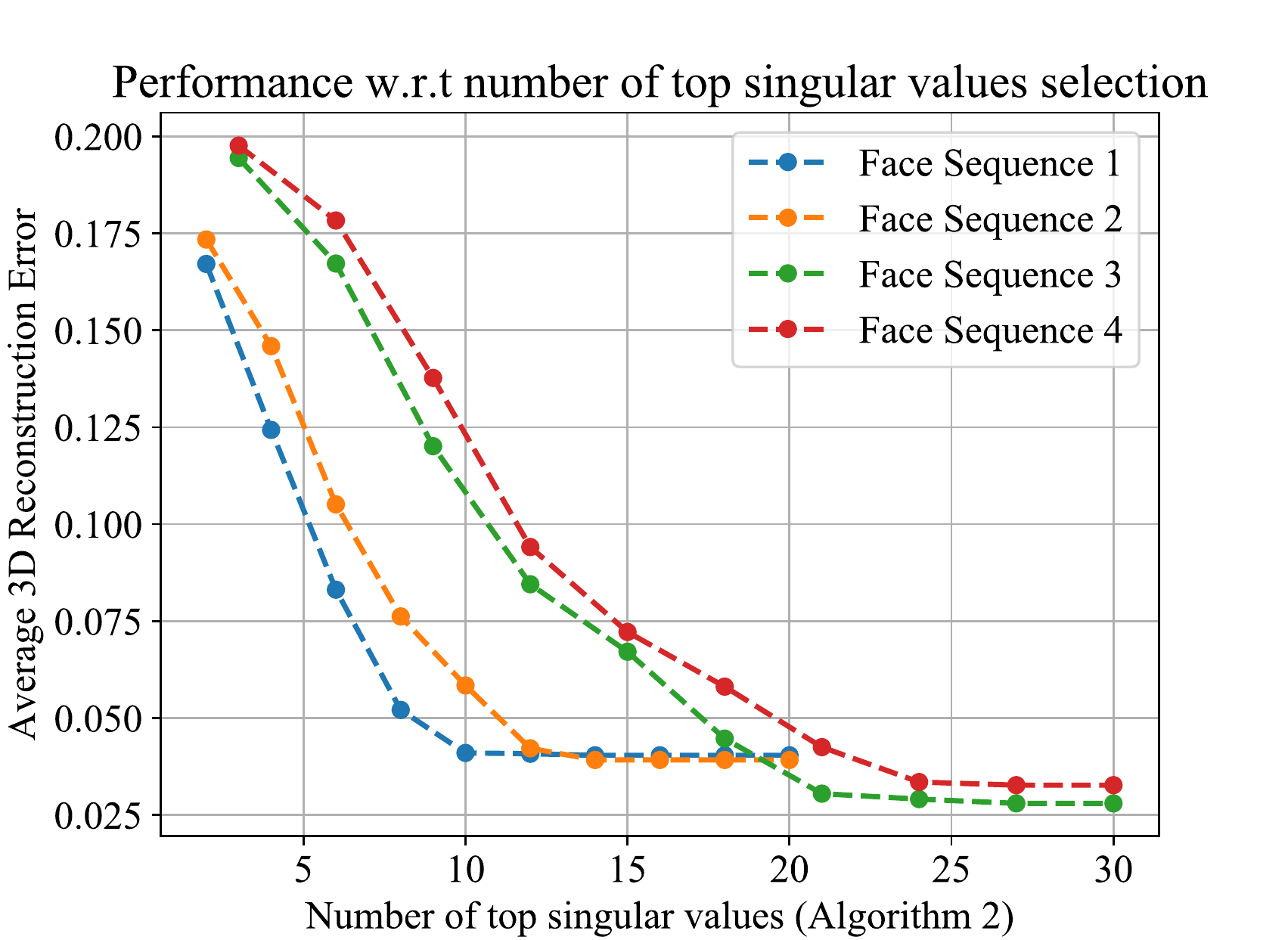}}
\caption{\small{ Change in the average 3D reconstruction accuracy with respect to number of singular vectors and singular values used. } }
\label{fig:evaluationtopsingular}
\end{figure}

\noindent
{\bf{3. Performance with change in the number of singular values:}}
The integral value of `$\m p$' in $\mathcal{G}(\m p, \m d)$ i.e., the number of top singular vectors to represent Grassmannians and, corresponding singular values to perform reconstruction can directly affect the performance of our algorithm. Yet, it has been observed over several experiments that we need relatively few singular values and singular vectors ---in comparison to the number of trajectories, to recover dense 3D reconstruction of the deforming object. Fig.(\ref{fig:algo1_sing_val}) and Fig. (\ref{fig:algo2_sing_val}) show the change in average 3D reconstruction with the different values of `$\m p$' for synthetic face dataset \cite{garg2013dense}. The graph shows that for both Algorithm 1 and Algorithm 2, 10-15 singular values are good enough for Face Sequence 1-2 and, 25-30 singular values are sufficient for Face Sequence 3-4. Note that these sequence have 28,880 trajectories and therefore, to reconstruct each vectorial point can be severely expensive. In constrast, our linear subspace representation can handle it easily.

\begin{figure*}
\centering
\subfigure [\label{fig:repro_face1}] 
{\includegraphics[width=0.24\textwidth, height=0.14\textheight]{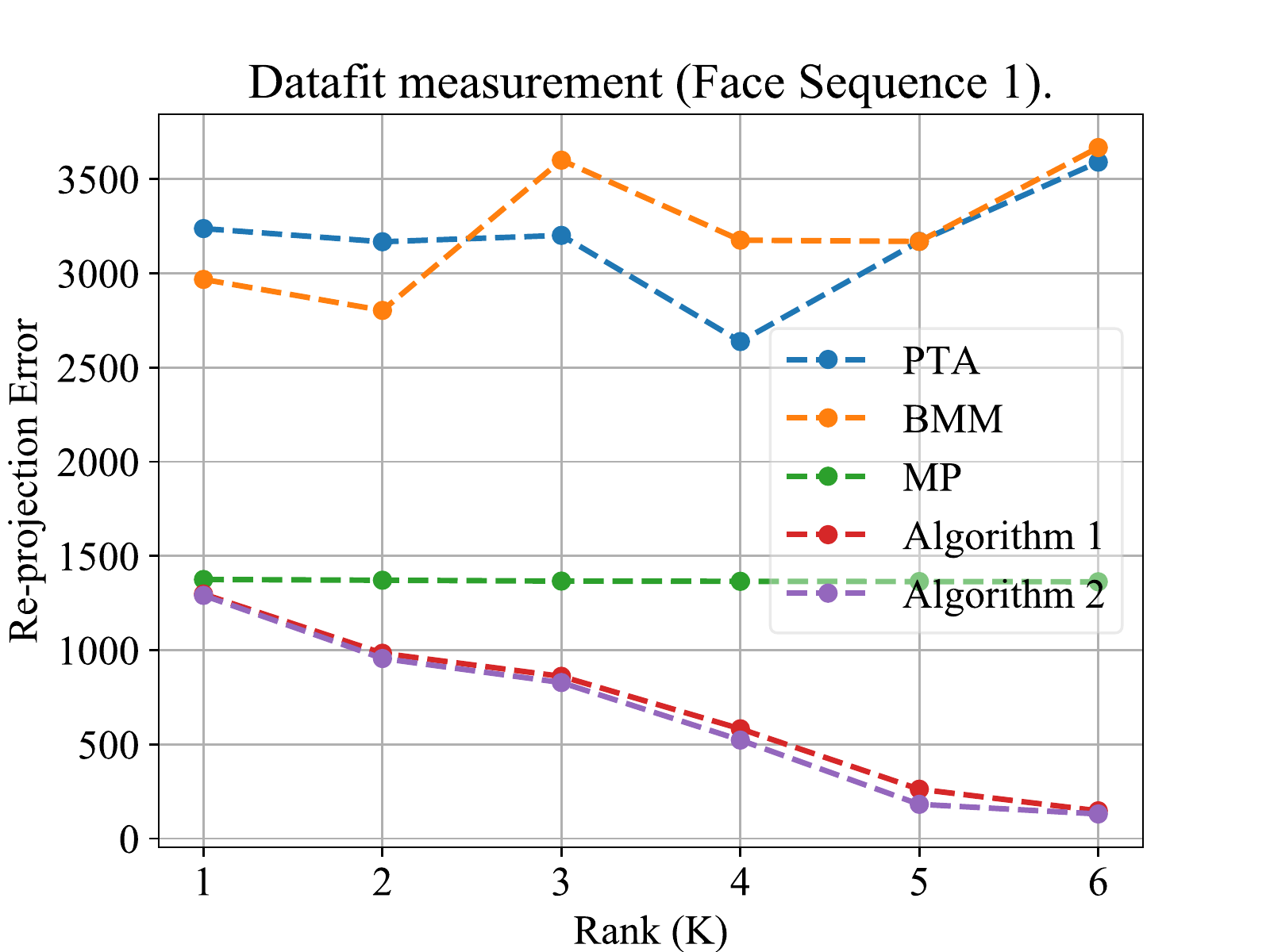}}
\subfigure [\label{fig:repro_face2}] {\includegraphics[width=0.24\textwidth, height=0.14\textheight]{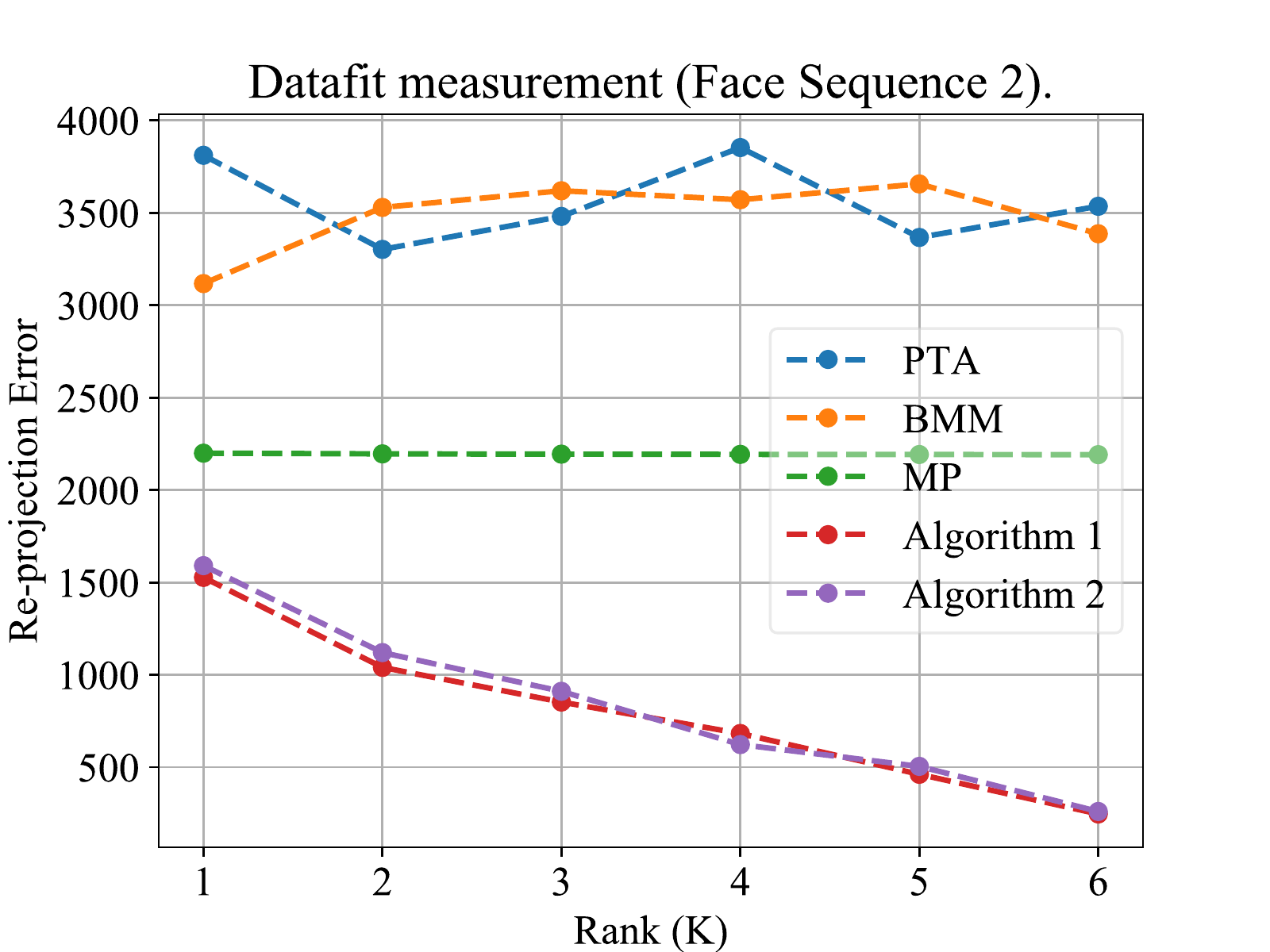}}
\subfigure [\label{fig:repro_face3}] 
{\includegraphics[width=0.25\textwidth, height=0.14\textheight]{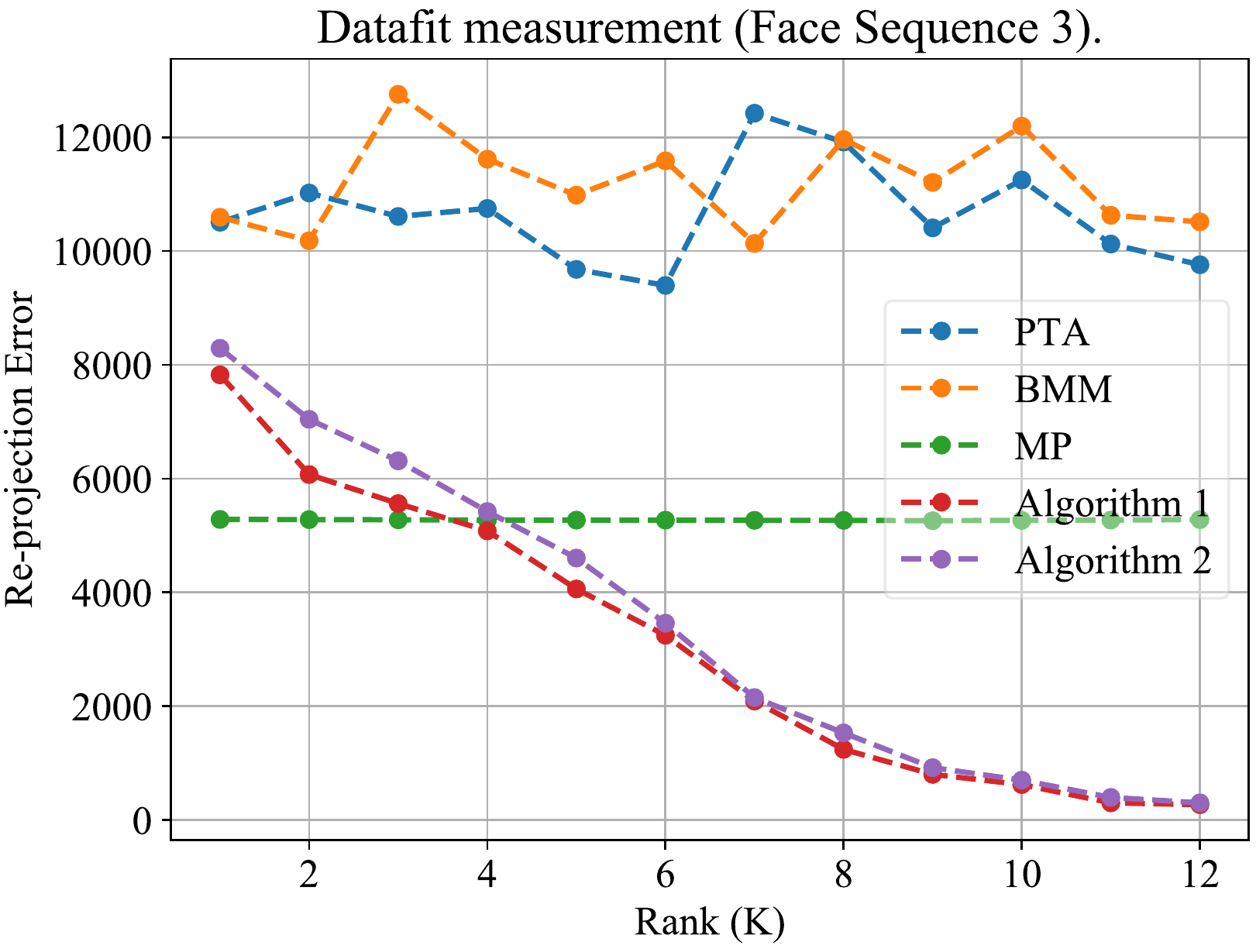}}
\subfigure [\label{fig:repro_face4}] {\includegraphics[width=0.25\textwidth, height=0.14\textheight]{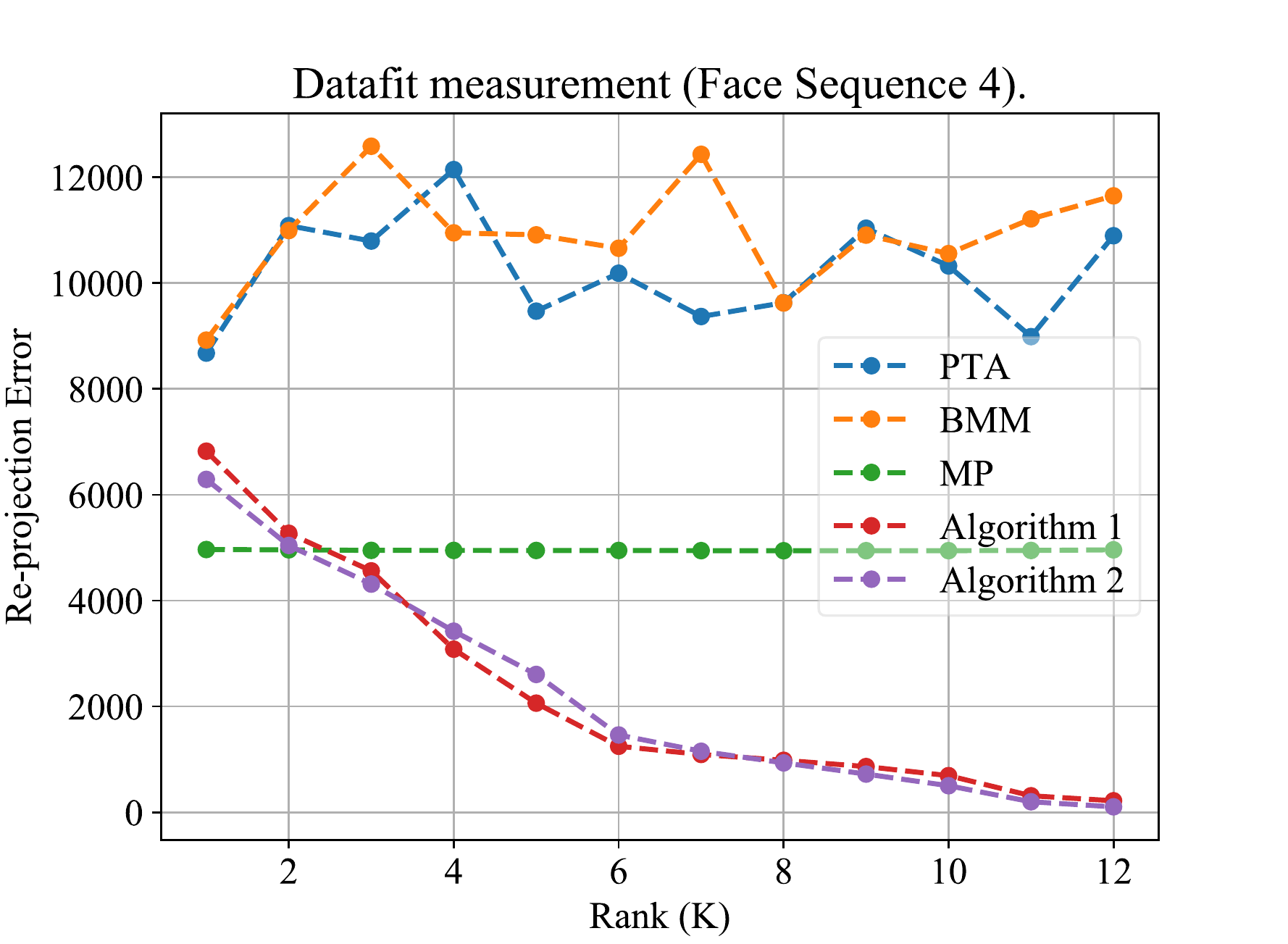}}
\caption{ \small{ The measure of re-projection error constraint i.e $\| \m W- \m R \m S\|_{\m F}$ as a function of rank on synthetic face dataset \cite{garg2013dense}. } }
\label{fig:repro_eval}
\end{figure*}

\begin{figure*}
\centering
\subfigure [\label{fig:gterr_face1}] 
{\includegraphics[width=0.24\textwidth, height=0.14\textheight]{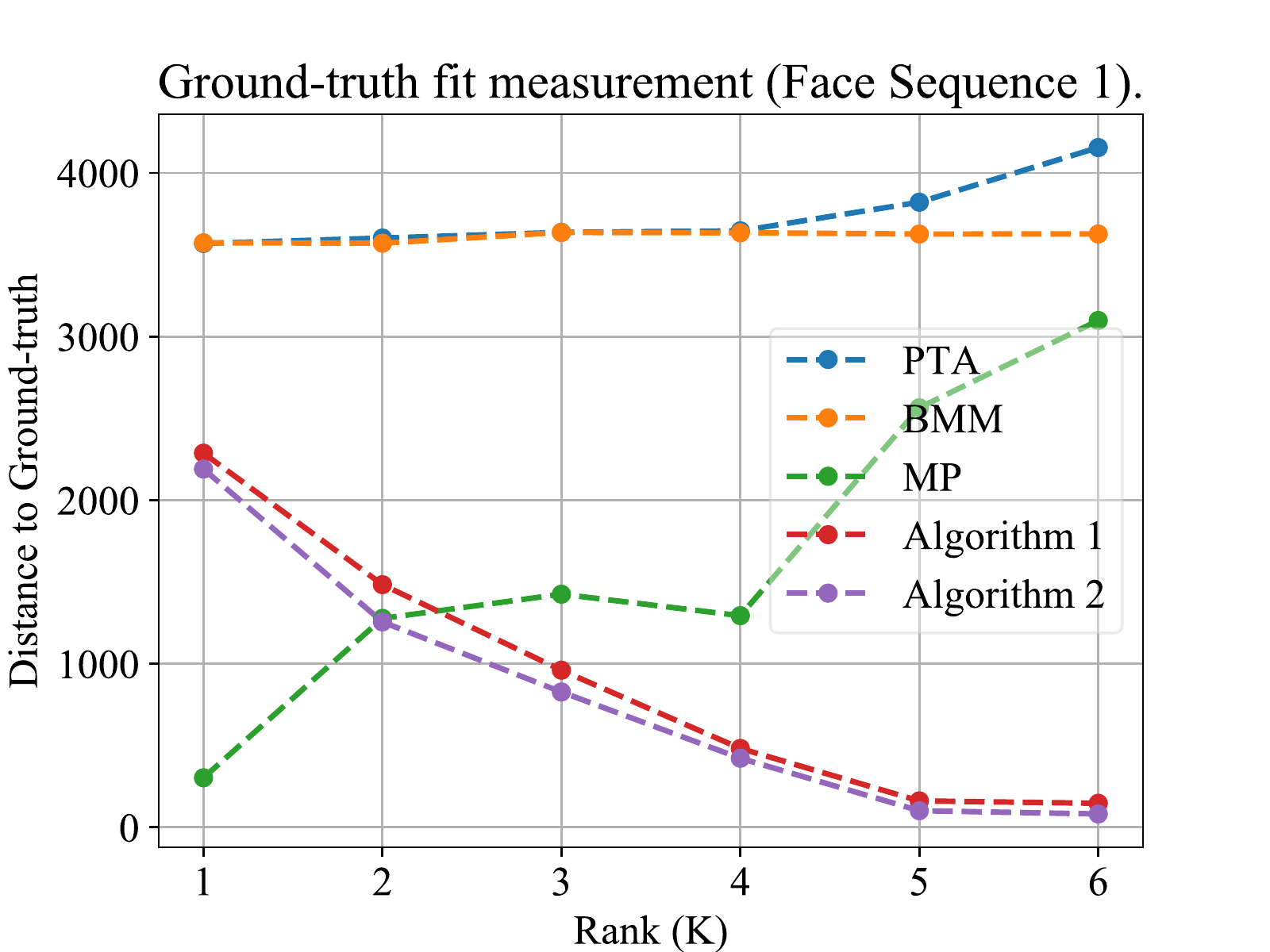}}
\subfigure [\label{fig:gterr_face2}] {\includegraphics[width=0.24\textwidth, height=0.14\textheight]{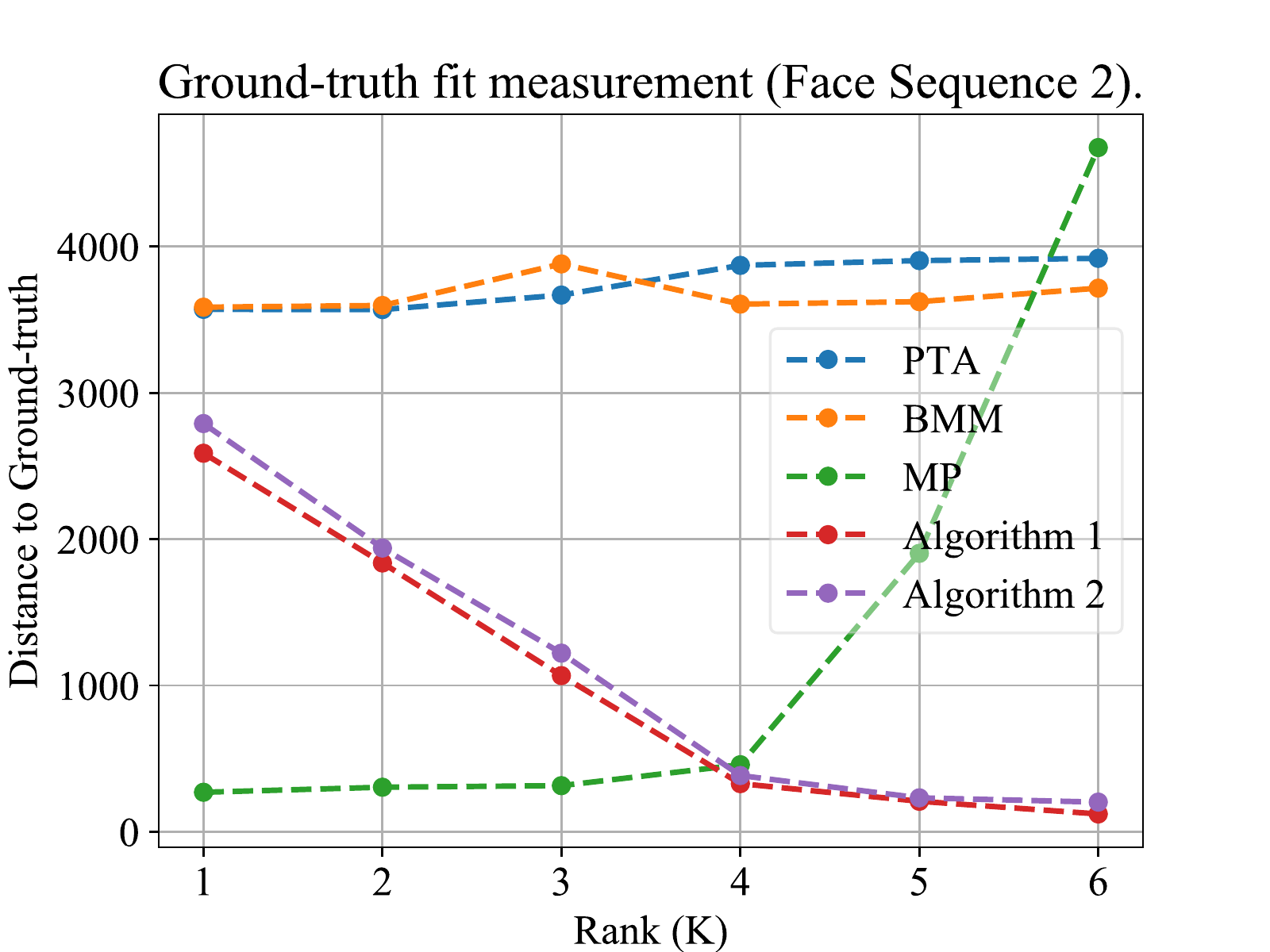}}
\subfigure [\label{fig:gterr_face3}] 
{\includegraphics[width=0.25\textwidth, height=0.14\textheight]{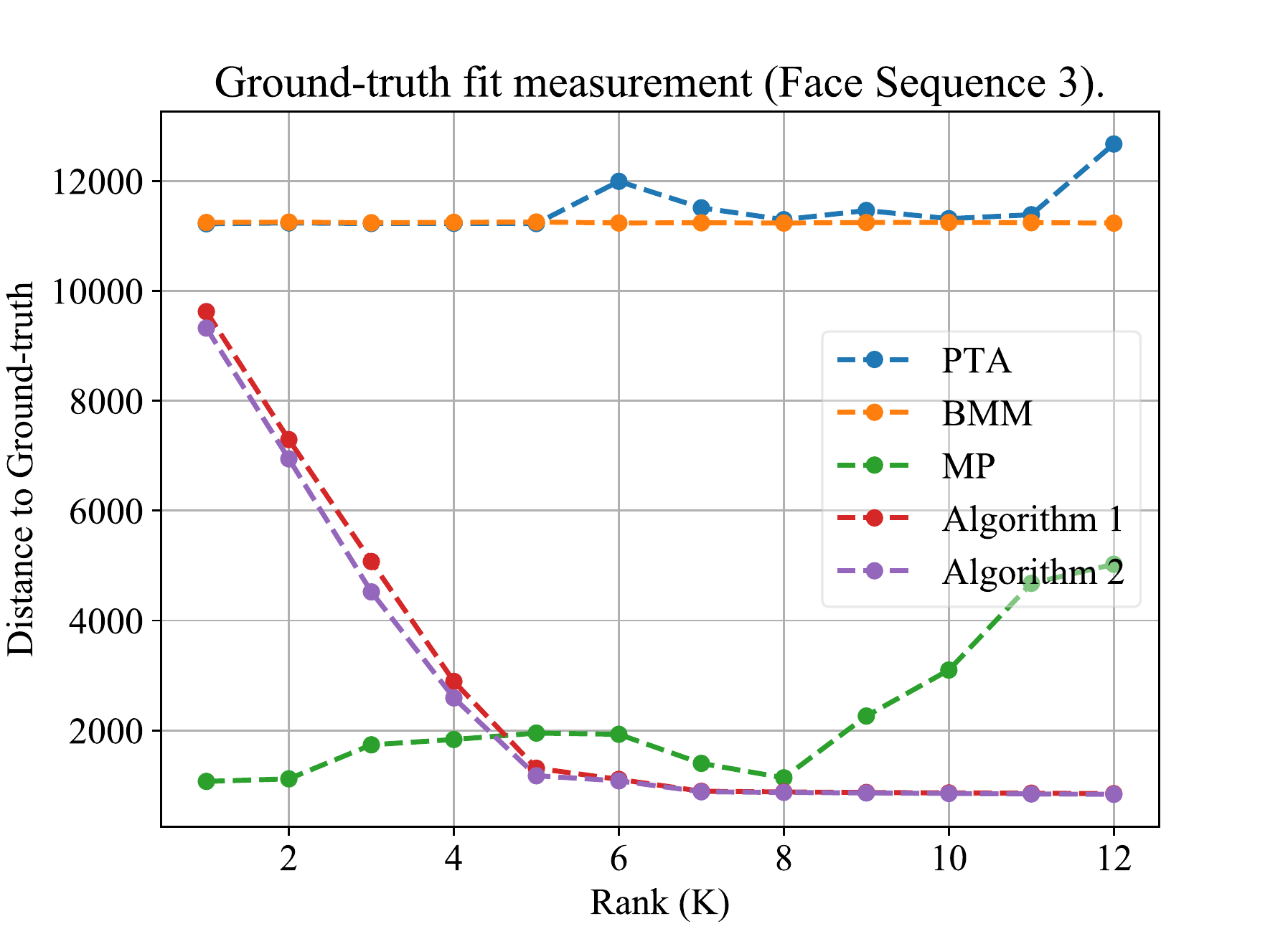}}
\subfigure [\label{fig:gterr_face4}] {\includegraphics[width=0.25\textwidth, height=0.14\textheight]{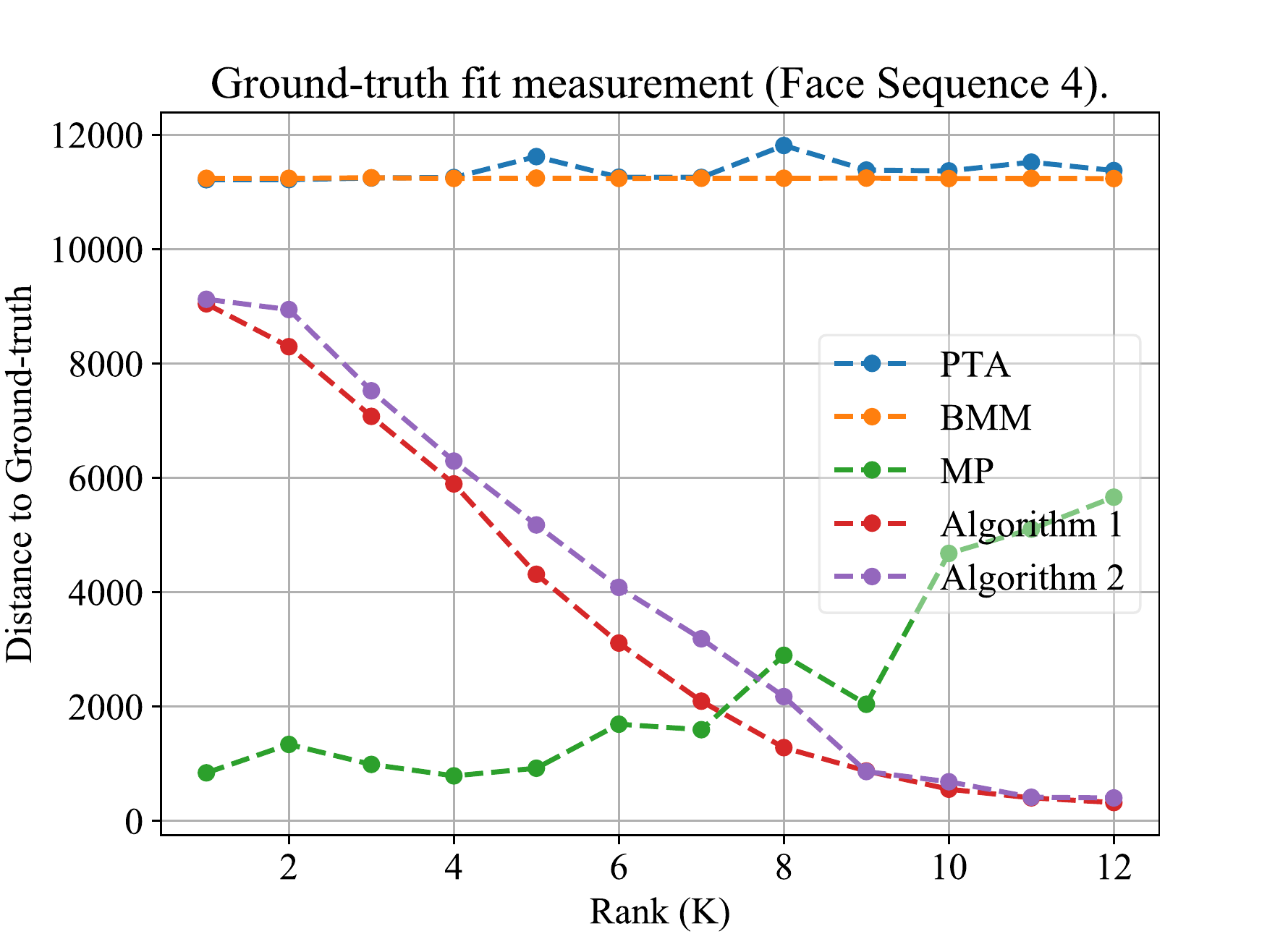}}
\caption{ \small{The measure of ground-truth shape fit i.e, the distance between the obtained shape and ground truth shape ($\|\m S_{\m G\m T} - \m S\|_{\m F}$) as a function of rank on synthetic face dataset \cite{garg2013dense}.}}
\label{fig:gterr_eval}
\end{figure*}

\noindent
\textbf{4. Analysis of Data-fit with Variation in the Rank:} Inspired by the recent work \cite{ornhag2020unified}, we performed this test to analyse the dependence of the algorithm on the output rank. Specifically, we measure the datafit and ground-truth fit which is defined as $\|\m W - \m R\m S\|_{\m F}$ and $\|\m S_{\m G\m T}-\m S\|_{\m F}$ respectively. This experiment revels the competence of the algorithm to fit the trajectory/shape based on the selection of the rank. The Fig.(\ref{fig:repro_eval}) and Fig.(\ref{fig:gterr_eval}) clearly indicates that sparse NRSfM algorithm \cite{dai2014simple} \cite{akhter2009nonrigid} fails to handle the dense deforming shapes. Neither, reprojection error nor ground-truth fit is maintained with increase in the 'K' value (rank).
Additionally, dense NRSfM approach with only global constraint \cite{paladini2012optimal} (MP) fails to capture the local deformation properly, hence, datafit to the ground-truth fails to correlate with the re-projection error with increase in the rank (K value). In contrast, our algorithm has expected trend to both reprojection error and ground-truth error with the variation in the rank.  Both of our algorithms have high correlation between the two measures. Note that MP \cite{paladini2012optimal} \cite{paladini2009factorization} algorithm estimates both rotation and translation for NRSfM problem, however, for consistency we did not consider the translation component for plotting these graphs.

\begin{figure}
\centering
\subfigure [\label{fig:algo1_ablation}] 
{\includegraphics[width=0.24\textwidth, height=0.14\textheight]{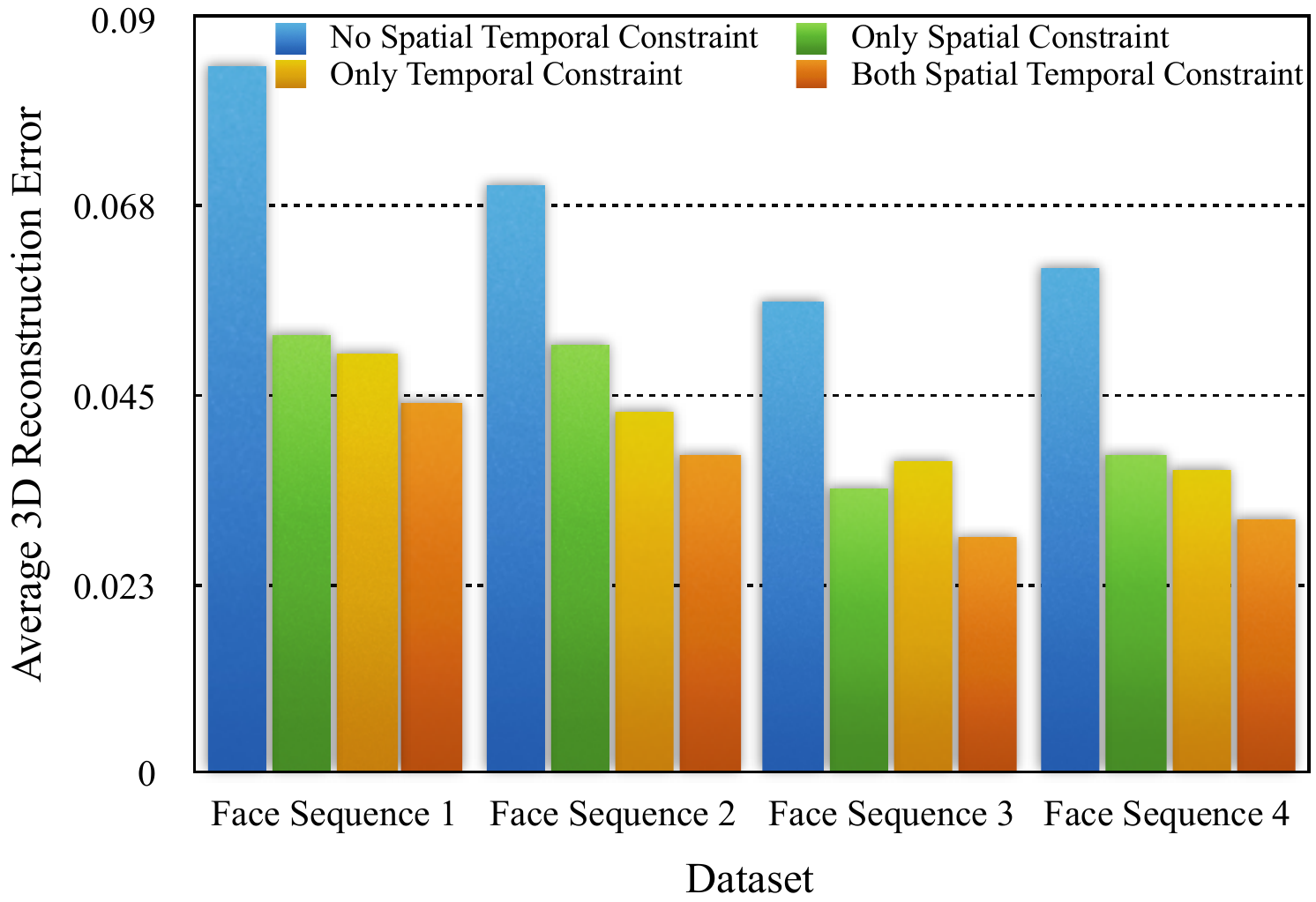}}
\subfigure [\label{fig:algo2_ablation}] {\includegraphics[width=0.24\textwidth, height=0.14\textheight]{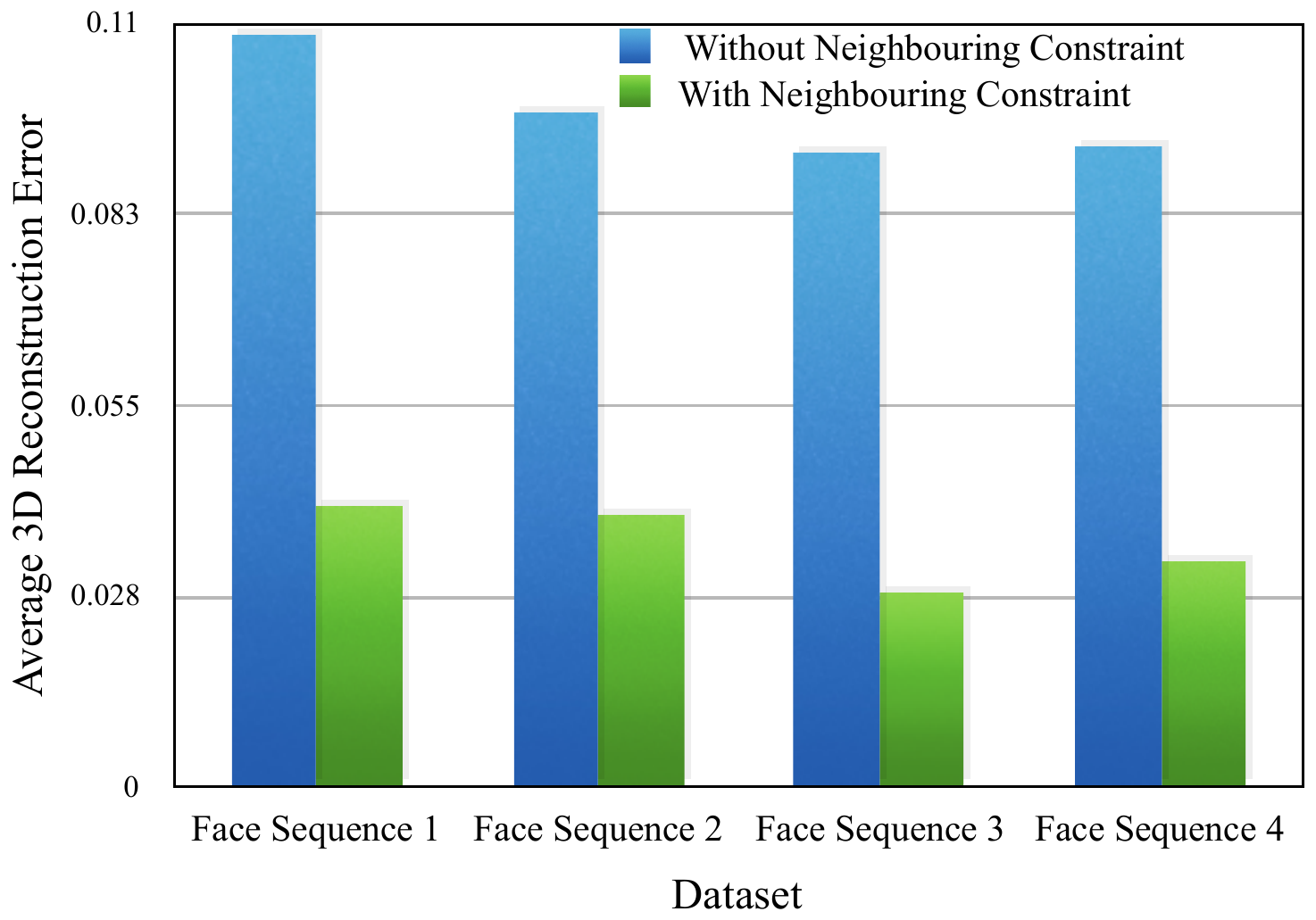}}
\caption{ \small{ (a)-(b) Ablation test results for Algorithm 1 and Algorithm 2 respectively.} }
\label{fig:abalationstudy}
\end{figure}

\noindent
{\textbf{5. Ablation Test:}} In this paper, we proposed two optimization algorithms that are composed of several constraints. To understand the importance of each constraint, we performed an ablation test. Algorithm 1 has both spatial and temporal subspace constraint, whereas, Algorithm 2 has spatial subspace constraint along with spatial neighbouring constraint.  To perform this task for Algorithm 1, we observe the performance of our formulation under four different setups: (a) without any spatio-temporal constraint (b) with only spatial constraint  (c) with only temporal constraint (TP), and (d) with both the spatial-temporal constraint. Fig.(\ref{fig:algo1_ablation}) show the variations in average 3D reconstruction errors under these setups on four synthetic face sequence. The statistics clearly illustrate the importance of both the constraints in our formulation. Similarly, for Algorithm 2, we analysed the performance before and after imposing the neighboring constraint. The result is shown in Fig.(\ref{fig:algo2_ablation}) that demonstrates the importance of the imposed constraint. Just to remind the readers that for Algorithm 2, we did not consider the temporal constraint. We argued that temporal information are generally not available in real-world cases.

\begin{figure}
\centering
\includegraphics[width=0.48\textwidth] {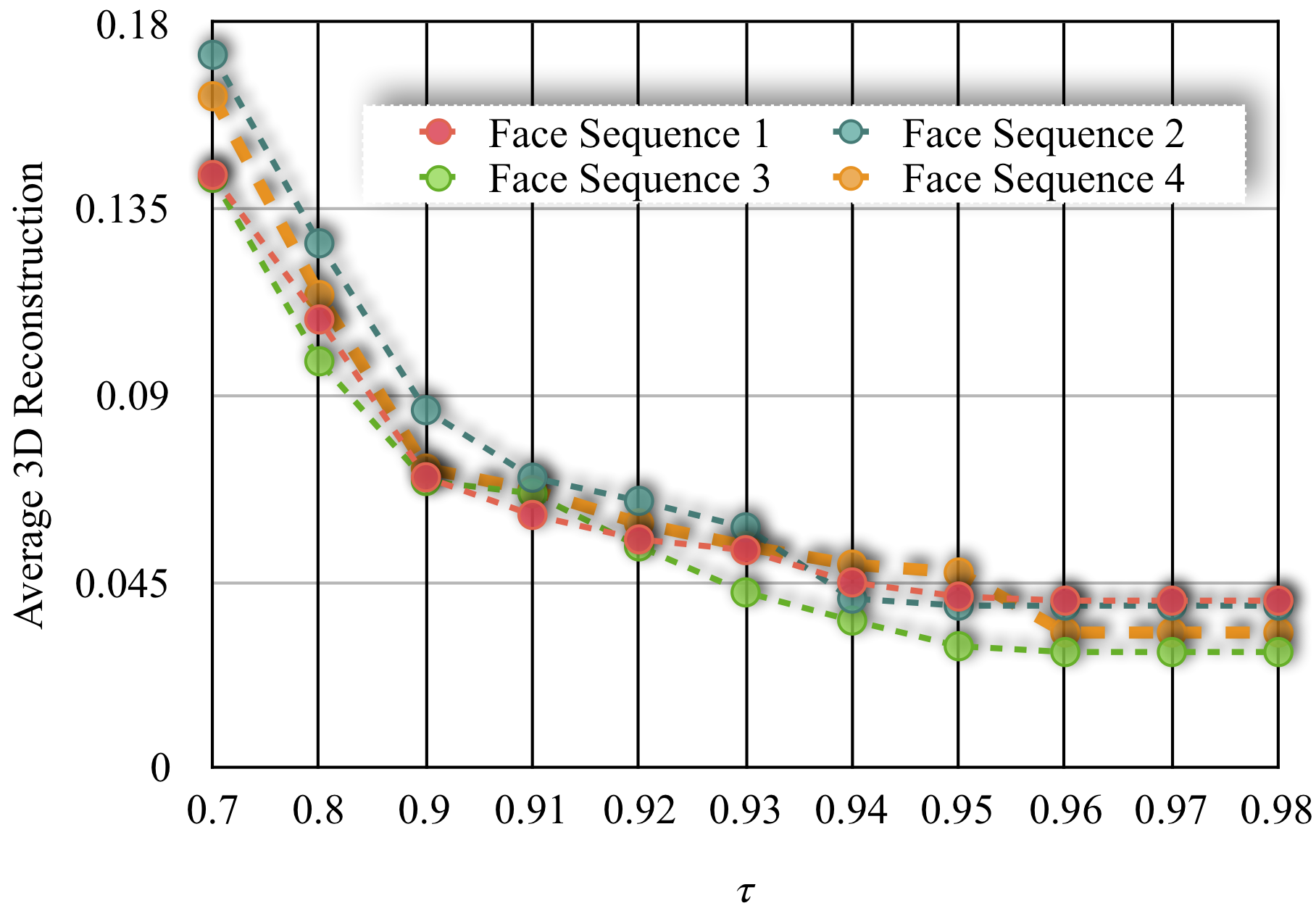}
\caption{\small{Variation in the 3D reconstruction accuracy w.r.t $\tau$.}}
\label{fig:dimesion_dependence}
\end{figure}

\noindent
\textbf{6. Dependence of the algorithm 2 on variable $\tilde{\m d}$:} Dimensionality reduction to group the grassmann points is one of the critical aspect of Algorithm 2. To determine the dimension to which we should project for better results is a key-concern. We used well-known procedure of cumulative energy of singular vectors to get the value of $\tilde{\m d}$. Mathematically, let $\Omega$ be the set that stack all the Grassmannians and $\sigma_{\m i}$ be the $\m i^\textrm{th}$ singular value of  $\Omega\Omega^{\m T}$, then 
\begin{equation}
\tilde{\m d} = \underset{\m d_\textrm{opt}} {\textrm{argmin}} \frac{\sum_{\m i=1}^{\m d_\textrm{opt}}\sigma_{\m i}}{\sum_{\m i=1}^{\m d}\sigma_{\m i}} \geq \tau
\end{equation}
where $\tau$ can vary from 0 to 1 and $\m d_\textrm{opt}$ (optimal dimension) is a positive integer. We put $\tau = 0.97$ for all our experiment. Fig.(\ref{fig:dimesion_dependence}) show the variations in the reconstruction error with the value of $\tau$ for synthetic face dataset \cite{garg2013dense}. It is observed that for different dataset the value of suitable $\tilde{\m d}$ is different. The point to note is that if the reduced dimension is less than the intrinsic dimension, the samples may lose important information for better grouping of Grassmannians. For our task, in general, $\tau = 0.97$ works well for all the dataset.


\section{Practical Limitations}
Firstly, our method assumes fairly good dense 2D feature correspondence is provided as input. Nonetheless, estimating robust dense 2D feature correspondences for a deforming surface across frames in itself is a very challenging problem to solve. The main challenges come from the fact that the illumination of the deforming object keeps changing over time. Consequently, a passive approach to establish correspondences may lead to wrong results.
Secondly, our representation may fail to handle non-rigid deformation such as stretching and squashing of an object. For example: stretching a rubber sheet or deflating a balloon. Such deformations are hard to handle due to substantial change in the global structure of the shape ---area/volume or projected object size can differ considerably over frames. Lastly, object deformation recorded under a distinct camera trajectory can provide different results.

\section{Conclusion and Future Work}
In this work, we introduced a new representation to solve the problem of dense non-rigid structure from motion. Exploiting both local and global deformation constraints, our algorithm uses the new representation to make the idea of joint segmentation and reconstruction scalable and therefore, is able to obtain the 3D structure of dense deforming surfaces with higher accuracy. Employing a unified spatial-temporal idea to blend the information from both shape and trajectory space, our algorithm demonstrated leading performance on the benchmark datasets. Later, we extended our formulation to a more practical setting, where, temporal shape information is not known a prior and input can be noisy. We used the assumption that a group of neighboring trajectories may span a similar linear subspace. To make our formulation robust to noise, we project the manifold representation to lower dimension for better clustering, thereby implicitly improving the 3D reconstruction. In particular, our algorithm shows notable accuracy in the presence of noise and complex deformations, where other methods may fail.

It has been observed that when the same object deformation is recorded under different camera motion, the non-rigid structure from motion algorithms behaves differently. In the future, we plan to extend our algorithm to handle such a situation robustly. Additionally, how far can we apply the new finding on smooth motion assumption in our pipeline and, when the low-rank model may fail is left as an extension to this work \cite{kumar2020non}.

\ifCLASSOPTIONcaptionsoff
  \newpage
\fi



%
\bibliography{bibliography/denseNRSFM.bib}

\newpage
\bibliographystyle{IEEETrans}
%

\begin{IEEEbiography}
[{\includegraphics[width=1in,height=1.25in,clip,keepaspectratio]{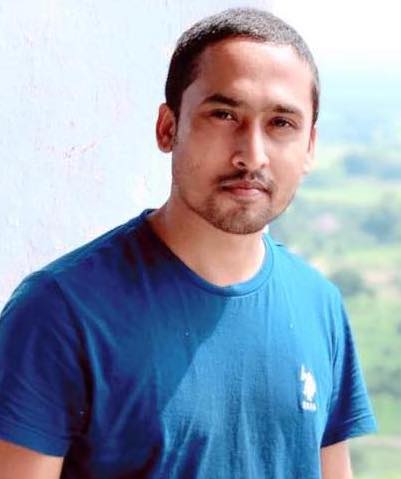}}]
{Suryansh Kumar} presently holds the position of Professorship for Computer Vision in the Department of Information Technology and Electrical Engineering at ETH Z\"urich, where he is advised by Prof. Dr. Luc Van Gool on 3D vision projects. He received Ph.D. in Engineering and Computer Science from the Australian National University in 2019. His Ph.D. dissertation on ``Non-Rigid Structure from Motion'' is nominated for J.G. Crawford Prize 2019 at the Australian National University. He received M.S in Computer Science and Engineering from International Institute of Information Technology, Hyderabad (IIIT-H) in 2013. Before joining Australian National University, he worked as a Visiting Scientist in the e-Motion Group at INRIA Rh\^one Alpes Grenoble. After that, he spend one year as a consultant engineer in a computer vision industry-Hyderabad, India. His research interests are geometric computer vision, robotics, abstract algebra, Geometric AI and mathematical optimization. He received best algorithm award from Disney Research for his work on ``Multi-body Non-Rigid Structure from Motion" in NRSFM Challenge at CVPR 2017, Hawaii, USA.
\end{IEEEbiography}

\begin{IEEEbiography}
[{\includegraphics[width=1in,height=1.25in,clip,keepaspectratio]{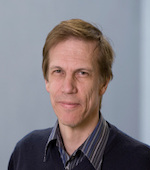}}]
{Luc Van Gool} is a Full Professor at ETH Z\"urich and KU Leuven. He studied Electrical Engineering at the University of Leuven in Belgium. In 1991 he received his Ph.D. degree from the same university, with a dissertation on the use of invariance in computer vision. In 1991 he became an assistant professor in Leuven and in 1996 professor. He still leads a research group in Leuven, that focuses on industrial applications of computer vision. In 1998 he became a full professor at the ETH Z\"urich, where he is the head of Computer Vision Lab in the Department of Information Technology and Electrical Engineering. With his research teams, Luc Van Gool is a partner in several national and international projects, e.g. the EU ACTS project Vanguard, the EU Esprit projects Improofs and Impact, and the EU Brite-Euram project Soquetec. He is also involved in several other projects, that range from fundamental research to application-driven developments. His major research interests include 3D Vision, 2D and 3D object recognition, texture analysis, range acquisition, stereo vision, robot vision, and optical flow. Luc Van Gool has been a member of the program committees of several leading international conferences, including the ICCV, ECCV, and CVPR. In 1998 he received a David Marr Prize at the International Conference on Computer Vision. He is also a cofounder and director of the company Eyetronics, that specialises on 3D modeling and animation, mainly for the entertainment industry and medical applications. The ``ShapeSnatcher'' product received one of the EU EITP prizes in 1998. He has received several awards for his work including FWO Excellence Prize 2015 and, IEEE Computer Society Distinguished Researcher award at ICCV 2017.
\end{IEEEbiography}

\begin{IEEEbiography}
[{\includegraphics[width=1.0in,height=1.25in,clip,keepaspectratio]{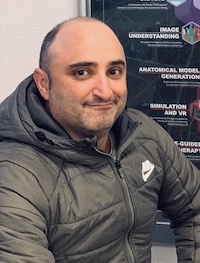}}]
{Carlos Eduardo Porto de Oliveira} Graduated with Bachelor in Social Communication with Film specialization. He is producing and directing documentaries, advertising films and all types of audiovisual content for different brands around the world. Much of his career was spent in Brazil, where he had the opportunity to specialize in animation and special effects. He was working for television channels in Brazil like MTV and ESPN, and also for biggest musical festivals developing all the content for big events. In Switzerland, he went through film studios, animation and film production companies, developing not only films but audio-visual and interactive installations. Currently, he works at ETH-CVL, where he is responsible for the audio-visual communication and special effects implementation, specifically directed to the most important scientific research in the CVL lab.
\end{IEEEbiography}

\begin{IEEEbiography}
[{\includegraphics[width=1.0in,height=1.25in,clip,keepaspectratio]{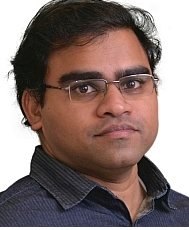}}]
{Anoop Cherian} is a Research Scientist with Mitsubishi Electric Research Labs (MERL) Cam-bridge, MA and an Adjunct Researcher affiliated to the Australian Centre for Robotic Vision (ACRV) at the Australian National University. Previously, he was a Postdoctoral Researcher in the LEAR team at INRIA at Grenoble. He received his B.Tech (honors) degree in computer science and Engineering from the National Institute of Technology, Calicut, India in 2002, his M.S. and Ph.D. degrees in computer science from the University of Minnesota, Minneapolis in 2010 and 2013 respectively. His research interests lie in the areas of computer vision and machine learning.
\end{IEEEbiography}

\begin{IEEEbiography}
[{\includegraphics[width=1.0in,height=1.25in,clip,keepaspectratio]{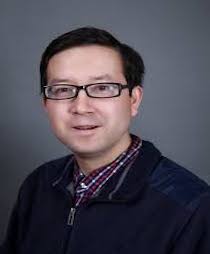}}]
{Yuchao Dai} is a Professor in School of Electronics and Information at Northwestern Polytechnical University, Xi'an, China. He was an ARC DECRA Fellow with the Research School of Engineering at the Australian National University, Canberra, Australia. He received the B.E. degree, M.E degree and Ph.D. degree all in signal and information processing from Northwestern Polytechnical University in 2005, 2008 and 2012, respectively. His research interests include structure from motion, multiview geometry, deep learning, compressive sensing and optimization. He won the Best Paper Award in CVPR 2012, DSTO Best Fundamental Contribution to Image Processing Paper Prize in 2014 and Best Algorithm award in CVPR NRSFM Challenge 2017. He served as an Area Chair for WACV 2019/2020 and ACM MM 2019.
\end{IEEEbiography}


\begin{IEEEbiography}
[{\includegraphics[width=1in,height=1.25in,clip,keepaspectratio]{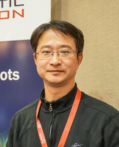}}]
{Hongdong Li}
is a Chief Investigator of the Australia Centre of Excellence for Robotic Vision, Australian National University. He is Associate Director (for Research) for ANU School of Engineering.  He was a visiting professor with the Robotics Institute, CMU during sabbatical in 2017. His research interests include geometric computer vision, pattern recognition and machine learning, vision perception for autonomous driving, and combinatorial optimization.  He is an Associate Editor for IEEE TPAMI, and served as Area Chair for recent years' CVPR, ICCV and ECCV. He was the winner of CVPR Best Paper Award 2012, Marr Prize (Honorable Mention) at ICCV 2017, IEEE ICPR and IEEE ICIP Best Student Paper Winner, DSTO Best Fundamental Contribution to Image Processing Paper Prize at DICTA 2014 and Best algorithm award in CVPR NRSFM Challenge 2017. He is a program co-chair for ACCV 2018 and ACCV 2022.  His research is funded by Australia Research Council, CSIRO, General Motors, Ford Motors, Microsoft Research etc.  He is Presentation Co-Chair for ICCV 2019 and AC for CVPR 2020 and ECCV 2020. \end{IEEEbiography}






\end{document}